\documentclass{ieeeaccess}
\usepackage{cite}
\usepackage{amsmath,amssymb,amsfonts}
\usepackage{algorithmic}
\usepackage{graphicx}
\usepackage{textcomp}
\def\BibTeX{{\rm B\kern-.05em{\sc i\kern-.025em b}\kern-.08em
    T\kern-.1667em\lower.7ex\hbox{E}\kern-.125emX}}

\usepackage{xspace}
\makeatletter
\DeclareRobustCommand\onedot{\futurelet\@let@token\@onedot}
\def\@onedot{\ifx\@let@token.\else.\null\fi\xspace}

\def\eg{\emph{e.g}\onedot} 
\def\ie{\emph{i.e}\onedot}

\def\etal{\emph{et al}\onedot}
\makeatother

\usepackage{colortbl}

\usepackage{algorithm2e}
\usepackage{multirow}

\usepackage[normalem]{ulem} 

\usepackage{url}
\usepackage{cite}

\def\thevol{8}
\def\theyear{2020}
\begin{document}
\history{Received July 2, 2020, accepted August 31, 2020, date of publication September 7, 2020, date of current version October 13, 2020.}
\doi{10.1109/ACCESS.2020.3022063}

\title{An Entropy Clustering Approach for Assessing Visual Question Difficulty}

\author{
\uppercase{Kento Terao}\authorrefmark{1},
\uppercase{Toru Tamaki}\authorrefmark{1},
\IEEEmembership{Member, IEEE},
\uppercase{Bisser Raytchev}\authorrefmark{1},
\IEEEmembership{Member, IEEE},
\uppercase{Kazufumi Kaneda}\authorrefmark{1},
\IEEEmembership{Member, IEEE},
\uppercase{Shin'ichi Satoh}\authorrefmark{2}
\IEEEmembership{Member, IEEE}
}

\address[1]{Hiroshima University, Hiroshima 739-8527 Japan}
\address[2]{National Institute of Informatics, Tokyo 101-8430 Japan}

\tfootnote{
This work was supported by 
JSPS KAKENHI grant number JP16H06540
and the PRMU mentorship program.
}

\markboth
{K. Terao \headeretal: Entropy Clustering Approach for Assessing Visual Question Difficulty}
{K. Terao \headeretal: Entropy Clustering Approach for Assessing Visual Question Difficulty}

\corresp{Corresponding author:
Toru Tamaki (e-mail: tamaki@hiroshima-u.ac.jp).}

\begin{abstract}

We propose a novel approach to identify the difficulty of visual questions for Visual Question Answering (VQA) without direct supervision or annotations to the difficulty. Prior works have considered the diversity of ground-truth answers of human annotators. In contrast, we analyze the difficulty of visual questions based on the behavior of multiple different VQA models. We propose to cluster the entropy values of the predicted answer distributions obtained by three different models: a baseline method that takes as input images and questions, and two variants that take as input images only and questions only. We use a simple k-means to cluster the visual questions of the VQA v2 validation set. Then we use state-of-the-art methods to determine the accuracy and the entropy of the answer distributions for each cluster. A benefit of the proposed method is that no annotation of the difficulty is required, because the accuracy of each cluster reflects the difficulty of visual questions that belong to it. Our approach can identify clusters of difficult visual questions that are not answered correctly by state-of-the-art methods. Detailed analysis on the VQA v2 dataset reveals that 1) all methods show poor performances on the most difficult cluster (about 10\% accuracy), 2) as the cluster difficulty increases, the answers predicted by the different methods begin to differ, and 3) the values of cluster entropy are highly correlated with the cluster accuracy. We show that our approach has the advantage of being able to assess the difficulty of visual questions without ground-truth (\ie, the test set of VQA v2) by assigning them to one of the clusters. We expect that this can stimulate the development of novel directions of research and new algorithms.


\end{abstract}

\begin{keywords}
Computer vision,
Visual Question Answering,
Entropy of answer distributions
\end{keywords}

\titlepgskip=-15pt

\maketitle

\section{Introduction}

Visual Question Answering (VQA) is one of the most challenging tasks in computer vision
\cite{WU2017CVIU,VQAv1}:
given a pair of question text and image (a visual question),
a system is asked to answer the question.
It has been attracting a lot of attention in recent years
because it has a large potential to impact many applications
such as smart support for the visually impaired \cite{Gurari_2018_CVPR},
providing instructions to autonomous robots \cite{embodiedqa},
and for intelligent interaction between humans and machines \cite{visual_dialog}.
Towards these goals, many methods and datasets have been proposed.
However, while VQA models typically try to predict answers to visual questions, we take a different approach in this paper; \ie, we analyze the difficulty of visual questions.

The VQA task is particularly challenging due to the diversity of annotations.
Unlike common tasks, such as classification where precise ground truth labels
are provided by the annotators,
a visual question may have multiple different answers
annotated by different crowd workers, as shown in Figure \ref{fig:VQA_question_examples}.
In VQA v2 \cite{balanced_vqa_v2}
and VizWiz \cite{vizwiz}, which are commonly used in this task,
each visual question was annotated by 10 crowd workers,
and almost half of the visual questions in these datasets have
multiple answers \cite{Gurari2017CHI,Bhattacharya_2019_ICCV},
as shown in Table \ref{tab:VQAv2stats} for VQA v2.
Each of the visual questions in Figure \ref{fig:VQA_question_examples}
is followed by ground truth answers
and corresponding entropy values.
Entropy values are large when ground truth answers
annotated by crowd workers are diverse,
and entropy is zero when crowd workers agree to a single answer.

\Figure[t]()[width=.95\linewidth]
    {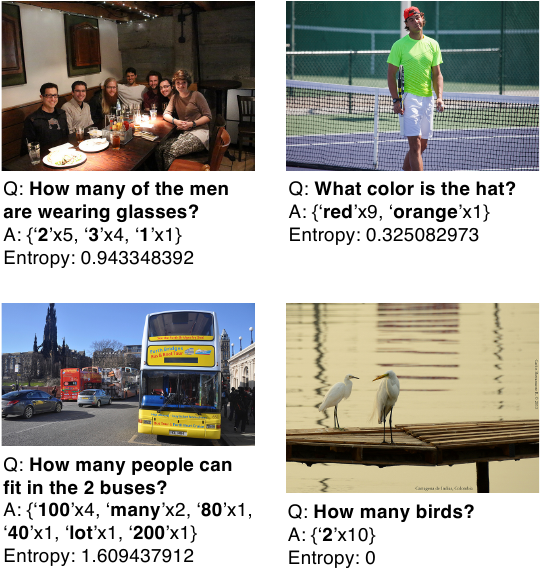}
    {Examples of visual questions and corresponding 10 answers
    of VQA v2 datasets, and
    corresponding entropy values.
    ``Q'' shows the question text, and ``A'' shows the 
    ground truth answers, where the mark ``x'' is used for indicating the number of
    crowd workers who had annotated that answer (\eg, `` `red'x9 '' signifies
    that there were nine people who had answered `red',
    `` `orange'x1 '' shows that one person answered `orange', and so on.).
    \label{fig:VQA_question_examples}
    }


The disagreement of crowd workers in ground truth annotations
has been an annoying issue for researchers dealing with tasks which involve crowdsource annotations
\cite{Daniel:2018:QCC,Soberon:2013:MCT,corr-1301-2774}.
Recently some works on VQA have tackled this issue.
Gurari \etal \cite{Gurari2017CHI} analyzed the number of
unique answers annotated by crowd workers
and proposed a model that predicts when crowdsourcing answers
(dis)agree by using binary classifiers.
Bhattacharya \etal \cite{Bhattacharya_2019_ICCV}
categorized reasons
why answers of crowd workers differ, and found
which co-occurring reasons arise frequently.

These works have revealed why multiple answers may arise
and when they disagree, 
however this is not enough to find out
how multiple answers make the visual question difficult for VQA models.
Malinowski \etal \cite{Malinowski_2015_ICCV} reported that
the disagreement harms the performance of the VQA model,
therefore the diversity of answers should be an important clue.
However, formulating the (dis)agreement as binary (single or multiple answers)
drops the information of the extent how diverse multiple answers are.
For example, suppose two different answers are given to a visual question.
This may mean that ``five people gave one answer
and the other five gave the other answer,''
or, that ``one gave one answer and the rest 9 gave the other.''
In the latter case,
the answer given by the first annotator may be noisy,
hence not suitable for taking into account.
To remove such noisy answers,
prior work \cite{Gurari2017CHI,Bhattacharya_2019_ICCV} employed
a minimum number of answer agreement.
If the agreement threshold is set to $m=2$
(at least two annotators are needed for each answer to be valid),
then the answer given by the single annotator is ignored.
However setting a threshold is ad-hoc and different threshold
may lead to different results when other datasets annotated by more (other than 10) workers 
would be available.

\begin{table}[t]
    \centering
    \caption{Numbers of unique answers per visual question
    of the validation set of VQA v2. The bottom row shows 
    averages of unique answers.}
    \label{tab:VQAv2stats}
\begin{tabular}{c@{\hspace{.3em}}|@{\hspace{.3em}}r@{\hspace{.3em}}r@{\hspace{.3em}}r@{\hspace{.3em}}r}
\#Ans      & Yes/No      & Number      & Other       & All          \\ \hline
1          & 41561       & 9775        & 18892       & 70228        \\
2          & 33164       & 6701        & 18505       & 58370        \\
3          & 5069        & 3754        & 15238       & 24061        \\
4          & 621         & 2110        & 12509       & 15240        \\
5          & 103         & 1528        & 10661       & 12292        \\
6          & 23          & 1239        & 9186        & 10448        \\
7          & 0           & 1062        & 7666        & 8728         \\
8          & 0           & 952         & 6169        & 7121         \\
9          & 0           & 726         & 4528        & 5254         \\
10         & 0           & 287         & 2325        & 2612         \\ \hline
total      & 80541       & 28134       & 105679      & 214354       \\
ave & 1.57$\pm$0.46 & 2.93$\pm$1.59 & 4.04$\pm$1.75 &2.97$\pm$1.60  \\
\end{tabular}
\end{table}

In this paper, we propose to use the entropy values of answer predictions produced by different VQA models
to evaluate the difficulty of visual questions for the models,
in contrast to prior work \cite{DBLP:conf/hcomp/YangGG18}
that uses the entropy of ground truth answers
as a metric of diversity or (dis)agreement of annotations.
In general, entropy is large when the distribution is broad,
and small when it has a narrow peak.
To the best of our knowledge, this is the first work to use entropy
for analysing the difficulty of visual questions.

The use of the entropy of answer distribution
enables us to analyse visual questions in a novel aspect.
Prior works have reported overall performance
as well as performances on three subsets of 
VQA v2 \cite{balanced_vqa_v2};
Yes/No (answers are yes or no for questions such as ``Is it ...'' and ``Does she ...''),
Numbers (answers are counts, numbers, or numeric, ``How many ...''),
and Others (other answers, ``What is ...'').
These three types have different difficulties
(\ie, Yes/No type is easier, Other type is harder),
and performances of each type are useful to highlight
how models behave to different types of visual questions.
In fact, usually the first two words carry the information of the entire question \cite{Gurari2017CHI},
and previous work \cite{Agrawal_2018_CVPR} uses this fact to switch the internal model to adopt 
suitable components to each type.
This categorization of question types is useful,
however not enough to find which visual questions are difficult.
If we can evaluate the difficulty of visual questions,
this could push forward the development of better VQA models.

Our goal is to present a novel way of analysing visual questions
by clustering the entropy values obtained from different models.
Images and questions convey different information \cite{Goyal2016,Das2016},
hence models that take images only or question only
are often used as baselines
\cite{VQAv1,Bhattacharya_2019_ICCV,balanced_vqa_v2}.
Datasets often have language bias \cite{balanced_vqa_v2,Jing_2020_AAAI,NIPS2018_7427,Agrawal_2018_CVPR}
and then questions only may be enough to answer reasonably.
However the use of the image information should
help to answer the question correctly.
Our key idea is that the entropy values of three models
(that use image only (I), question only (Q), and both (Q+I))
are useful to characterize each visual question.

\begin{figure}[t]
    \centering
    \includegraphics[width=\linewidth]{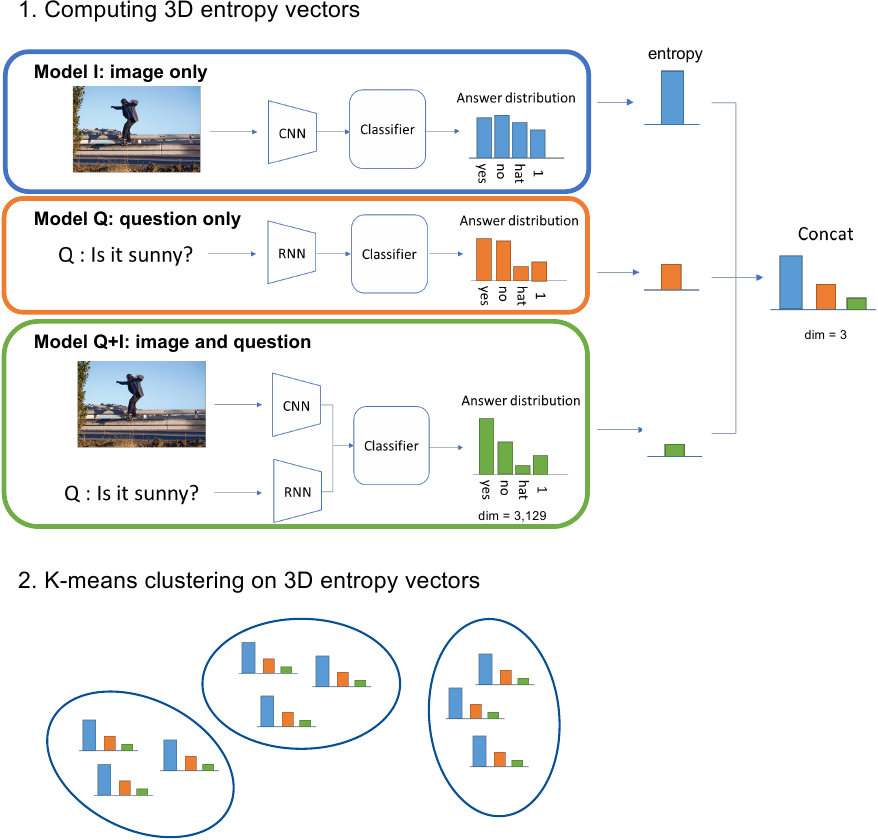}
    \caption{
    Overview of our entropy clustering approach.
    First, three models (I, Q, and Q+I) are used to predict answer distributions.
    The entropy values of the predicted answer distributions are computed
    to construct a 3D entropy vector.
    Then, clustering is performed on these 3D entropy vectors
    to analyse the accuracy for each cluster.
    }
    \label{fig:method_overview}
\end{figure}


The contributions of this work can be summarized as follows.
\begin{itemize}
    \item Instead of using the entropy of ground truth annotations,
    we use the entropy of the predicted answer distribution for the first time
    to analyse how diverse predicted answers are. 
    We show that entropy values of different models are useful to characterize visual questions.

    \item We propose an entropy clustering approach 
    to categorize the difficulty levels of visual questions
    (see Figure \ref{fig:method_overview}).
    After training three different models (I, Q, and Q+I),
    predicting answer distributions and computing entropy values,
    the visual questions are clustered.
    This is simple yet useful, and enables us to find
    which visual questions are most difficult to answer.

    \item We discuss the performances of several state-of-the-art methods.
    Our key insight is that the difficulty
    of visual question clusters is common to all methods,
    and tackling the difficult clusters may lead to
    the development of a next generation of VQA methods.

\end{itemize}




\section{Related work}

The task of VQA has attracted a lot of attention in recent years.
Challenges have been conducted since 2016,
and many datasets have been proposed.
In addition to the normal VQA task,
related tasks have emerged,
such as 
EmbodiedQA \cite{embodiedqa},
TextVQA \cite{singh2019TowardsVM},
and
VQA requiring external knowledge \cite{fvqa,Marino_2019_CVPR,Singh_2019_ICCV,Garcia_2020_AAAI}.
Still the basic framework of VQA is active and challenging,
and some tasks include VQA as an important component,
such as visual question generation \cite{mostafazadeh2016ACL_VQG,Li_2018_CVPR},
visual dialog \cite{visual_dialog,Jain_2018_CVPR}, and image captions \cite{Shen_2019_ICCV}.

VQA datasets have two types of answers.
For multiple-choice \cite{balanced_vqa_v2,zhu2016cvpr,Yu_2015_ICCV},
several candidate answers are shown to annotators for each question.
For open-ended questions \cite{VQAv1,balanced_vqa_v2,vizwiz,Johnson_2017_CVPR,COCO-QA},
annotators are asked to answer in free text,
hence answers tend to differ for many reasons \cite{Bhattacharya_2019_ICCV}.
Currently two major datasets,
VQA \cite{VQAv1,balanced_vqa_v2} and VizWiz \cite{vizwiz},
suffer from this issue
because visual questions in these datasets
were answered by 10 crowd workers,
while other datasets
\cite{COCO-QA,zhu2016cvpr,kafle2017analysis,krishnavisualgenome,fvqa,Johnson_2017_CVPR,Yu_2015_ICCV,FM-IQA}
have one answer per visual question.

This disagreement between annotators has recently been investigated in several works.
Bhattacharya \etal \cite{Bhattacharya_2019_ICCV}
proposed 9 reasons why and when answers differ:
low-quality image (LQI), answer not present (IVE),
invalid (INV), difficult (DFF),
ambiguous (AMB), subjective (SBJ),
synonyms (SYN), granular (GRN), and spam (SMP).
The first six reasons come from both/either question and/or image,
and the last three reasons are due to issues inherent to answers.
They found that ambiguity occurs the most,
and co-occurs with synonyms (same but different wordings) and 
granular (same but different concept levels).
This work gives us quite an important insight about visual questions,
however only for those that have multiple different answers annotated.
Gurari \etal \cite{Gurari2017CHI} investigated the number of
unique answers annotated by crowd workers,
but didn't consider how answers differ if disagreed.
Instead they use a threshold of agreement to show
how many annotators answered the same.
Yang \etal \cite{DBLP:conf/hcomp/YangGG18}
investigated the diversity of ground truth annotations.
They trained a model to estimate the number of unique answers
in the ground truth.
Their motivation is collecting almost all diverse ground truth answers
with a limited budget for crowdsourcing.
If more answers are expected, then they continue to ask crowdworkers to provide more answers.
If enough answers have been collected, they stop collecting answers to that question.
%
%

\begin{figure}[t]
    \centering
    \includegraphics[width=.7\linewidth]{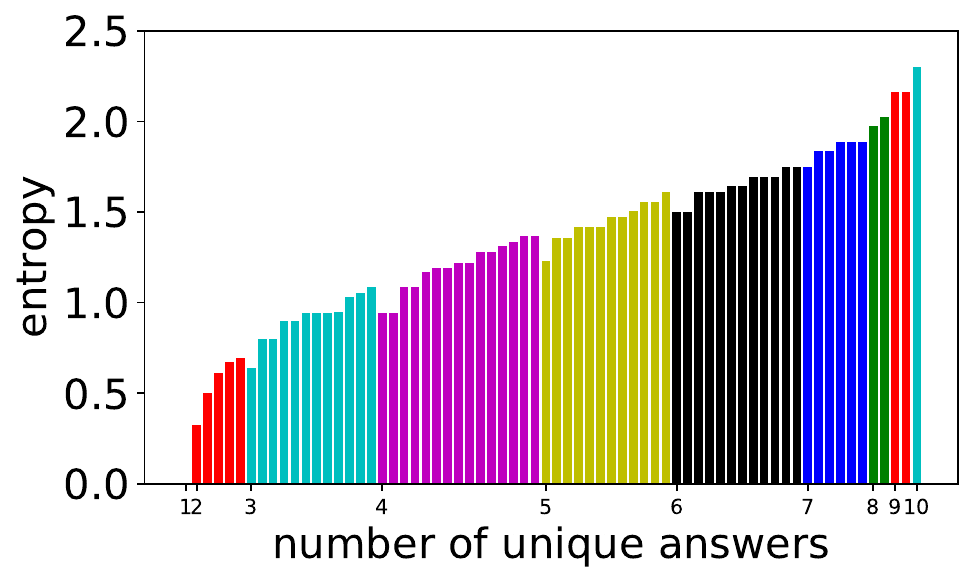}
    \caption{Entropy values of all possible combinations of unique number of answers.}
    \label{fig:entropy_count_relation}
\end{figure}

Our approach is to use the entropy of the answer distributions
of the predictions of VQA models.
This is a novel aspect, and complementary to the prior works.
Entropy takes into account by a single number
the fraction of multiple answers as well as the distribution of answers.
It therefore provides another modality to analyse visual questions
at a fine-grained level.
Figure \ref{fig:entropy_count_relation} shows
how entropy values change for the same number of unique answers.
The leftmost bar's value is zero because there is only a single answer (\ie all answers agree),
and the rightmost bar represents the case when all 10 answers are different.
In between, entropy values are sorted inside the same number of unique answers.
In the experiments we will see that the entropy of the answer distributions of VQA model predictions is consistent with the entropy of the ground truth answers, and also with the number of unique answers.

For computing the entropy,
we use three different VQA models
(image only, question only, and both)
with the expectation that images and question texts
convey different information.
This has been studied in some recent works
such as \cite{NIPS2018_7427} that
utilizes the difference between normal VQA (Q+I) and 
question-only (Q) models.
Many recent works capture the difference of visual and textual information
by using the attention mechanism between.
Some co-attention models
\cite{yu2018beyond,Nam_2017_CVPR,Xu_2016_ECCV}
use visual and textual attention in each modality or in an one-way manner (\eg, from question to image).
Some other works (such as 
DCN \cite{Nguyen_2018_CVPR},
BAN \cite{NIPS2018_7429}, and
MCAN \cite{Yu_2019_CVPR})
investigate ``dense'' co-attention that 
use bidirectional attention between images and questions.
More recent works try to capture a more complex
visual-textual information
\cite{Chen_2020_CVPR,Wang_2020_CVPR,Agarwal_2020_CVPR,Patro_2020_AAAI}.
%
Our work instead tries to keep our approach as simple as possible by using three independently trained models
to obtain the entropy.


We should note that this approach is different from uncertainty of prediction.
Teney \etal \cite{Teney_2018_CVPR}
proposed a model using soft scores because
scores may indicate uncertainty in ground truth annotations,
and minimizing the loss between ground truth and prediction answer distribution.
This approach is useful,
yet it doesn't show the nature of visual questions.

Our approach is closely related to hard example mining \cite{Wang_2015_ICCV,Shrivastava_2016_CVPR} and hardness / failure prediction \cite{Wang_2018_ECCV}. Hard example mining approaches determine which examples are difficult to train during training, while hardness prediction jointly trains the task classifier and an auxiliary hardness prediction network. Compared to these works, our approach  differs in the following two aspects. First, the VQA task is multi-modal and assessing the difficulty of visual questions has not been considered before. Second, our approach is off-line and can determine the difficulty without ground-truth, \ie, before actually trying to answer the visual questions in the test set.

\section{Clustering visual questions with entropy}






Here we formally define the entropy.
Let $A$ be the set of possible answers that a VQA model would predict,
and $P(x=a)$ or $P(a)$ the probability distribution
predicted by the VQA model,
which satisfies $\sum_{a \in A} P(a) = 1$ and $\forall a \in A, P(a) \ge 0$.
The entropy $H$ of the VQA model prediction is defined by
\begin{equation}
H = -\sum_{a \in A} P(a) \ln P(a).
\end{equation}

\subsection{Clustering method}


To perform clustering, we hypothesize that ``easy visual questions
lead to low entropy while difficult visual questions to high entropy.''
A similar concept has been reported in terms of the human consensus with multiple ground truth annotations \cite{Malinowski_2015_ICCV},
but in this paper we address the relation between the difficulty and the entropy of answer distributions produced by VQA models.
This is reasonable because for easy visual questions 
VQA systems can predict answer distributions
in which the correct answer category has large probability while other categories are low.
In contrast, difficult visual questions makes VQA systems generate
broad answer distributions because many answer candidates may be equally plausible.
Entropy can capture the diversity of
predicted answer distributions,
and also that of ground truth annotations in the same manner.

We prepare three different models
that use as input image only (I), question only (Q), and both question and image (Q+I).
In this case, we expect the following three levels of difficulty of visual questions:
\begin{itemize}
\item Level 1: Reasonably answered by using question only.
\item Level 2: Difficult to answer with question only but good with images.
\item Level 3: Difficult even if both image and question are provided.
\end{itemize}
For a certain visual question,
it is of level 1 if the answer distribution of the Q model has low entropy.
It is of level 2 if the Q model is high entropy and the Q+I model is low entropy.
If both the Q and Q+I models have high entropy, then the visual question is of level 3.
(We show a procedure to determine these levels in
section \ref{Clustering results}.)
This concept is realised by the following procedure.
1) Train the I, Q, and Q+I models on the training set with image only, questions only, and both images and questions, respectively.
2) Evaluate the validation set by using the three models and 
    compute answer distributions and entropy values of each of visual questions.
3) Perform clustering on the validation set with entropy values.
Clustering features are the entropy values of the three models.

\subsection{Datasets and setting}

We use VQA v2 \cite{balanced_vqa_v2}:
it consists of training, validation, and test sets.
To train models, we use the training set
(82,783 images, 443,757 questions, and 4,437,570 answers).
We use the validation set
(40,504 images, 214,354 questions, and 2,143,540 answers)
for clustering and analysis.

We choose Pythia v0.1 \cite{pythia2018arxiv,singh2018pythia}
as a base model, 
and modify it so that
it takes questions only (Q model),
or images only (I model).
To do so, we simply set either
image features or question features to zero vectors.
With no modification, it is Q+I model (\ie Pythia v0.1).
As in prior works \cite{Anderson_2018_CVPR,lu2019vilbert,Teney_2018_CVPR,Yu_2019_CVPR},
3129 answers
in the training set
that occur at least 8 times
are chosen as candidates,
which results in 
a multi-class problem predicting answer distributions of 3129 dimension.
Note that other common choices for the number of answers
are 3000 \cite{yu2018beyond,fukui-etal-2016-multimodal}
and 1000 \cite{NIPS2016_6202}.
Even when different numbers are used, our entropy clustering approach works and we expect our findings to hold.

To compare the performance with state-of-the-art methods,
we use
BUTD \cite{Anderson_2018_CVPR},
MBF \cite{Yu_2017_ICCV},
MFH \cite{yu2018beyond},
BAN \cite{NIPS2018_7429} (including BAN-4 and BAN-8),
MCAN \cite{Yu_2019_CVPR} (including small and large),
and
Pythia v0.3 \cite{singh2018pythia,singh2019TowardsVM}.

The metric for performance evaluation
is the following, which is commonly used for this dataset \cite{VQAv1}:
\begin{equation}
\text{accuracy} = 
\min \left( \frac{\text{\parbox{3.5cm}{\centering \# humans that provided that answer}}}{3}, 1 \right),
\label{eq:accuracy_definition}
\end{equation}
in other words, an answer is 100\% correct
if at least three annotated answers match that answer.


First we show the performance of each model
in Table \ref{tab:vqa_accuracy}. 
As expected, the I model performs worst
because there is no clue of questions in the image.
In contrast, Q model performs reasonably better,
particularly for Yes/No type.
Average performances of different models (excluding I and Q)
are about 84\%, 47\%, and 58\%
for types of Yes/No, Number, and Other, respectively.
%
%
Note that we show average and standard deviations (std) in Table \ref{tab:vqa_accuracy},
and the std values look relatively large.
This is natural and it is due to the definition of VQA accuracy (Eq. \eqref{eq:accuracy_definition}).
For each prediction, accuracy is discrete: 0, 33.3, 66.6, or 100 depending on
how many people provided that ground truth answer.
Averaging these discrete values results 
in a large std.
(In other words, large discretization errors lead to a large std.
For example, 10 predictions with accuracy of 100
and 35 with accuracy of 0 result in $22.22 \pm 41.57$.)
It is quite common for VQA papers to 
report average accuracy only without std,
probably because std is large for any models and not useful for comparison.
In this paper we report std of accuracy
as well as std for entropy and the reasons to differ.

\begin{table}[t]
	\centering
	\caption{Accuracy of models on the validation set of VQA v2.}
	\label{tab:vqa_accuracy}
\scalebox{0.9}{
	\begin{tabular}{l|c@{\ \ }c@{\ \ }c@{\ \ }c}
Model & Overall & Yes/No & Number & Other \\
\hline
 I   & 24.65$\pm$41.42 & 64.21$\pm$44.51 & 0.27$\pm$3.16 & 0.99$\pm$7.48 \\
 Q   & 44.83$\pm$46.76 & 68.48$\pm$43.00 & 32.05$\pm$43.45 & 30.21$\pm$42.91 \\
 Q+I \cite{pythia2018arxiv}& 67.47$\pm$43.35 & 84.52$\pm$33.01 & 47.55$\pm$46.25 & \textbf{59.78}$\pm$45.00 \\
\hline
BUTD \cite{Anderson_2018_CVPR} & 63.79$\pm$44.61 & 81.20$\pm$35.84 & 43.90$\pm$45.93 & 55.81$\pm$45.78 \\
MFB \cite{Yu_2017_ICCV}        & 65.14$\pm$44.21 & 83.11$\pm$34.31 & 45.32$\pm$46.16 & 56.72$\pm$45.60 \\
MFH \cite{yu2018beyond}        & 66.23$\pm$43.85 & 84.12$\pm$33.45 & 46.71$\pm$46.27 & 57.79$\pm$45.40 \\
BAN-4 \cite{NIPS2018_7429}     & 65.87$\pm$43.90 & 83.57$\pm$33.88 & 47.23$\pm$46.17 & 57.34$\pm$45.43 \\
BAN-8 \cite{NIPS2018_7429}     & 66.00$\pm$43.87 & 83.48$\pm$33.95 & 47.20$\pm$46.17 & 57.69$\pm$45.40 \\
MCAN-small \cite{Yu_2019_CVPR} & 67.20$\pm$43.42 & 84.91$\pm$32.68 & \textbf{49.35}$\pm$46.23 & 58.46$\pm$45.18 \\
MCAN-large \cite{Yu_2019_CVPR} & \textbf{67.47}$\pm$43.33 & \textbf{85.33}$\pm$32.24 & 48.96$\pm$46.23 & 58.78$\pm$45.13 \\
Pythia v0.3 \cite{singh2019TowardsVM} & 65.91$\pm$44.42 & 84.30$\pm$33.56 & 44.90$\pm$46.47 & 57.49$\pm$46.07 \\
	\end{tabular}
}
\end{table}

Next 
in Table \ref{tab:vqa_entropy}
we show 
the entropy values of the predicted answer distributions
by different models for each of the three types, as well as ground truth
annotations.
Average entropy values of models (excluding I and Q) for each type are
0.25, 1.62, 1.72, respectively.
Yes/No type has smaller entropy
than the others because answer distributions
tend to gather around only two candidates (``Yes'' and ``No'').
Note that the range of entropy values is different
for model predictions and for the ground truth answers.
Entropy ranges from 
$0$ (single answer) to $2.303$ (10 different answers)
for ground truth answers,
and 
from $0$ ($1$ for a single entry, otherwise $0$)
to $8.048$ (uniform values of $1/3129$) for model predictions.

\begin{table}[t]
    \centering
    \caption{Entropy of models on the validation set of VQA v2.}
    \label{tab:vqa_entropy}

    \begin{tabular}{l|c@{\ \ }c@{\ \ }c@{\ \ }c}
Model & Overall & Yes/No & Number & Other \\
\hline
 I & 4.19$\pm$0.42 & 4.16$\pm$0.42 & 4.19$\pm$0.43 & 4.21$\pm$0.41 \\
 Q & 1.80$\pm$1.32 & 0.59$\pm$0.22 & 2.28$\pm$0.94 & 2.60$\pm$1.22 \\
 Q+I \cite{pythia2018arxiv} & 0.84$\pm$1.06 & 0.20$\pm$0.27 & 1.39$\pm$1.18 & 1.19$\pm$1.15 \\
\hline
BUTD \cite{Anderson_2018_CVPR} & 1.24$\pm$1.33 & 0.32$\pm$0.29 & 1.86$\pm$1.25 & 1.77$\pm$1.45 \\
MFB \cite{Yu_2017_ICCV}        & 1.76$\pm$1.86 & 0.42$\pm$0.31 & 2.07$\pm$1.77 & 2.71$\pm$1.95 \\
MFH \cite{yu2018beyond}        & 1.63$\pm$1.77 & 0.40$\pm$0.31 & 2.00$\pm$1.76 & 2.46$\pm$1.89 \\
BAN-4 \cite{NIPS2018_7429}     & 0.99$\pm$1.20 & 0.21$\pm$0.27 & 1.60$\pm$1.25 & 1.43$\pm$1.31 \\
BAN-8 \cite{NIPS2018_7429}     & 0.95$\pm$1.17 & 0.20$\pm$0.26 & 1.53$\pm$1.23 & 1.36$\pm$1.27 \\
MCAN-small \cite{Yu_2019_CVPR} & 1.21$\pm$1.71 & 0.17$\pm$0.27 & 1.66$\pm$1.82 & 1.89$\pm$1.91 \\
MCAN-large \cite{Yu_2019_CVPR} & 1.15$\pm$1.64 & 0.16$\pm$0.26 & 1.63$\pm$1.76 & 1.78$\pm$1.84 \\
Pythia v0.3 \cite{singh2019TowardsVM} & 0.59$\pm$0.82 & 0.13$\pm$0.22 & 0.86$\pm$0.93 & 0.89$\pm$0.87 \\
\hline
GT & 0.67$\pm$0.68  & 0.25$\pm$0.29 & 0.66$\pm$0.68 & 0.99$\pm$0.71  \\
    \end{tabular}

\end{table}

\begin{table*}[t]

 \centering
 \caption{Clustering results for the validation set of VQA v2.
 Each column corresponds to a different cluster
 and colors indicate cluster types
 (level 1 in gray, level 2 in yellow, and level 3 in red).}
 \label{tab:clustering_result}

\definecolor{light_gray}{rgb}{0.9,0.9,0.9}
\definecolor{light_red}{rgb}{1,0.8,0.8}
\definecolor{light_yellow}{rgb}{1,0.98,0.9}

\newcolumntype{g}{>{\columncolor{light_gray}}r}
\newcolumntype{d}{>{\columncolor{light_red}}r}
\newcolumntype{y}{>{\columncolor{light_yellow}}r}

\scalebox{0.75}{
 \begin{tabular}{cl|ggyyyyyddd}
 & cluster&0&1&2&3&4&5&6&7&8&9 \\
\hline
\multirow{3}{*}{\rotatebox{90}{\parbox[b]{3em}{\centering base\\ model\\ entropy}}}
 & I&3.77$\pm$0.27&4.47$\pm$0.24&4.22$\pm$0.39&4.09$\pm$0.40&4.22$\pm$0.40&4.19$\pm$0.40&4.23$\pm$0.40&4.27$\pm$0.39&4.23$\pm$0.41&4.42$\pm$0.41 \\
 & Q&0.60$\pm$0.24&0.61$\pm$0.23&2.69$\pm$0.32&1.78$\pm$0.29&4.09$\pm$0.47&1.48$\pm$0.41&2.61$\pm$0.36&4.01$\pm$0.49&2.68$\pm$0.49&4.33$\pm$0.62 \\
 & Q+I\cite{pythia2018arxiv}&0.20$\pm$0.26&0.21$\pm$0.27&0.24$\pm$0.27&0.25$\pm$0.28&0.63$\pm$0.46&1.45$\pm$0.45&1.46$\pm$0.35&2.25$\pm$0.43&2.77$\pm$0.47&3.79$\pm$0.56 \\
\cline{2-12}
\multirow{8}{*}{\rotatebox{90}{\parbox[b]{8em}{\centering state-of-the-art\\ entropy}}}
 & BUTD\cite{Anderson_2018_CVPR}&0.38$\pm$0.45&0.42$\pm$0.48&0.85$\pm$0.98&0.77$\pm$0.83&1.73$\pm$1.35&1.79$\pm$0.92&2.03$\pm$0.97&3.13$\pm$1.06&3.01$\pm$0.90&3.98$\pm$0.91 \\
 & MFB\cite{Yu_2017_ICCV}&0.55$\pm$0.67&0.57$\pm$0.68&1.41$\pm$1.41&1.15$\pm$1.16&2.77$\pm$1.80&2.39$\pm$1.43&2.85$\pm$1.48&4.44$\pm$1.36&4.03$\pm$1.36&5.47$\pm$1.27 \\
 & MFH\cite{yu2018beyond}&0.49$\pm$0.56&0.51$\pm$0.57&1.23$\pm$1.27&1.00$\pm$1.03&2.47$\pm$1.75&2.19$\pm$1.30&2.60$\pm$1.37&4.21$\pm$1.42&3.83$\pm$1.33&5.37$\pm$1.33 \\
 & BAN-4\cite{NIPS2018_7429}&0.25$\pm$0.37&0.27$\pm$0.39&0.61$\pm$0.81&0.53$\pm$0.67&1.32$\pm$1.21&1.46$\pm$0.83&1.68$\pm$0.89&2.67$\pm$1.07&2.63$\pm$0.89&3.61$\pm$0.99 \\
 & BAN-8\cite{NIPS2018_7429}&0.23$\pm$0.36&0.26$\pm$0.37&0.57$\pm$0.77&0.50$\pm$0.64&1.21$\pm$1.16&1.40$\pm$0.82&1.60$\pm$0.87&2.55$\pm$1.07&2.54$\pm$0.90&3.48$\pm$1.01 \\
 & MCAN-small\cite{Yu_2019_CVPR}&0.23$\pm$0.45&0.25$\pm$0.49&0.70$\pm$1.16&0.54$\pm$0.89&1.74$\pm$1.78&1.66$\pm$1.3&2.0$\pm$1.45&3.66$\pm$1.67&3.35$\pm$1.46&4.95$\pm$1.52 \\
 & MCAN-large\cite{Yu_2019_CVPR}&0.21$\pm$0.42&0.23$\pm$0.45&0.64$\pm$1.07&0.51$\pm$0.84&1.62$\pm$1.71&1.59$\pm$1.23&1.9$\pm$1.37&3.5$\pm$1.65&3.21$\pm$1.43&4.82$\pm$1.53 \\
 & Pythia v0.3\cite{singh2019TowardsVM}&0.14$\pm$0.28&0.16$\pm$0.29&0.28$\pm$0.50&0.25$\pm$0.44&0.69$\pm$0.78&0.89$\pm$0.68&0.99$\pm$0.69&1.59$\pm$0.85&1.76$\pm$0.80&2.20$\pm$0.90 \\
\cline{2-12}
\multirow{3}{*}{\rotatebox{90}{\parbox[b]{3em}{test set\\ entropy}}}
 & I &3.77$\pm$0.26&4.47$\pm$0.24&4.21$\pm$0.39&4.08$\pm$0.40&4.22$\pm$0.41&4.19$\pm$0.41&4.23$\pm$0.40&4.28$\pm$0.40&4.24$\pm$0.41&4.43$\pm$0.41 \\
 & Q &0.60$\pm$0.23&0.61$\pm$0.23&2.70$\pm$0.32&1.78$\pm$0.29&4.09$\pm$0.47&1.48$\pm$0.41&2.62$\pm$0.37&4.02$\pm$0.49&2.68$\pm$0.50&4.33$\pm$0.61 \\
 & Q+I\cite{pythia2018arxiv} &0.18$\pm$0.25&0.20$\pm$0.26&0.24$\pm$0.27&0.26$\pm$0.29&0.63$\pm$0.45&1.45$\pm$0.45&1.46$\pm$0.35&2.25$\pm$0.44&2.78$\pm$0.49&3.78$\pm$0.56 \\
\hline
\hline
\multirow{3}{*}{\rotatebox{90}{\parbox[b]{3em}{\centering base\\ model\\ acc.}}}
 & I&53.13$\pm$47.09&54.81$\pm$46.93&0.67$\pm$6.88&1.61$\pm$11.58&0.75$\pm$6.89&2.33$\pm$13.20&0.69$\pm$5.87&1.05$\pm$7.18&0.90$\pm$6.28&1.13$\pm$7.00 \\
 & Q&69.70$\pm$42.54&66.73$\pm$43.73&32.91$\pm$45.13&46.91$\pm$47.63&16.32$\pm$35.09&34.99$\pm$42.62&24.13$\pm$37.8&8.71$\pm$23.97&14.54$\pm$29.74&5.50$\pm$18.76 \\
 & Q+I\cite{pythia2018arxiv}&86.15$\pm$31.32&82.87$\pm$34.48&\textbf{84.53}$\pm$32.9&\textbf{83.08}$\pm$34.53&\textbf{67.79}$\pm$42.20&47.10$\pm$44.24&45.73$\pm$43.64&26.04$\pm$38.10&22.53$\pm$34.56&9.32$\pm$25.05 \\
\cline{2-12}
 \multirow{8}{*}{\rotatebox{90}{\parbox[b]{8em}{\centering state-of-the-art\\ accuracy}}}
 & BUTD\cite{Anderson_2018_CVPR}&82.55$\pm$34.81&78.85$\pm$37.62&78.32$\pm$38.36&77.39$\pm$39.08&60.18$\pm$44.87&45.77$\pm$44.21&43.06$\pm$43.62&23.53$\pm$36.76&22.6$\pm$35.03&9.93$\pm$25.37 \\
 & MFB\cite{Yu_2017_ICCV}&84.20$\pm$33.29&80.84$\pm$36.24&79.77$\pm$37.26&78.73$\pm$38.16&60.93$\pm$44.76&47.09$\pm$44.38&43.56$\pm$43.56&24.25$\pm$37.11&23.16$\pm$35.32&10.39$\pm$25.85 \\
 & MFH\cite{yu2018beyond}&85.38$\pm$32.21&81.89$\pm$35.41&80.72$\pm$36.51&80.06$\pm$37.09&62.84$\pm$44.23&47.97$\pm$44.36&44.82$\pm$43.69&25.17$\pm$37.79&23.77$\pm$35.47&10.97$\pm$26.82 \\
 & BAN-4\cite{NIPS2018_7429}&84.89$\pm$32.65&81.35$\pm$35.75&80.11$\pm$36.95&79.70$\pm$37.32&61.98$\pm$44.48&47.97$\pm$44.20&45.27$\pm$43.72&24.67$\pm$37.24&23.84$\pm$35.47&10.76$\pm$26.41 \\
 & BAN-8\cite{NIPS2018_7429}&84.88$\pm$32.64&81.25$\pm$35.85&80.51$\pm$36.67&79.67$\pm$37.37&62.79$\pm$44.23&48.33$\pm$44.24&45.65$\pm$43.74&25.07$\pm$37.62&24.00$\pm$35.59&10.62$\pm$26.21 \\
 & MCAN-small\cite{Yu_2019_CVPR}&86.06$\pm$31.49&82.76$\pm$34.63&81.31$\pm$35.97&80.78$\pm$36.50&63.52$\pm$43.96&\textbf{49.86}$\pm$44.24&46.67$\pm$43.72&26.10$\pm$38.02&25.46$\pm$36.33&11.56$\pm$27.42 \\
 & MCAN-large\cite{Yu_2019_CVPR}&\textbf{86.44}$\pm$31.10&\textbf{83.18}$\pm$34.25&81.58$\pm$35.79&80.86$\pm$36.53&63.79$\pm$43.88&49.33$\pm$44.30&\textbf{47.27}$\pm$43.78&26.35$\pm$38.05&25.50$\pm$36.27&11.58$\pm$27.30 \\
 & Pythia v0.3\cite{singh2019TowardsVM}&85.56$\pm$32.46&82.29$\pm$35.46&82.35$\pm$36.42&80.19$\pm$37.90&65.21$\pm$45.60&48.54$\pm$46.50&47.13$\pm$46.57&\textbf{27.31}$\pm$42.05&\textbf{26.32}$\pm$40.98&\textbf{11.70}$\pm$30.28 \\
\hline
\hline
\multirow{8}{*}{\rotatebox{90}{\parbox[b]{8em}{\centering GT statistics}}}
 & entropy &0.30$\pm$0.36&0.29$\pm$0.36&0.60$\pm$0.60&0.50$\pm$0.54&0.99$\pm$0.66&0.97$\pm$0.65&1.13$\pm$0.65&1.37$\pm$0.67&1.45$\pm$0.65&1.34$\pm$0.70 \\
 \cline{3-12}
 & ave \# ans&1.72$\pm$0.98&1.69$\pm$0.96&2.71$\pm$1.91&2.39$\pm$1.67&3.98$\pm$2.30&3.84$\pm$2.26&4.43$\pm$2.37&5.42$\pm$2.57&5.75$\pm$2.53&5.39$\pm$2.68 \\
\cline{3-12}
& total &42637&52600&20235&21643&10631&13516&19010&12608&12620&8854 \\
\cline{2-12}
 & Yes/No&35483&44426&22&262&6&288&23&13&13&5 \\
 & Number&1194&1471&2770&5778&253&4489&4969&1255&3790&2165 \\
 & Other&5960&6703&17443&15603&10372&8739&14018&11340&8817&6684 \\
\cline{2-12}
 & \# agree &20762&26338&6770&8528&1488&1912&1988&954&699&789 \\
 & \# disagree&21875&26262&13465&13115&9143&11604&17022&11654&11921&8065 \\
 \hline
 \hline
\multirow{9}{*}{\rotatebox{90}{reasons to differ \cite{Bhattacharya_2019_ICCV}}}
 & LQI&0.05$\pm$0.04&0.05$\pm$0.04&0.02$\pm$0.03&0.03$\pm$0.04&0.01$\pm$0.03&0.03$\pm$0.04&0.03$\pm$0.04&0.02$\pm$0.04&0.03$\pm$0.04&0.06$\pm$0.08 \\
 & IVE&0.48$\pm$0.21&0.48$\pm$0.21&0.11$\pm$0.14&0.18$\pm$0.19&0.10$\pm$0.10&0.23$\pm$0.20&0.19$\pm$0.19&0.15$\pm$0.15&0.24$\pm$0.20&0.21$\pm$0.16 \\
 & INV&0.26$\pm$0.14&0.25$\pm$0.14&0.02$\pm$0.03&0.03$\pm$0.05&0.01$\pm$0.02&0.04$\pm$0.06&0.03$\pm$0.04&0.02$\pm$0.02&0.03$\pm$0.04&0.03$\pm$0.03 \\
 & DFF&0.09$\pm$0.07&0.08$\pm$0.07&0.08$\pm$0.10&0.11$\pm$0.11&0.07$\pm$0.07&0.15$\pm$0.12&0.13$\pm$0.12&0.10$\pm$0.09&0.17$\pm$0.14&0.12$\pm$0.08 \\
 & AMB&0.75$\pm$0.13&0.76$\pm$0.12&0.95$\pm$0.05&0.93$\pm$0.07&0.96$\pm$0.04&0.91$\pm$0.08&0.93$\pm$0.06&0.94$\pm$0.06&0.92$\pm$0.07&0.91$\pm$0.08 \\
 & SBJ&0.32$\pm$0.23&0.30$\pm$0.23&0.13$\pm$0.09&0.14$\pm$0.11&0.12$\pm$0.08&0.13$\pm$0.12&0.12$\pm$0.09&0.11$\pm$0.09&0.11$\pm$0.09&0.10$\pm$0.09 \\
 & SYN&0.25$\pm$0.27&0.25$\pm$0.26&0.81$\pm$0.18&0.72$\pm$0.25&0.87$\pm$0.12&0.65$\pm$0.27&0.72$\pm$0.24&0.80$\pm$0.18&0.68$\pm$0.24&0.72$\pm$0.21 \\
 & GRN&0.35$\pm$0.22&0.35$\pm$0.22&0.79$\pm$0.15&0.71$\pm$0.20&0.82$\pm$0.11&0.66$\pm$0.21&0.71$\pm$0.19&0.76$\pm$0.16&0.67$\pm$0.19&0.69$\pm$0.17 \\
 & SPM&0.03$\pm$0.02&0.02$\pm$0.01&0.01$\pm$0.01&0.01$\pm$0.01&0.01$\pm$0.01&0.01$\pm$0.01&0.01$\pm$0.01&0.01$\pm$0.01&0.01$\pm$0.01&0.02$\pm$0.02 \\
 & OTH&0.01$\pm$0.01&0.01$\pm$0.01&0.00$\pm$0.00&0.00$\pm$0.00&0.00$\pm$0.00&0.00$\pm$0.00&0.00$\pm$0.00&0.00$\pm$0.00&0.00$\pm$0.00&0.01$\pm$0.01 \\
\hline
\end{tabular}
  }
  
\end{table*}

\subsection{Clustering results}
\label{Clustering results}

Now we show the clustering results in 
Table \ref{tab:clustering_result}.
We used $k$-means to cluster the 3-d vectors of 214,354  visual questions into $k=10$ clusters.
Note that many factors (\eg
initialization and number of clusters,
chosen algorithms)
affect the clustering result, but we will show that
similar clustering results are obtained with different parameter settings
in experiments.
Here we use the simplest algorithm, and a reasonable number of clusters.


Each column of Table \ref{tab:clustering_result} shows the
statistics for each cluster.
Clusters are numbered in ascending order of the entropy for the Q+I model.
The top rows with `\textbf{base model entropy}' show the entropy values
for the three base models.

To find three levels of visual questions,
we divide the clusters by the following simple rule.
For each cluster,
\begin{itemize}
\item if $\text{`Q entropy'} < 1$ then it is level 1,
\item else 
\begin{itemize}
\item if $\text{`Q+I entropy'} > 2$ then it is level 3,
\item otherwise level 2.
\end{itemize}
\end{itemize}
Column colors of Table \ref{tab:clustering_result} indicate levels;
level 1 (clusters 0 and 1) are in gray,
level 2 in yellow (2 to 6), and level 3 (7, 8, and 9) in red.


Below we describe other rows of Table \ref{tab:clustering_result}.
\begin{itemize}

\item \textbf{base model acc.}\ 
Accuracy values of the three base models.
Accuracy of Q+I model tends to decrease as Q+I entropy increases,
which we will discuss later.

\item \textbf{state-of-the-art entropy and accuracy}\ 
Entropy and accuracy values of 9 state-of-the-art methods.

\item \textbf{test set entropy}\ 
Entropy values of the test set of VQA v2.
We assign test visual questions to one of these clusters
(we will discuss this later).

\item \textbf{GT statistics}\
Statistics of ground truth annotations.
Row `entropy' shows entropy values of ground truth annotations.
Row `ave \# ans' shows the average number of unique answers per visual question.
These two rows show how ground truth answers differ in each cluster.

Row `total' shows total numbers of visual questions.
Rows `yes/no', `number', and `other' shows
numbers of each type in that cluster.
Rows `\# agree' and `\# disagree'
show numbers of visual questions for which 10 answers agree (all are the same)
and disagree (all are not the same), as in \cite{Bhattacharya_2019_ICCV}.

\item \textbf{reasons to differ}\ 
Average values obtained by reason classifiers \cite{Bhattacharya_2019_ICCV}
that output values 
from 0 (not that reason) to 1 (it is this reason) to each reason independently.
We train classifiers on the subset of the VQA v2 training set
provided by \cite{Bhattacharya_2019_ICCV},
then apply to VQA v2 validation set.

\end{itemize}

\subsection{Discussion}


\subsubsection{Entropy suggests accuracy.}
\begin{figure}
    \centering
    \includegraphics[width=\linewidth]{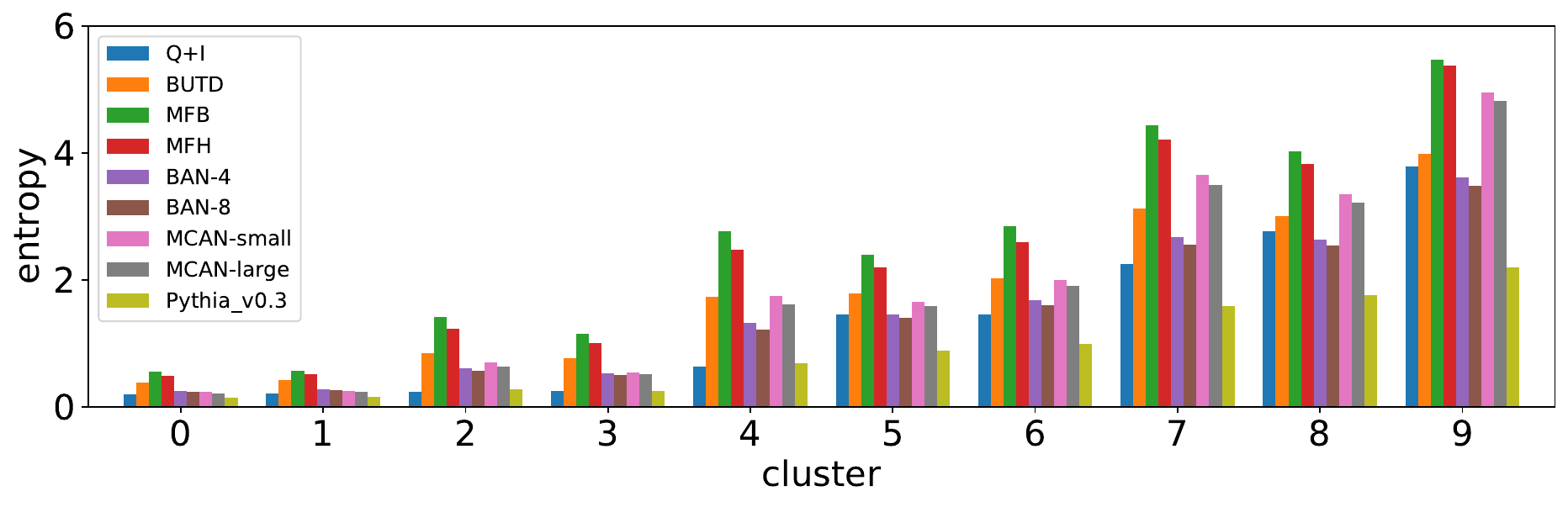}
    \includegraphics[width=\linewidth]{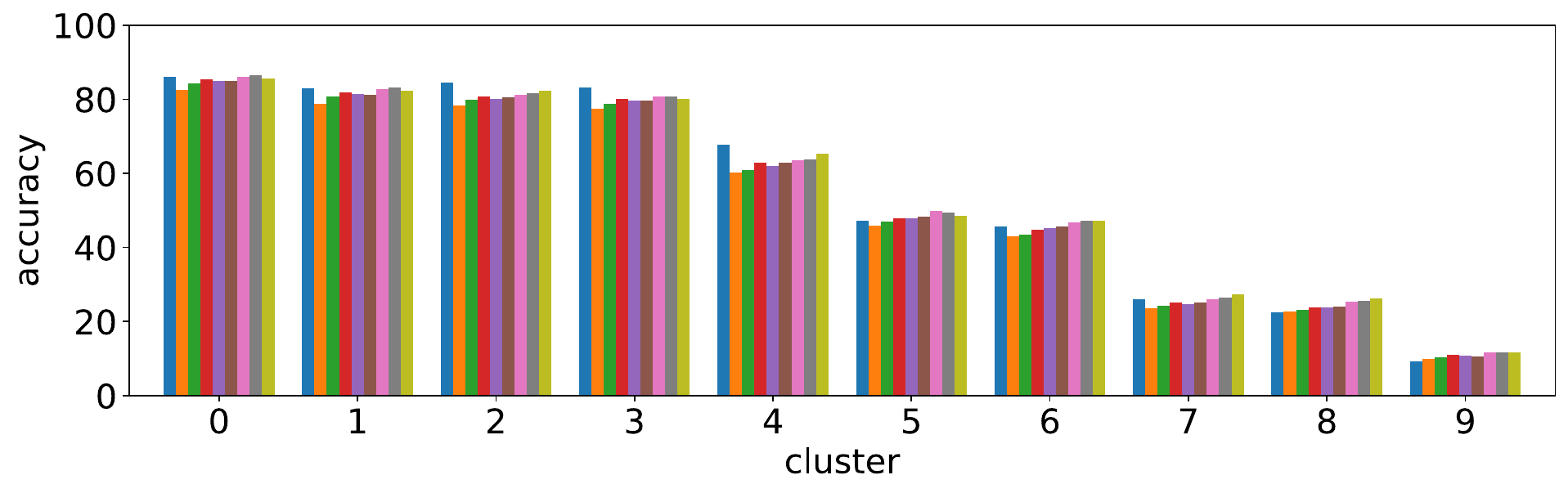}
    \includegraphics[width=\linewidth]{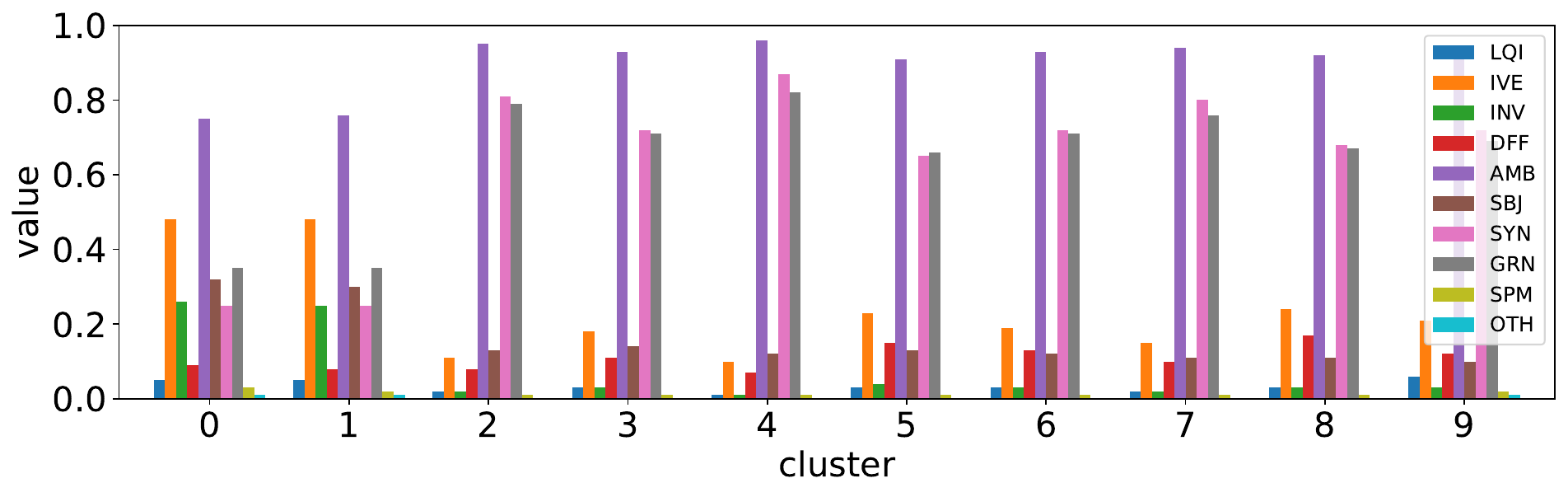}
    
    \caption{Values of (top) entropy, (middle) accuracy, and (bottom) reasons to differ
    for each cluster.
    Entropy values increase while accuracy decreases
    from cluster 0 (left) to 9 (right), while the
    predicted values of 9 reasons to differ are not well correlated to the order of clusters.
    Note that the legend is common for all
    following plots of entropy and accuracy.
    }
    \label{fig:entropy_accuracy_relation}
\end{figure}

\begin{figure}
    \centering
    \includegraphics[width=\linewidth]{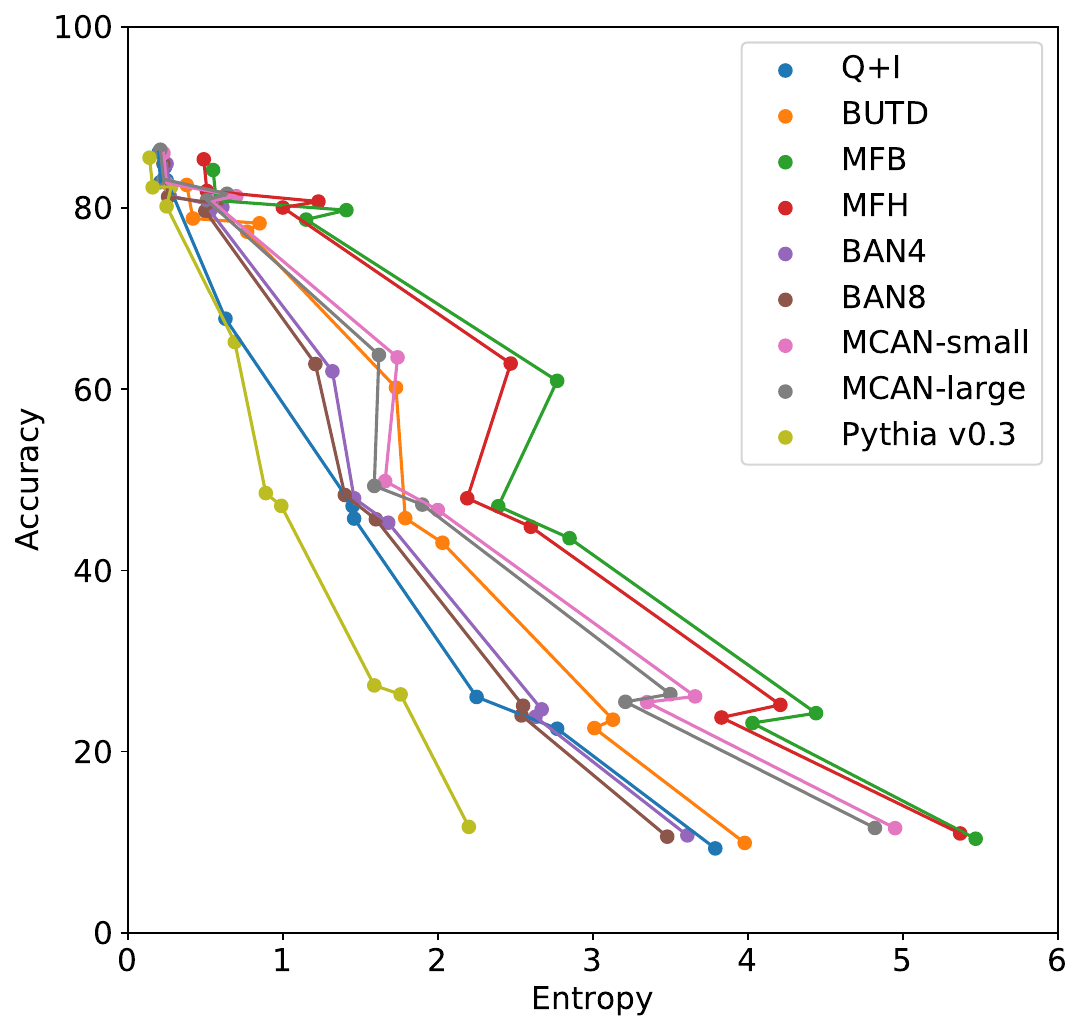}
    \caption{Scatter plot of averages of entropy and accuracy of different models. 
    Dots of each model are connected by lines in the order of cluster from top-left (cluster 0) to bottom-right (cluster 9).}
    \label{fig:entropy_accuracy_relation2}
\end{figure}

We performed the clustering by using the entropy values of the
three models based on Pythia v0.1 \cite{pythia2018arxiv,singh2018pythia}.
Using a different base model may lead to different clustering results,
however the values of entropy and accuracy of different state-of-the-art models
exhibit similar trends; 
entropy values increase while accuracy decreases
from cluster 0 to 9,
as shown in Figure \ref{fig:entropy_accuracy_relation}.
This suggests that clusters with large (or small) entropy values have low (high) accuracy,
as shown in Figure \ref{fig:entropy_accuracy_relation2},
and this tells us that entropy values are an important cue for predicting accuracy.

\subsubsection{Entropy is different from reasons to differ and question types.}
Most frequent reasons to differ
shown in \cite{Bhattacharya_2019_ICCV} are AMB, SYN, and GRN,
but Figure \ref{fig:entropy_accuracy_relation} shows that
predicted values of those reasons are not well correlated to the order of clusters.
For question types, Number and Other types looks not related to these clusters.
Therefore our approach using entropy captures different aspects of visual questions.

\subsubsection{Cluster 0 is easy, cluster 9 is hard.}
Level 1 (clusters 0 and 1) are dominant,
and covers 44\% of the entire validation set, including 99\% of Yes/No type.
Low entropy values and few number of unique answers (row `ave \# ans')
of these cluster can be explained by the fact that
typical answers are either `Yes' or `No'.
Accuracy of Yes/No type is expected to be about 85\% (Table \ref{tab:vqa_accuracy}),
and it is close to the accuracy for these clusters.
In contrast, level 3 (clusters 7, 8, and 9) looks much more difficult to answer.
In particular, accuracy values of cluster 9 are about 10\%
compared to over 80\% of level 1.
This is due to the fact that visual questions with disagreed answers gather in this level;
GT entropy is about 1.3, with more than five unique answers.
However, values of DFF, AMB, SYN and GRN of level 3 are not so different from level 2,
which may suggest that the quality of visual questions is not the main reason for difficulty.

\subsubsection{Difficulty of the test set can be predicted.}
This finding enables us to evaluate the difficulty of visual questions
in \emph{the test set}.
To see this,
we applied the same base models (that are already trained and used for clustering)
to visual questions in the test set,
and computed entropy values to assign each visual questions
to one of the 9 clusters.
Rows with `test set entropy' in Table \ref{tab:clustering_result}
show the average entropy values of those test set visual questions.
Assuming that the validation and test sets are similar in nature,
we now are able to \emph{evaluate and predict the difficulty
of test-set visual questions without computing accuracy}.
This is the most interesting result,
and we have released a list \cite{vqd}\footnote{
Clustering results are available online at \url{https://github.com/tttamaki/vqd},
in which we show lists of pairs of question IDs and clusters
for both the validation and test sets
of the VQA v2 dataset.
}
that shows
which visual questions in the train / val / test sets belong to which cluster.
This would be extremely useful when developing a new model
incorporating the difficulty of visual questions,
and also when evaluating performances
for different difficulty levels (not for different question types).

\begin{figure*}
    \centering

\begin{minipage}[b]{.2\linewidth}\centering
\includegraphics[width=\linewidth]{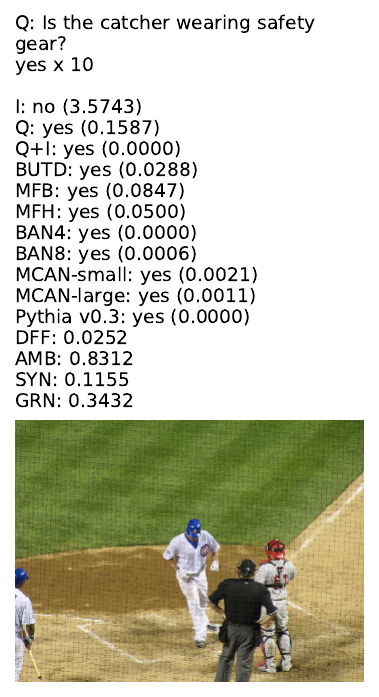}
\includegraphics[width=\linewidth]{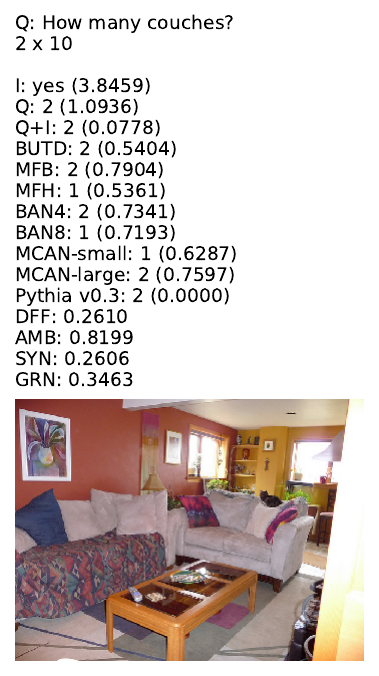}
\includegraphics[width=\linewidth]{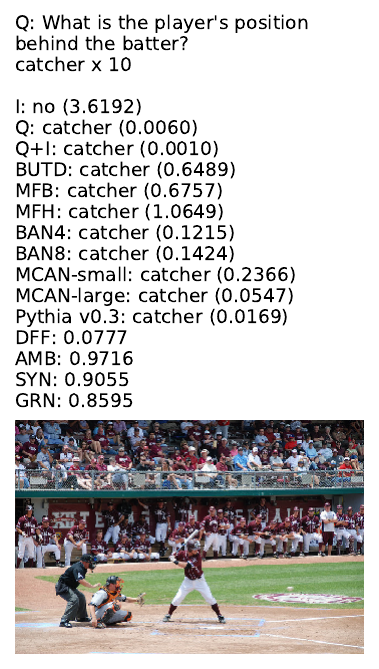}
\\
cluster 0\\
\end{minipage}
\hspace{1em}
\begin{minipage}[b]{.2\linewidth}\centering
\includegraphics[width=\linewidth]{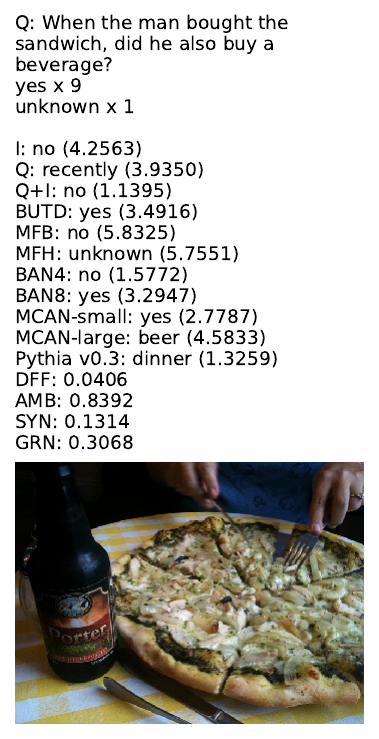}
\includegraphics[width=\linewidth]{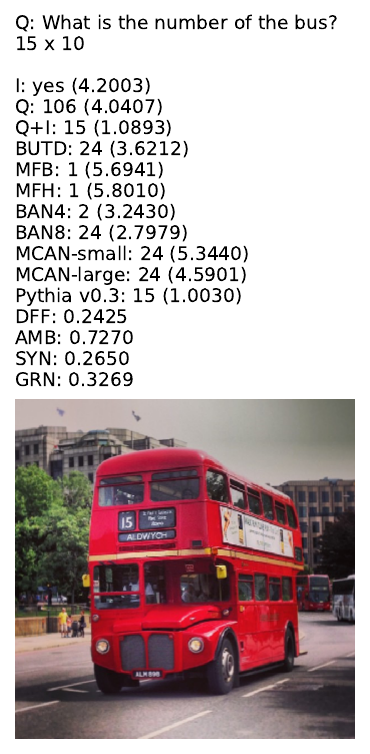}
\includegraphics[width=\linewidth]{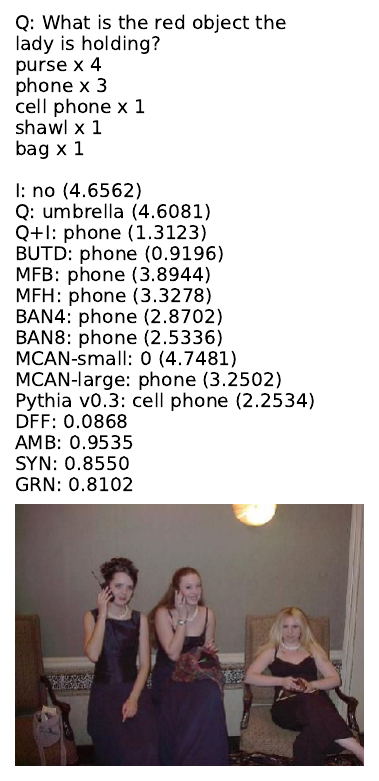}
\\
cluster 4\\
\end{minipage}
\hspace{1em}
\begin{minipage}[b]{.2\linewidth}\centering
\includegraphics[width=\linewidth]{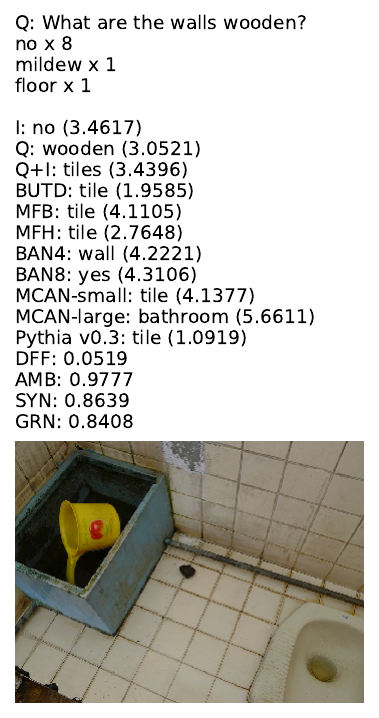}
\includegraphics[width=\linewidth]{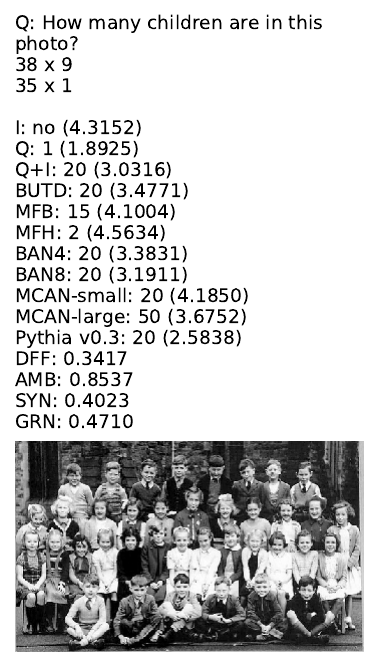}
\includegraphics[width=\linewidth]{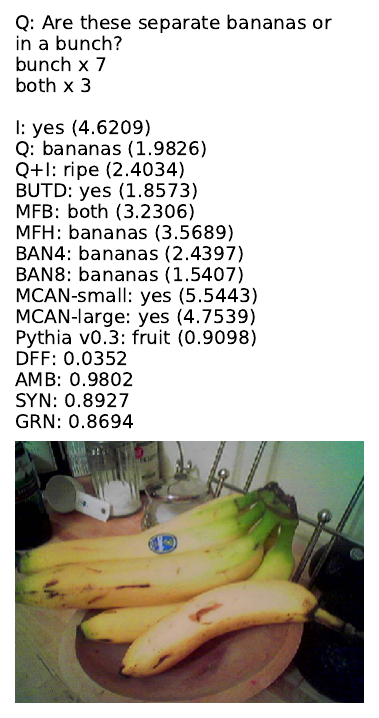}
\\
cluster 8\\
\end{minipage}
\hspace{1em}
\begin{minipage}[b]{.2\linewidth}\centering
\includegraphics[width=.8\linewidth]{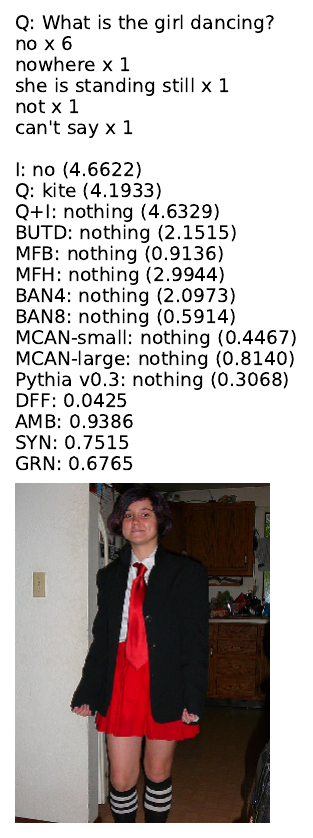}
\includegraphics[width=\linewidth]{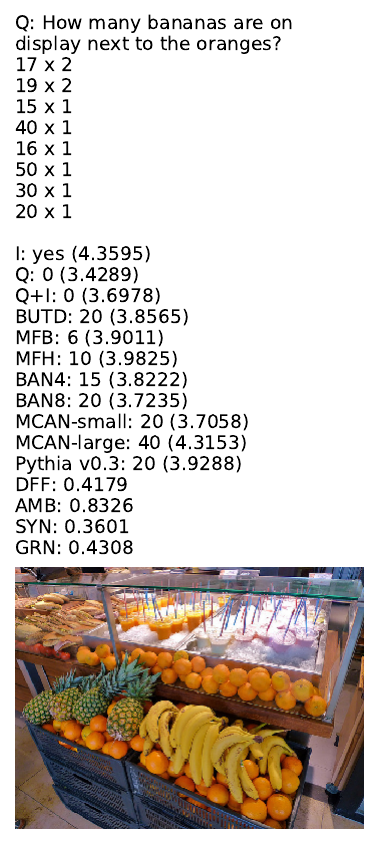}
\includegraphics[width=\linewidth]{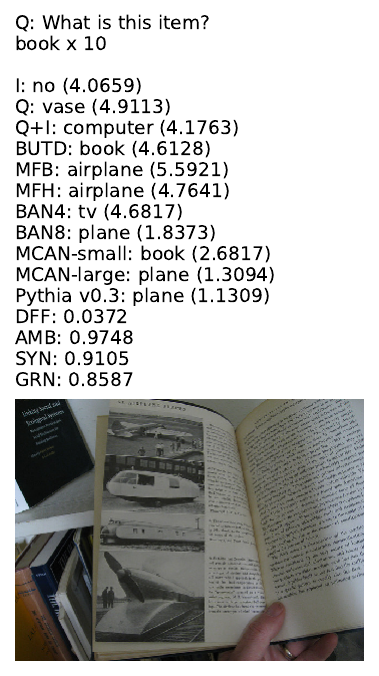}
\\
cluster 9\\
\end{minipage}

    \caption{Examples of visual questions in cluster 0, 4, 8, and 9 (from left to right).
    For each visual question, question text, answers,
    predicted answers and entropy values (in parenthesis) of each method
    are shown, followed by
    values of DFF, AMB, SYN and GRN \cite{Bhattacharya_2019_ICCV}.
    More samples are available online \cite{vqd}.
    }
    \label{fig:examples}
\end{figure*}

\subsubsection{Qualitative evaluation of cluster difficulty}
Figure \ref{fig:examples} shows some examples
of visual questions in each level (from cluster 0, 4, 8, and 9).
Entropy values of different methods tend to be larger in cluster 9,
and 
visual questions in cluster 9 seem to be more difficult than those in cluster 0.
To answer easy questions like ``Is the catcher wearing safety gear?''
or ``What is the player's position behind the batter?'' in cluster 0, images are not necessary 
and the Q model can correctly answer with low entropy.
The question in cluster 9 at the bottom
looks pretty difficult for the models to answer
because of the ambiguity of the question (``What is this item?'') and of the image (containing the photos of vehicles on the page of the book)
even when the human annotators agree on the single ground-truth annotation.

\subsection{Disagreement of predictions of different models}

For difficult visual questions
the number of unique answers is large, \ie annotators highly disagree,
while for easy questions numbers are small and they agree
(5.39 for cluster 9, 1.72 for cluster 0).
Now the following question arises; how much do different models (dis)agree, \ie
do they produce the same answer or different answers?

\begin{figure}
    \centering

\def\cs{.49}
\includegraphics[page=1,width=\cs\linewidth]{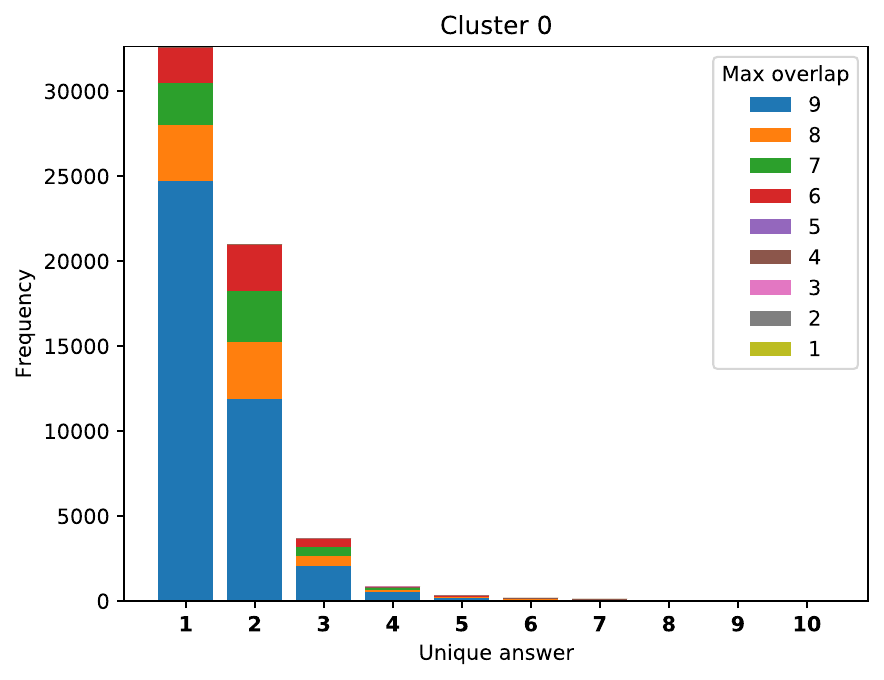}
\includegraphics[page=2,width=\cs\linewidth]{images_nolegend/hist.pdf}
\includegraphics[page=3,width=\cs\linewidth]{images_nolegend/hist.pdf}
\includegraphics[page=4,width=\cs\linewidth]{images_nolegend/hist.pdf}
\includegraphics[page=5,width=\cs\linewidth]{images_nolegend/hist.pdf}
\includegraphics[page=6,width=\cs\linewidth]{images_nolegend/hist.pdf}
\includegraphics[page=7,width=\cs\linewidth]{images_nolegend/hist.pdf}
\includegraphics[page=8,width=\cs\linewidth]{images_nolegend/hist.pdf}
\includegraphics[page=9,width=\cs\linewidth]{images_nolegend/hist.pdf}
\includegraphics[page=10,width=\cs\linewidth]{images_nolegend/hist.pdf}

    \caption{Histograms of visual questions
    with numbers of unique answers of ground truth annotations,
    and max overlap of predicted answers by 9 models.
    }
    \label{fig:model_disagreement}
\end{figure}

\begin{figure}
    \centering

\def\cs{.49}
\includegraphics[page=1,width=\cs\linewidth]{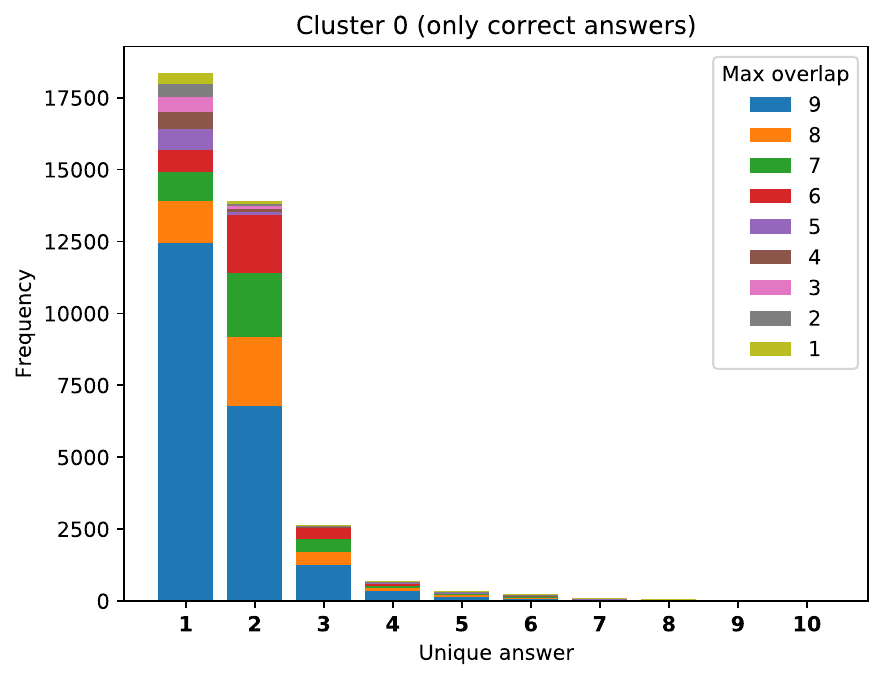}
\includegraphics[page=2,width=\cs\linewidth]{images_nolegend/hist_correct.pdf}
\includegraphics[page=3,width=\cs\linewidth]{images_nolegend/hist_correct.pdf}
\includegraphics[page=4,width=\cs\linewidth]{images_nolegend/hist_correct.pdf}
\includegraphics[page=5,width=\cs\linewidth]{images_nolegend/hist_correct.pdf}
\includegraphics[page=6,width=\cs\linewidth]{images_nolegend/hist_correct.pdf}
\includegraphics[page=7,width=\cs\linewidth]{images_nolegend/hist_correct.pdf}
\includegraphics[page=8,width=\cs\linewidth]{images_nolegend/hist_correct.pdf}
\includegraphics[page=9,width=\cs\linewidth]{images_nolegend/hist_correct.pdf}
\includegraphics[page=10,width=\cs\linewidth]{images_nolegend/hist_correct.pdf}

    \caption{Histograms of visual questions (that are correctly answered by models)
    with numbers of unique answers of ground truth annotations,
    and max overlap of predicted answers by 9 models.
    }
    \label{fig:model_disagreement_correct_only}
\end{figure}

To see this, we define the overlap of model predictions.
We have 9 models (BUTD, MFB, MFH, BAN-4/8, MCAN-small/large, Pythia v0.3 and v0.1 (Q+I)),
and we define the ``overlap'' of the answers to be 9 when all models predict the same answer.
For example, if we have two different answers to a certain question,
each answer produced (supported) by respectively four and five models,
then the answer overlaps are four and five, and we call the larger one a \emph{max overlap}.
Therefore, larger max overlap indicates a higher degree of agreement among the models. 
Figure \ref{fig:model_disagreement} shows histograms of visual questions
with different number of unique answers.
The legend shows the details of max overlap.
Figure \ref{fig:model_disagreement_correct_only} shows similar histograms
but the max overlap is counted with correct model answers only.

For clusters 0 and 1,
almost visual questions have one or two unique answers,
and the models highly agree (max overlap of 9 is dominant).
This is expected because most visual questions in these clusters
are of Yes/No type, and models tend to agree by predicting either of two answers.
Apparently clusters 2, 3, and 4 look similar; dominant max overlap is 9.
This means that
all of 9 models predict the same answer to
almost half of visual questions even when
annotators disagree to five different answers.
In contrast, models predict different answers
to visual questions of clusters 6 -- 9
even when annotators agree and there is a single ground truth answer
(this is the case in the middle of cluster 8 column in Figure \ref{fig:examples}).
Filling this gap may be a promising research direction for the next generation VQA models.

\section{Conclusions}

We have presented a novel way of evaluating the difficulty
of visual questions of the VQA v2 dataset.
Our approach is surprisingly simple,
using three base models (I, Q, Q+I),
predicting answer distributions,
and computing entropy values
to perform clustering with a simple $k$-means.
Experimental results have shown that
these clusters are strongly correlated with
entropy and accuracy values
of many models including state-of-the-art methods.

Our work can be used in many different ways.
One example is to use our work to classify the difficulty of visual questions
to switch different network branches, 
like in \cite{Agrawal_2018_CVPR} which has a question-type classifier to use different branches.
Another example is to apply a curriculum learning framework by
training a model with easy visual questions first, then 
gradually using more difficult ones.
Another possible direction would include judging
if questions generated by VQA and visual dialog models are appropriate to ask. 
However, using our work as a component
is not the only possible way, because 
our work provides additional insights into visual questions.
For example, cluster 9 contains many visual questions
that require reading text in the image, which is recently explored
as a new task by TextVQA \cite{singh2019TowardsVM}.
This cluster also has visual questions of different types of difficulty,
therefore the results of our work
have the potential to inspire more interesting new tasks
that have never been explored.
By providing the correspondences between clusters
and visual questions in the test set as an indicator of difficulty \cite{vqd},
our approach explores a novel aspect of evaluating
performances of VQA models, suggesting a promising direction
for future development of a next generation of VQA models.


\appendix

\section*{Additional Experiments}

\subsection{Further analysis}

Here we give a more detailed analysis of the clustering results.

\subsubsection{Robustness to clustering initialization}

\begin{figure}[t]
\def\myfigheight{1.3cm}
\begin{minipage}[t]{.48\linewidth}
    \centering
    \includegraphics[height=\myfigheight]{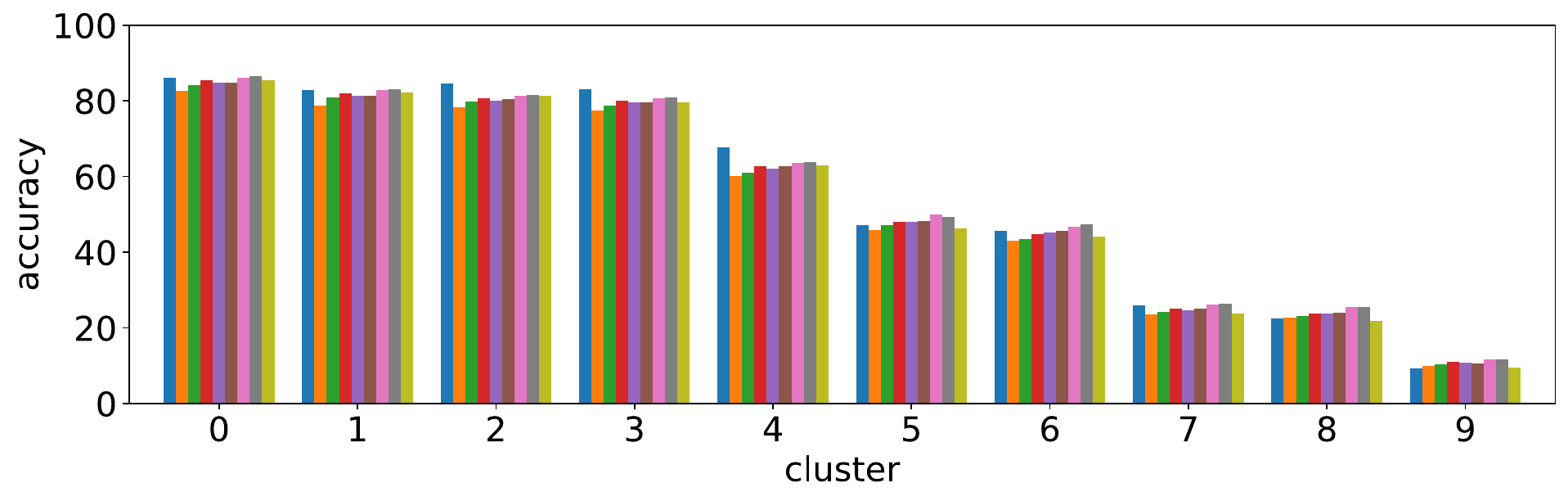}
    \includegraphics[height=\myfigheight]{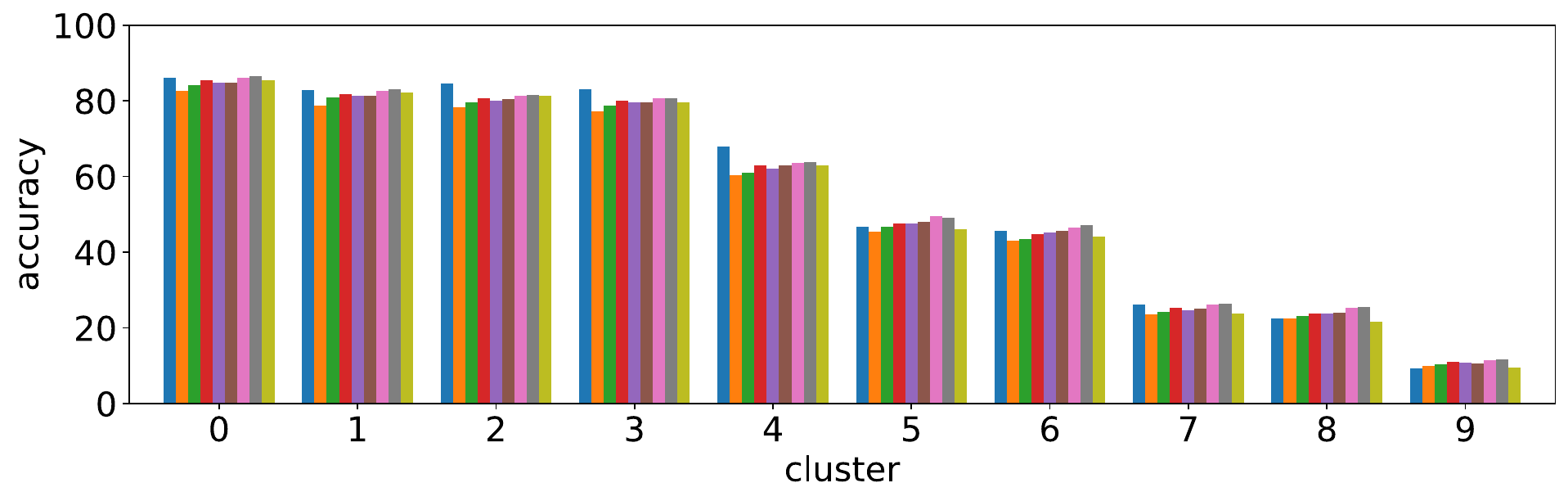}
    \includegraphics[height=\myfigheight]{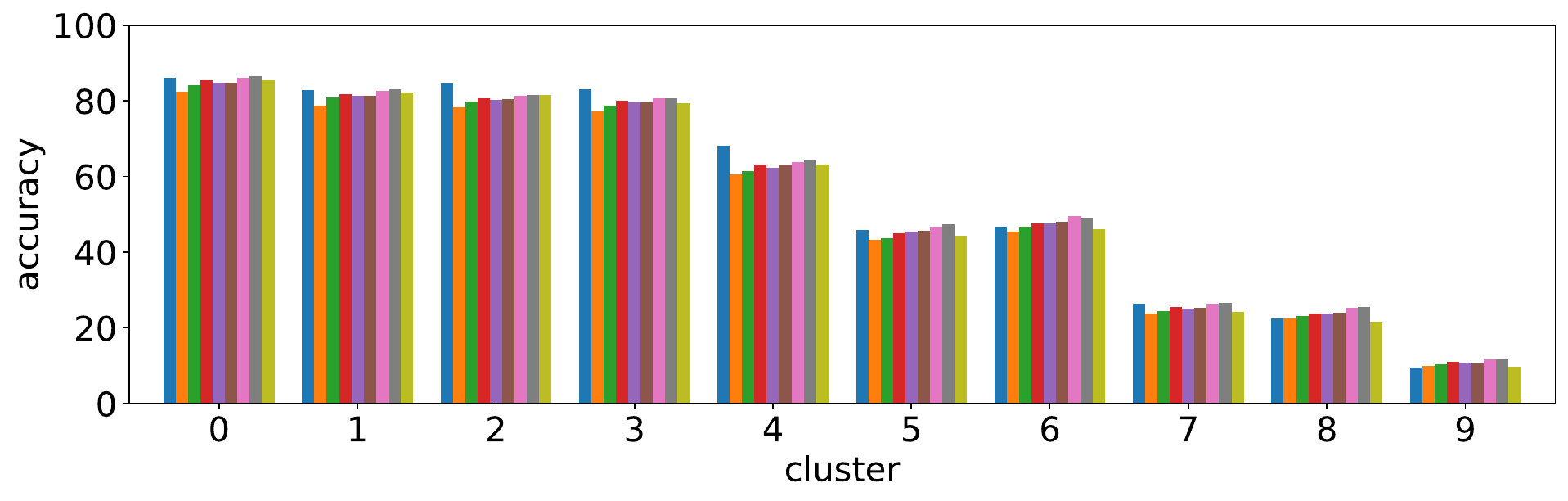}
    \includegraphics[height=\myfigheight]{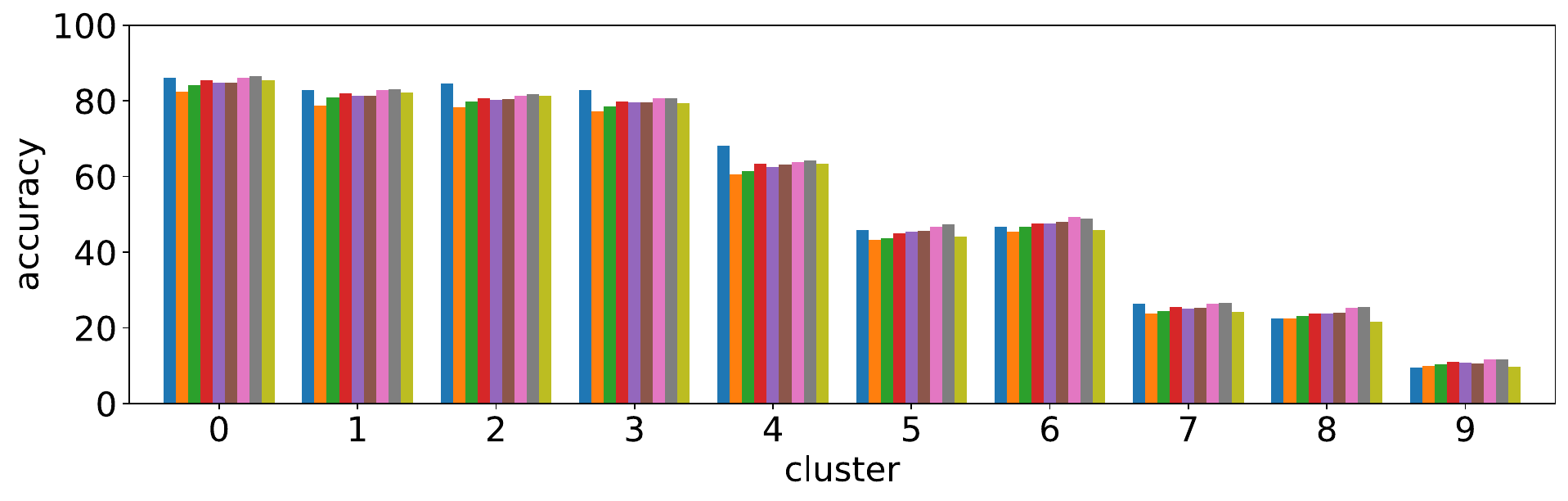}
    \includegraphics[height=\myfigheight]{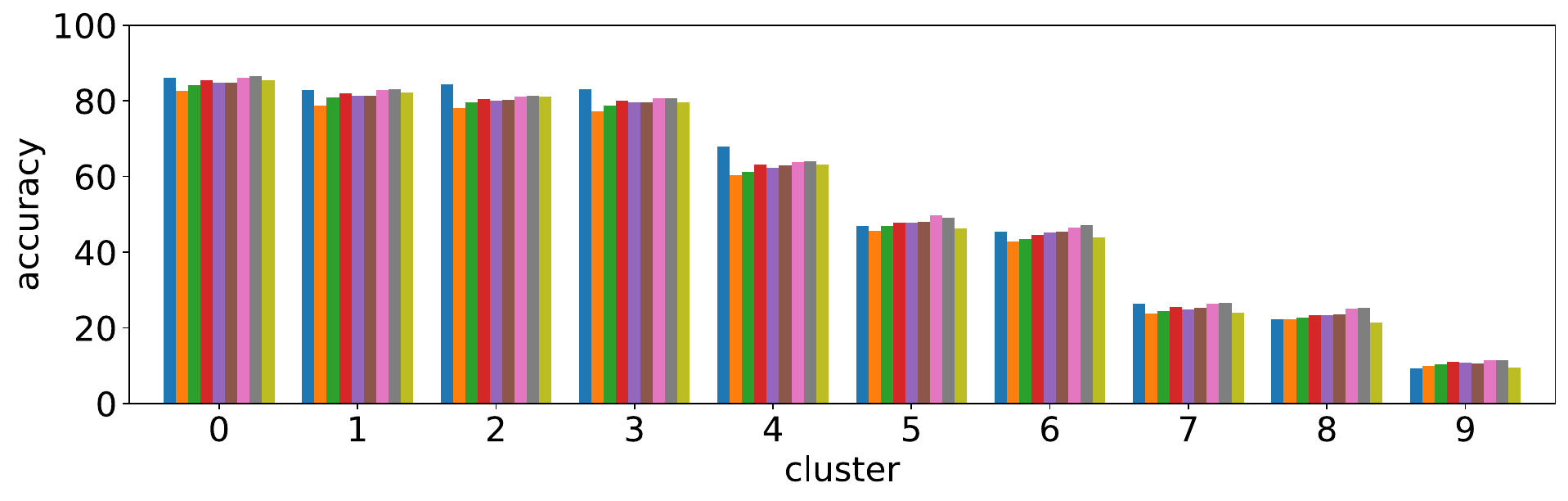}
    \includegraphics[height=\myfigheight]{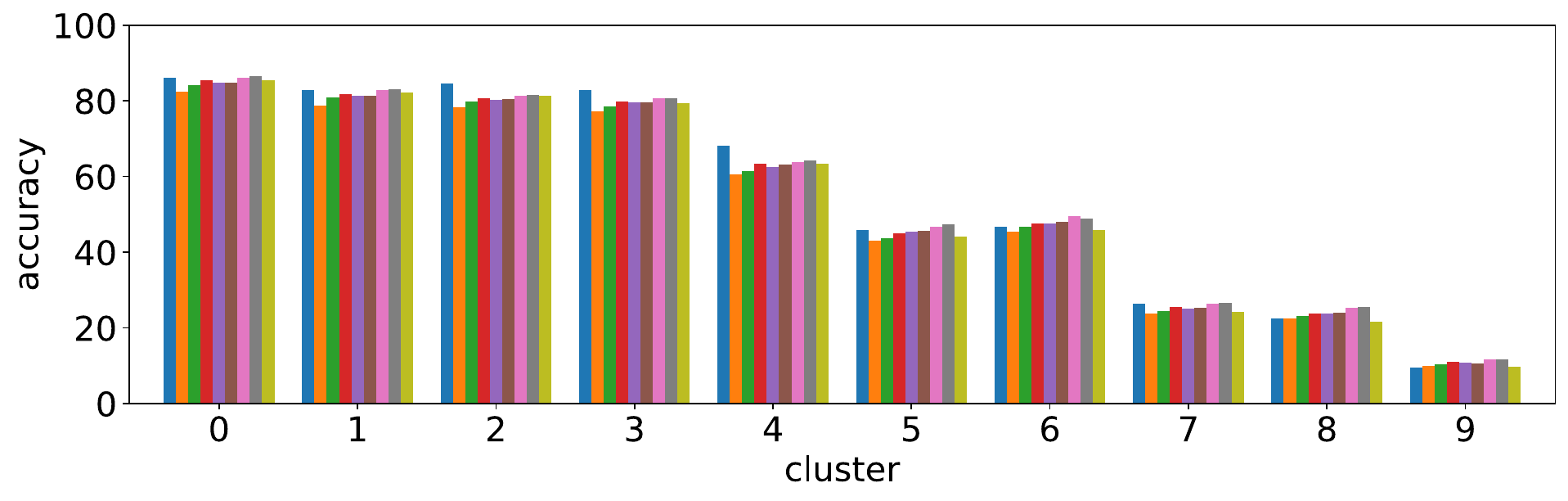}
\end{minipage}
\hfill
\begin{minipage}[t]{.48\linewidth}
    \centering
    \includegraphics[height=\myfigheight]{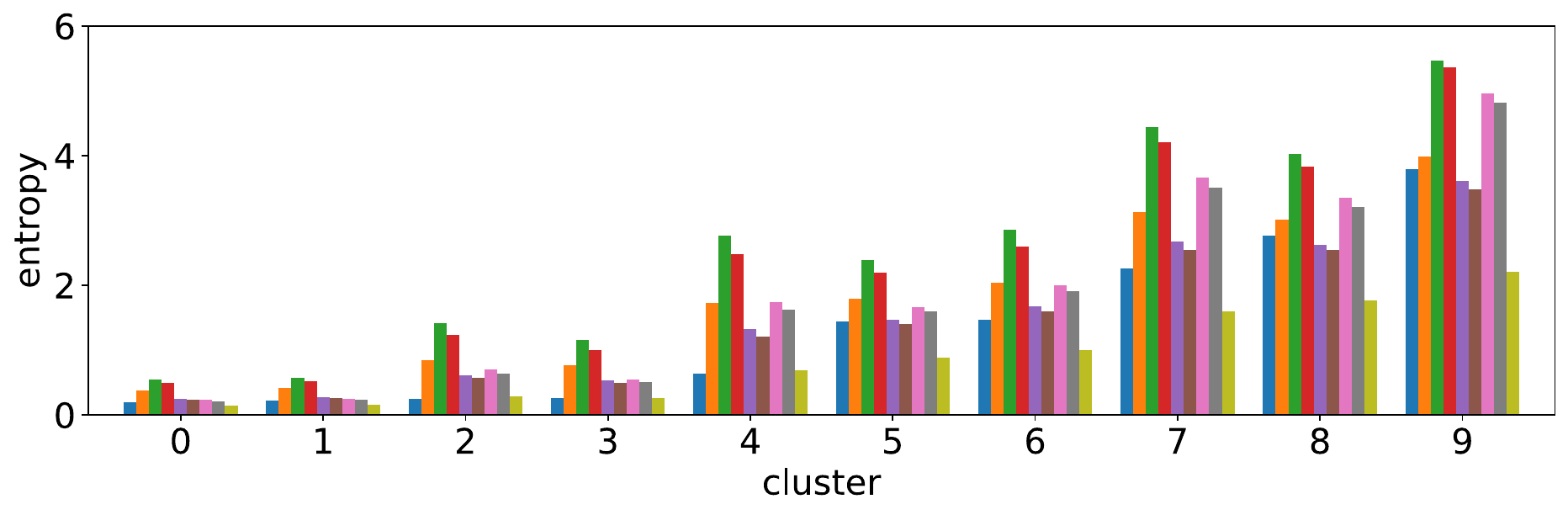}
    \includegraphics[height=\myfigheight]{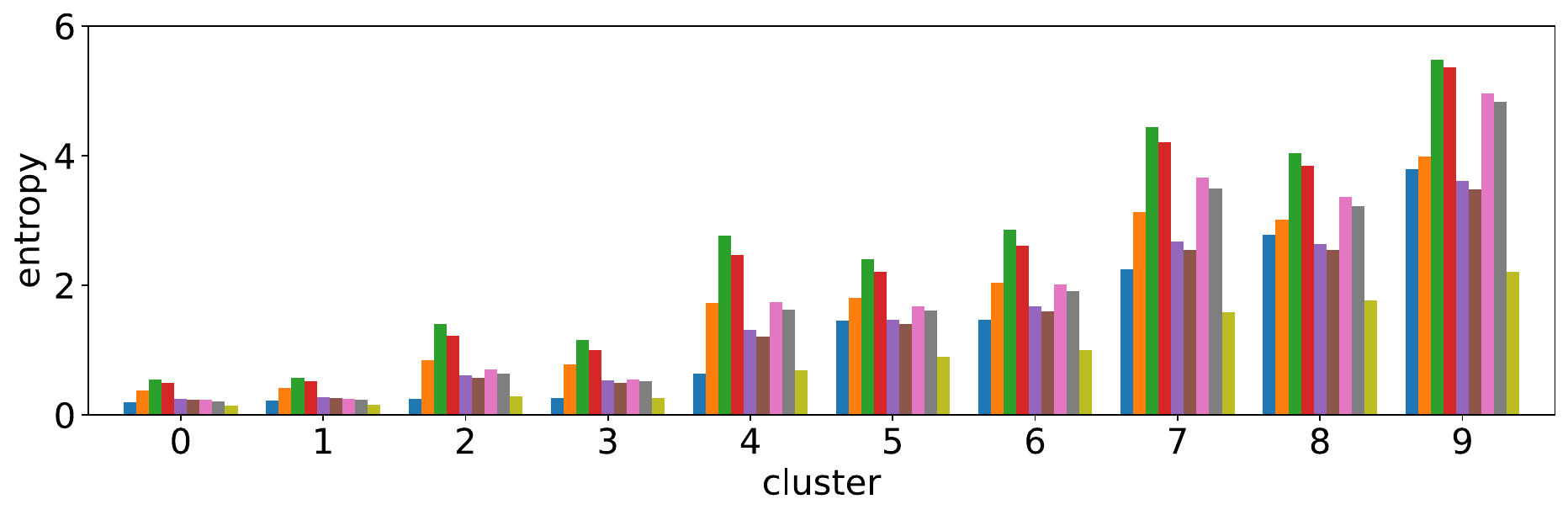}
    \includegraphics[height=\myfigheight]{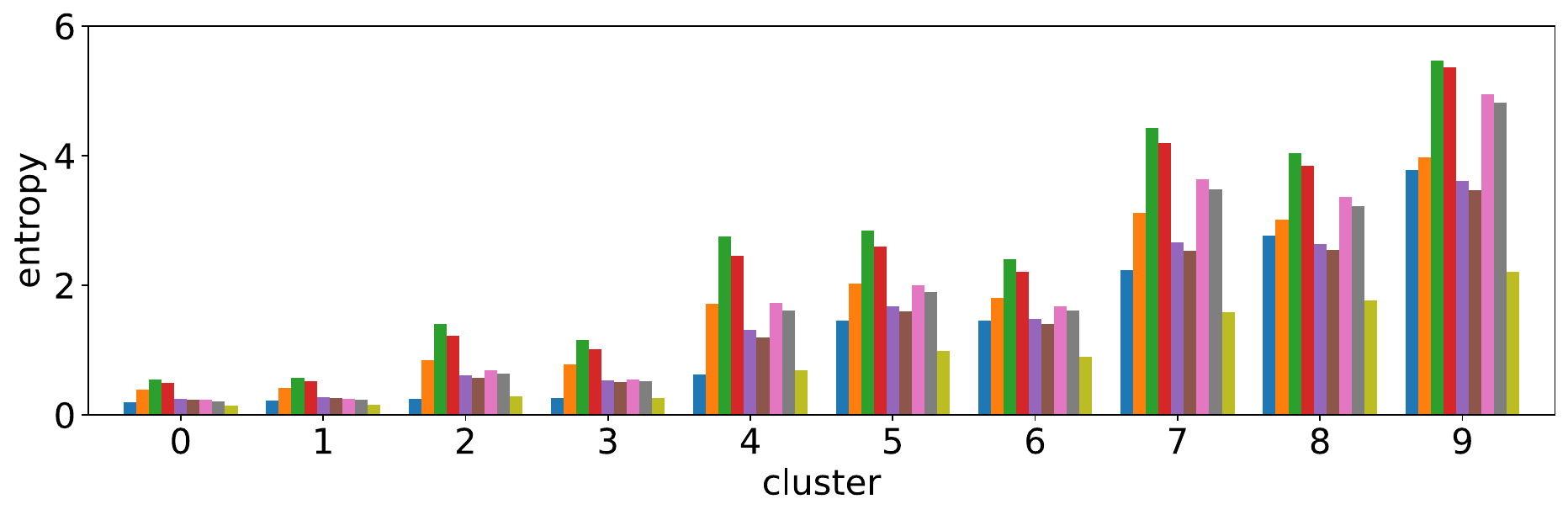}
    \includegraphics[height=\myfigheight]{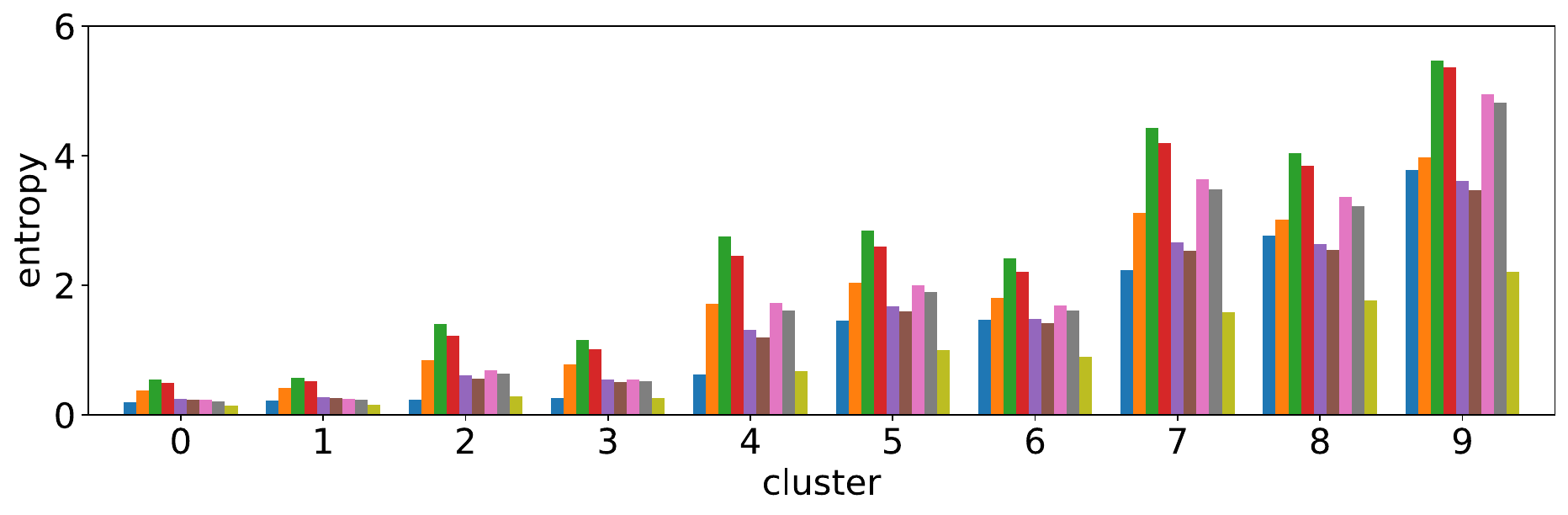}
    \includegraphics[height=\myfigheight]{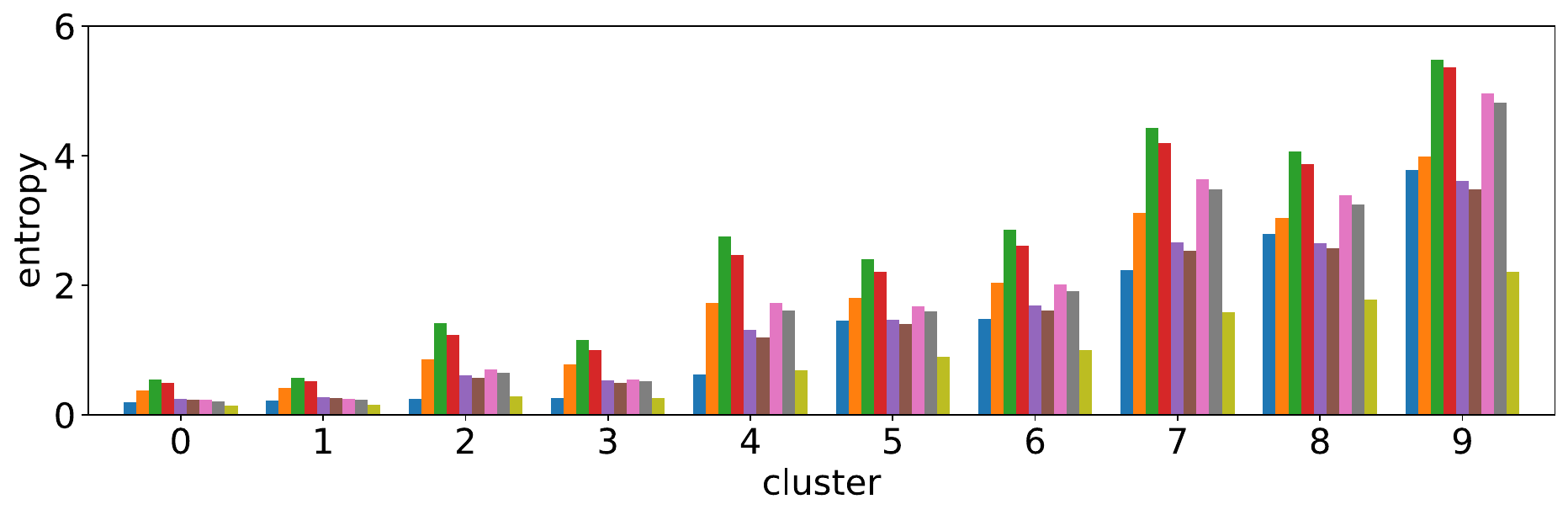}
    \includegraphics[height=\myfigheight]{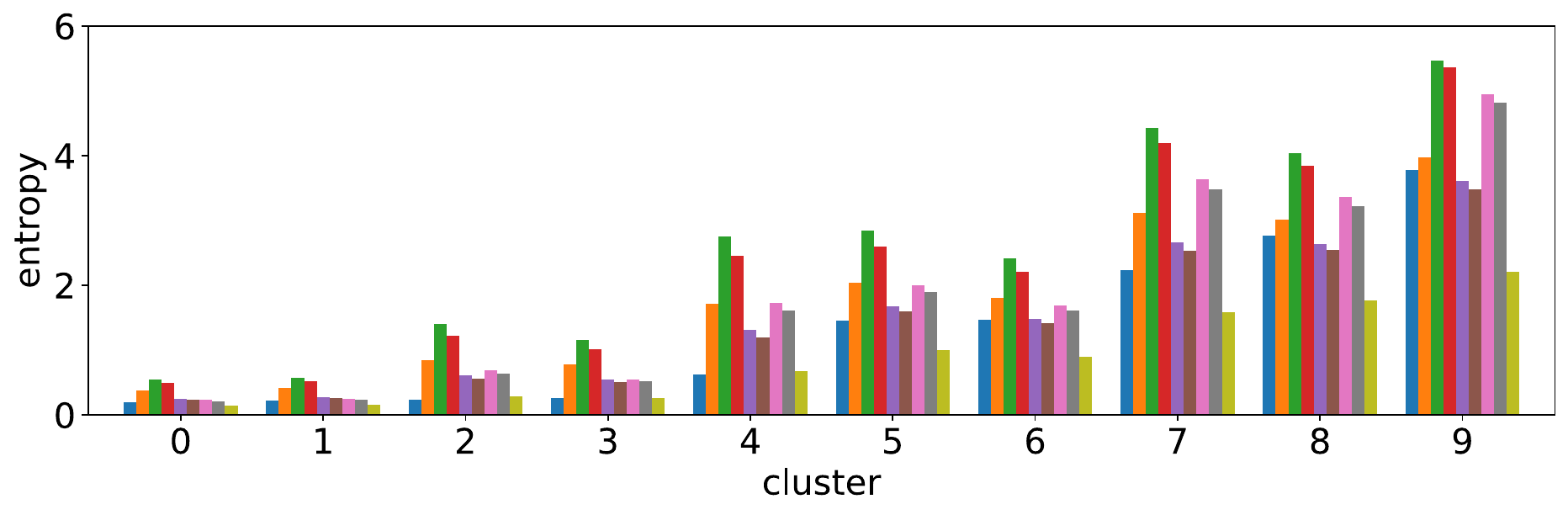}
\end{minipage}
        
    \caption{Values of (left) accuracy and (right) entropy
    for each cluster in different runs of $k$-means clustering with k-means++ initialization.
    The top row is the same with Figure \ref{fig:entropy_accuracy_relation},
    and the rest show five different clustering results obtained when using
     different initializations for the $k$-means.
    }
    \label{fig:accuracy_entropy_cluster_relation}
\end{figure}

\begin{figure}[t]
\def\myfigheight{1.3cm}
\begin{minipage}[t]{.48\linewidth}
    \centering
    \includegraphics[height=\myfigheight]{supp_nolegend/i_q_iq_k10/accuracy_cluster_i_q_iq_0.pdf}
    \includegraphics[height=\myfigheight]{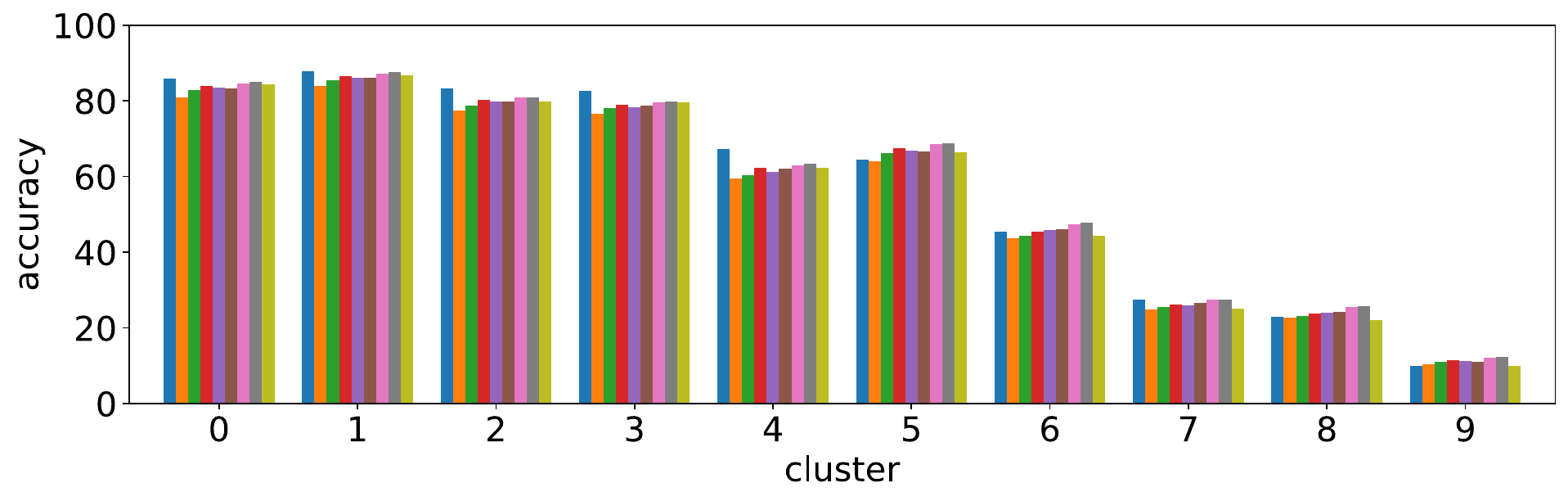}
    \includegraphics[height=\myfigheight]{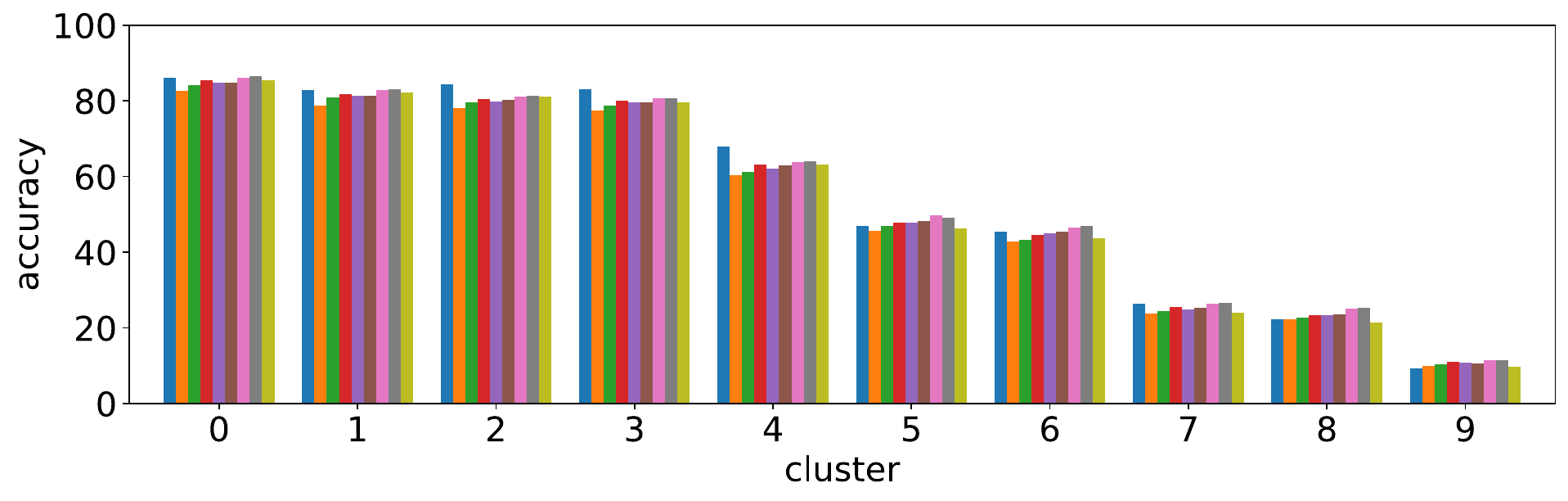}
    \includegraphics[height=\myfigheight]{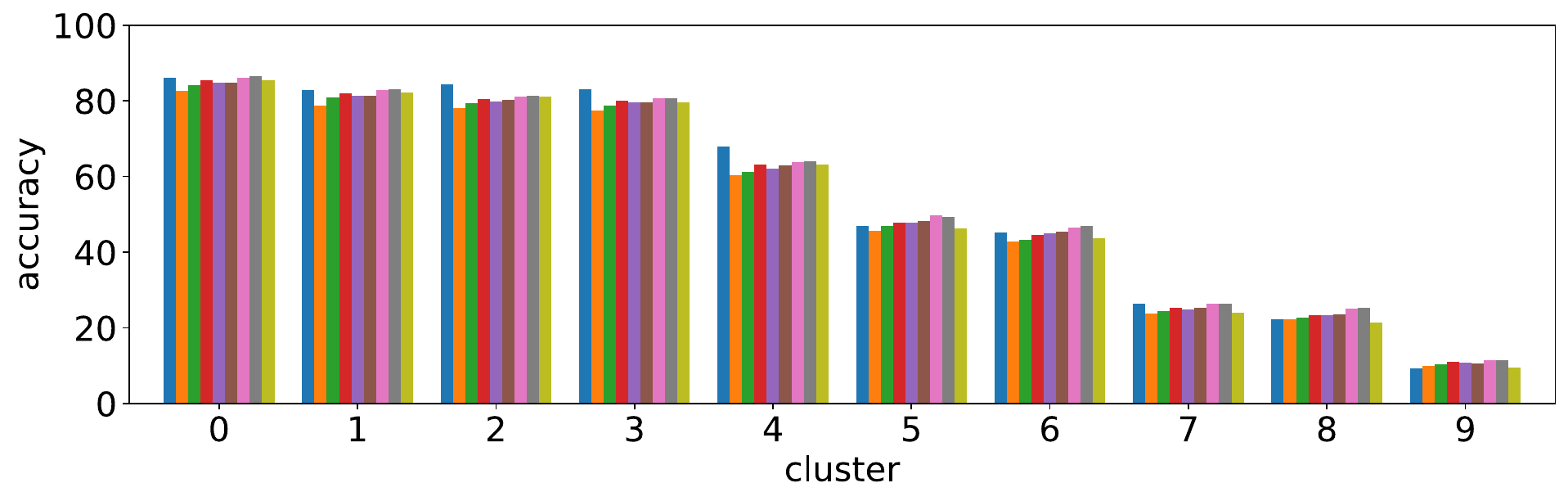}
    \includegraphics[height=\myfigheight]{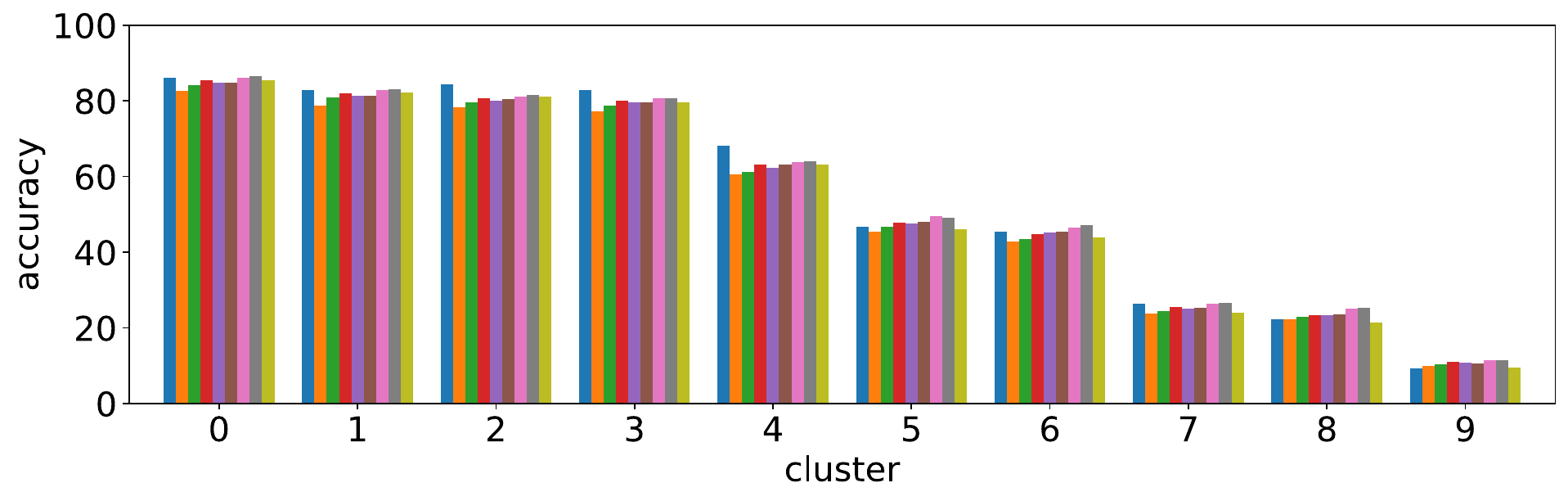}
    \includegraphics[height=\myfigheight]{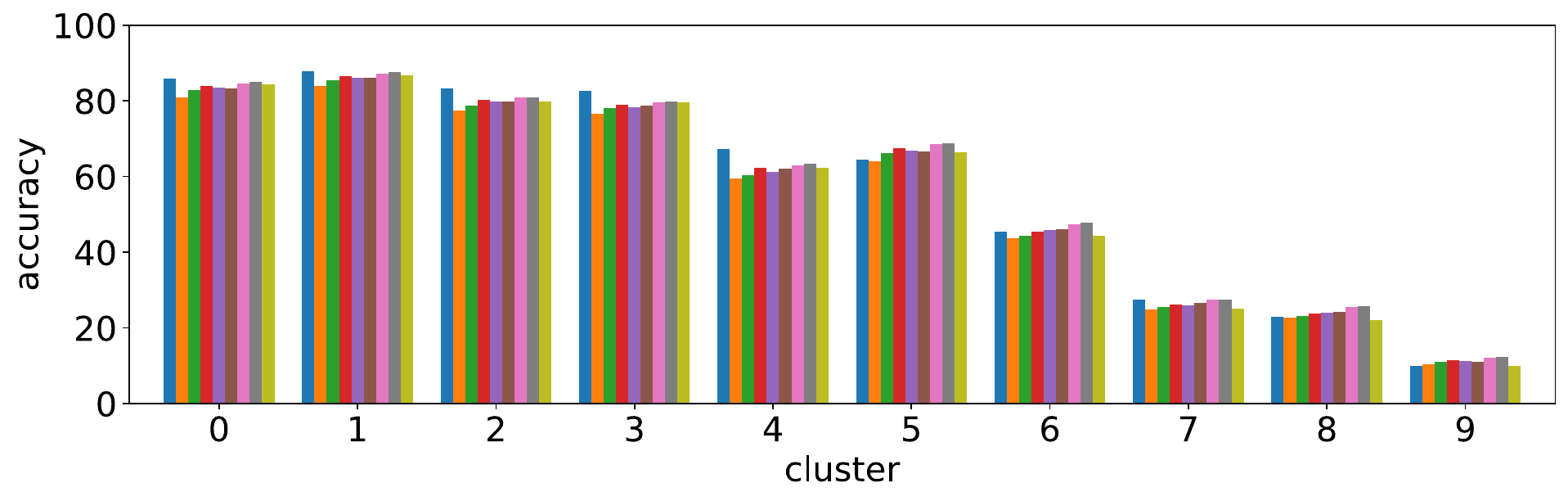}
\end{minipage}
\hfill
\begin{minipage}[t]{.48\linewidth}
    \centering
    \includegraphics[height=\myfigheight]{supp_nolegend/i_q_iq_k10/entropy_cluster_i_q_iq_0.pdf}
    \includegraphics[height=\myfigheight]{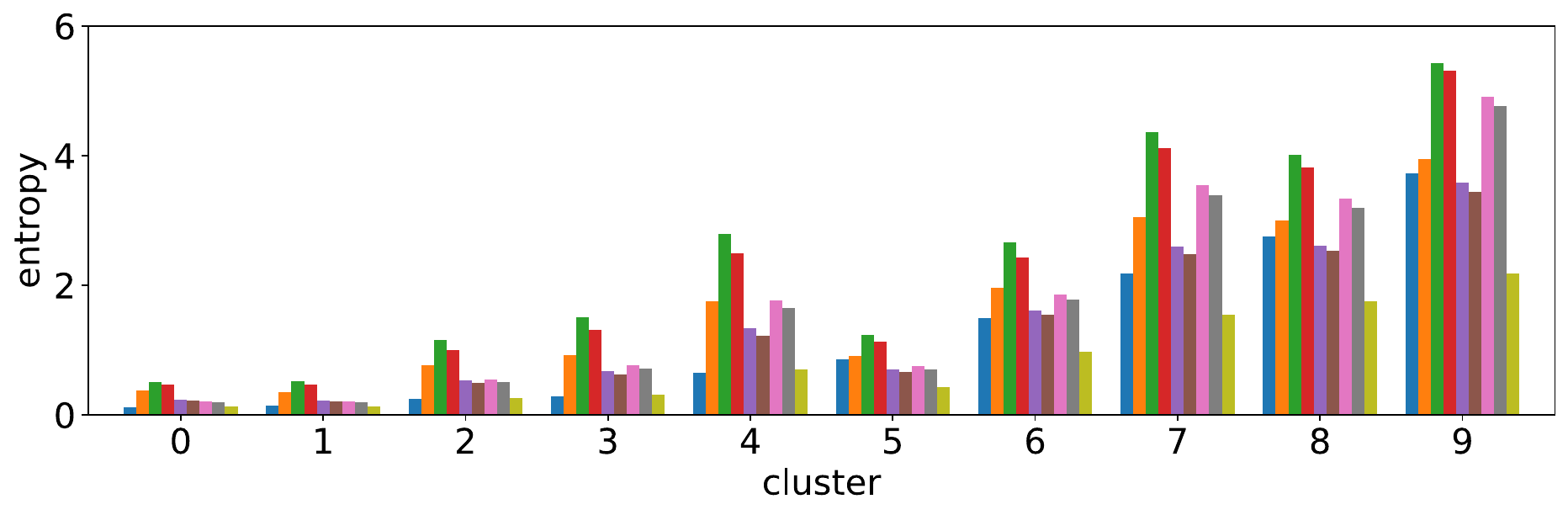}
    \includegraphics[height=\myfigheight]{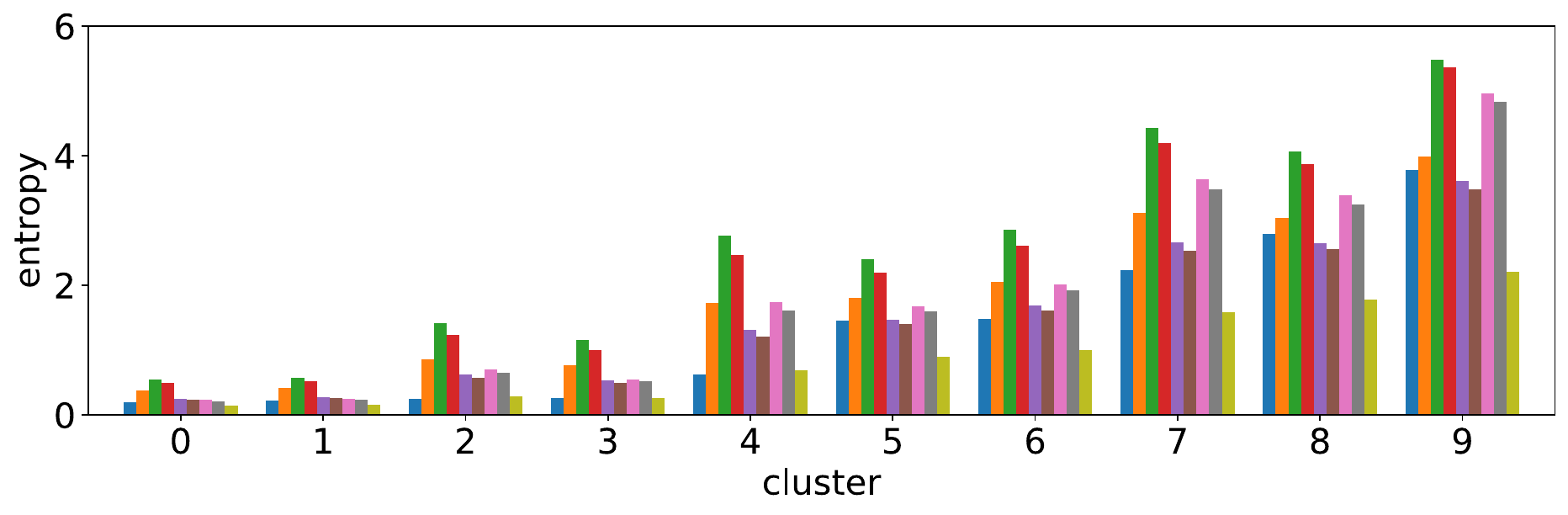}
    \includegraphics[height=\myfigheight]{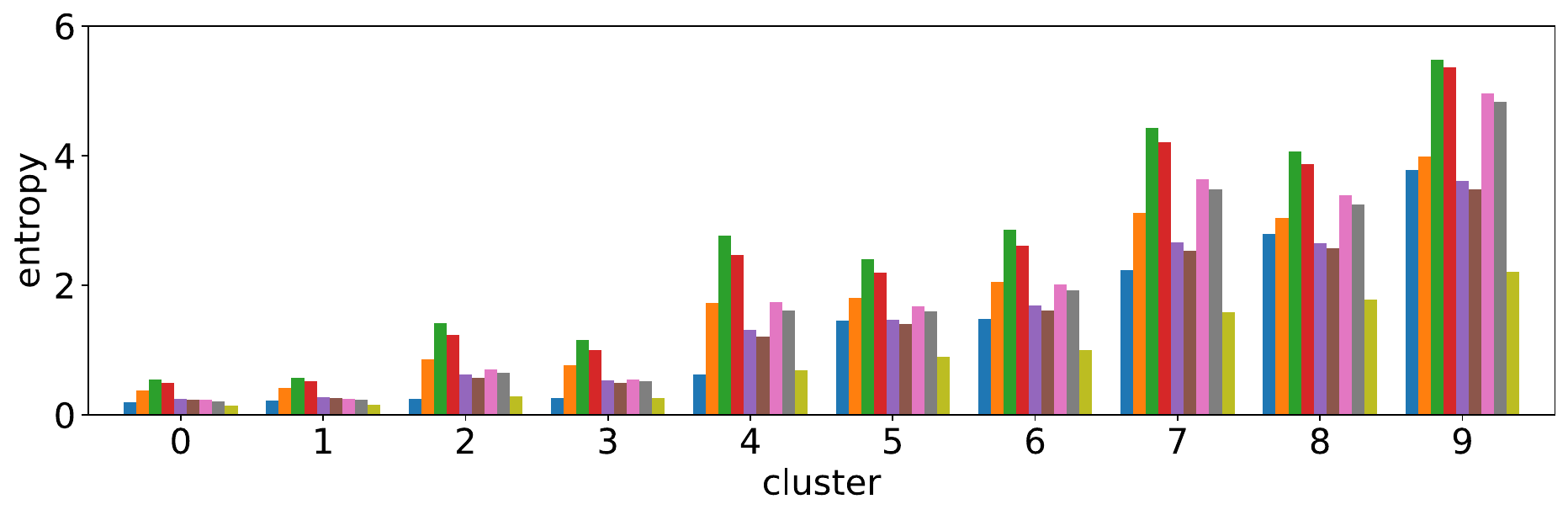}
    \includegraphics[height=\myfigheight]{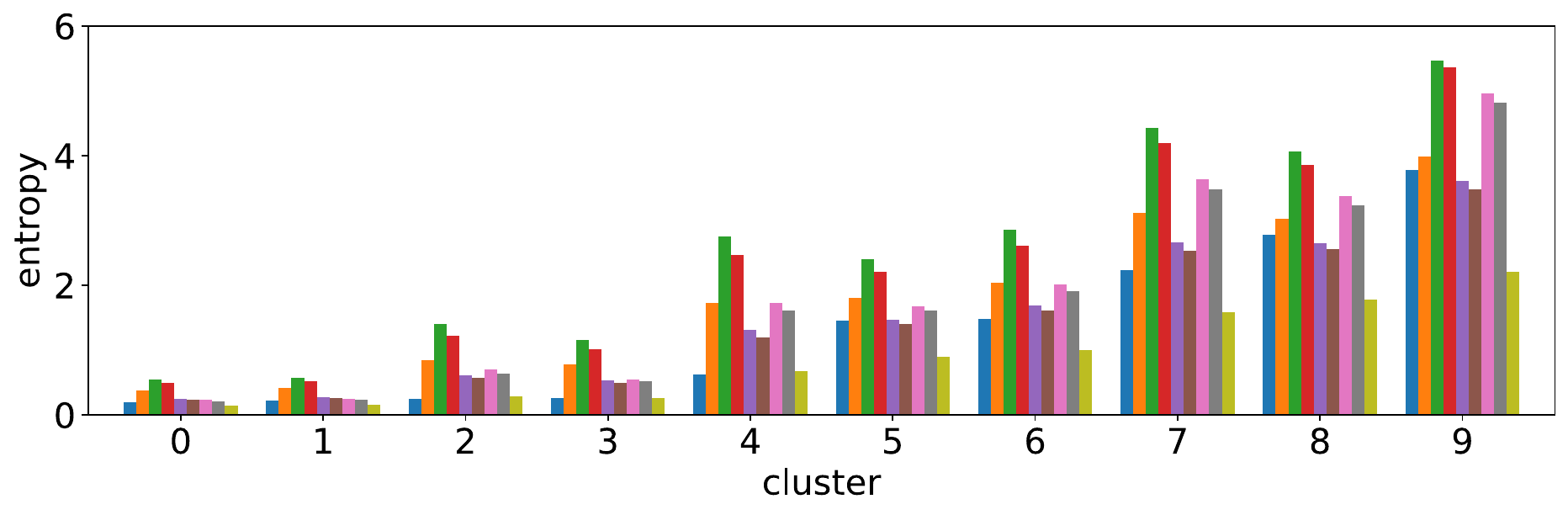}
    \includegraphics[height=\myfigheight]{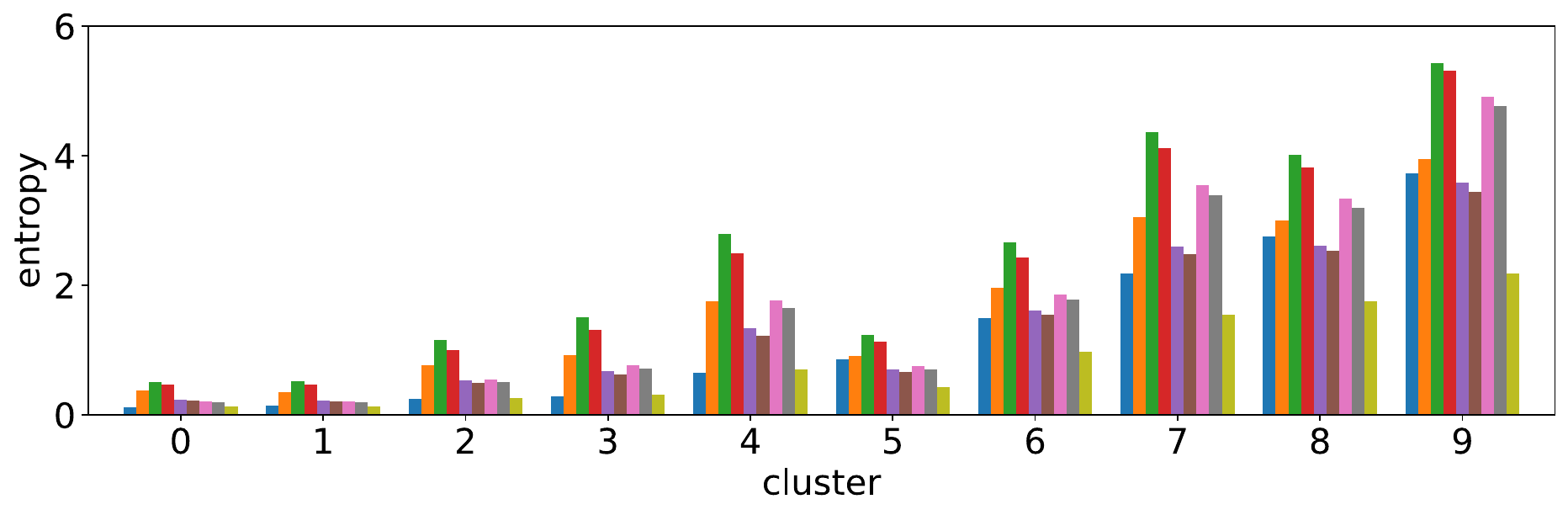}
\end{minipage}
        
    \caption{Values of (left) accuracy and (right) entropy
    for each cluster in different runs of $k$-means clustering with random initialization.
    The top row is the same with Figure \ref{fig:entropy_accuracy_relation},
    and the rest are five different clustering results with
     different initialization of $k$-means.
    }
    \label{fig:accuracy_entropy_cluster_relation2}
\end{figure}

In section \ref{Clustering results},
we argue that similar results are obtained while
many factors including initialization
affect the clustering result.
Figures \ref{fig:accuracy_entropy_cluster_relation}
and \ref{fig:accuracy_entropy_cluster_relation2}
show results corresponding to Figure \ref{fig:entropy_accuracy_relation},
but repeated 5 more times with different initialization of the $k$-means clustering algorithm:
the k-means++ initialization scheme \cite{Arthur2007kmeans++} for Figure \ref{fig:accuracy_entropy_cluster_relation},
and random initialization for Figure \ref{fig:accuracy_entropy_cluster_relation2}.

These figures show that we obtain similar results
even with different initializations of the $k$-means algorithm.
This demonstrates the robustness of our approach to the clustering initialization.

\subsubsection{Robustness to the number of clusters}

\begin{figure}[t]
\def\myfigheight{1.3cm}
\begin{minipage}[t]{.48\linewidth}
    \centering
    \includegraphics[height=\myfigheight]{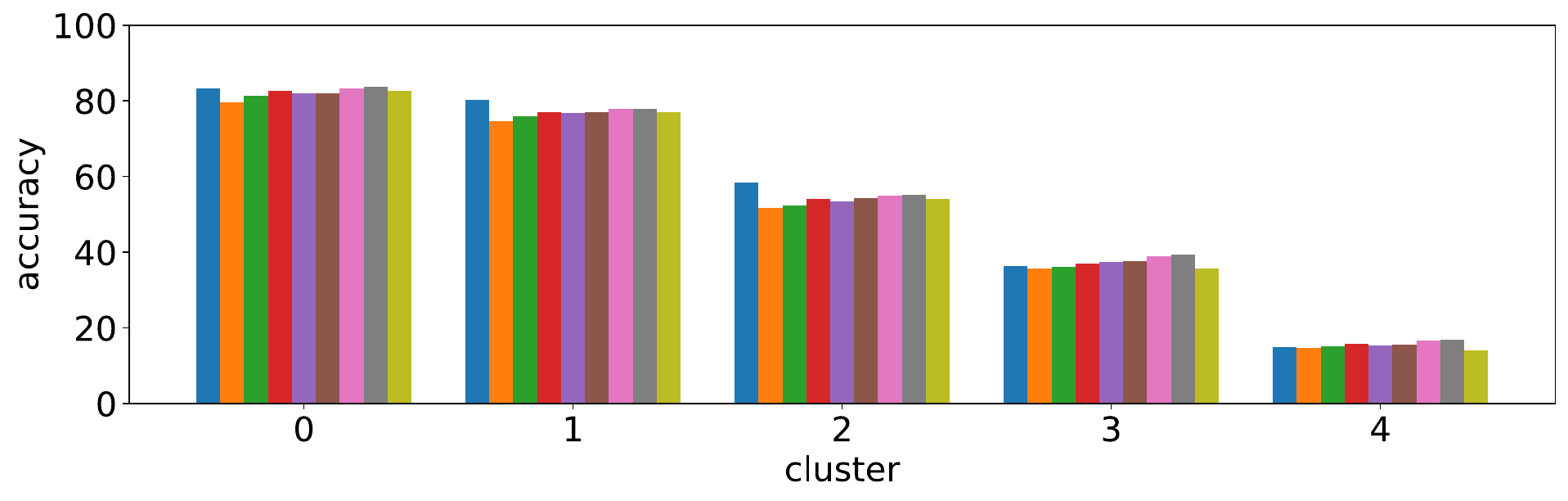}
    \includegraphics[height=\myfigheight]{supp_nolegend/i_q_iq_k10/accuracy_cluster_i_q_iq_0.pdf}
    \includegraphics[height=\myfigheight]{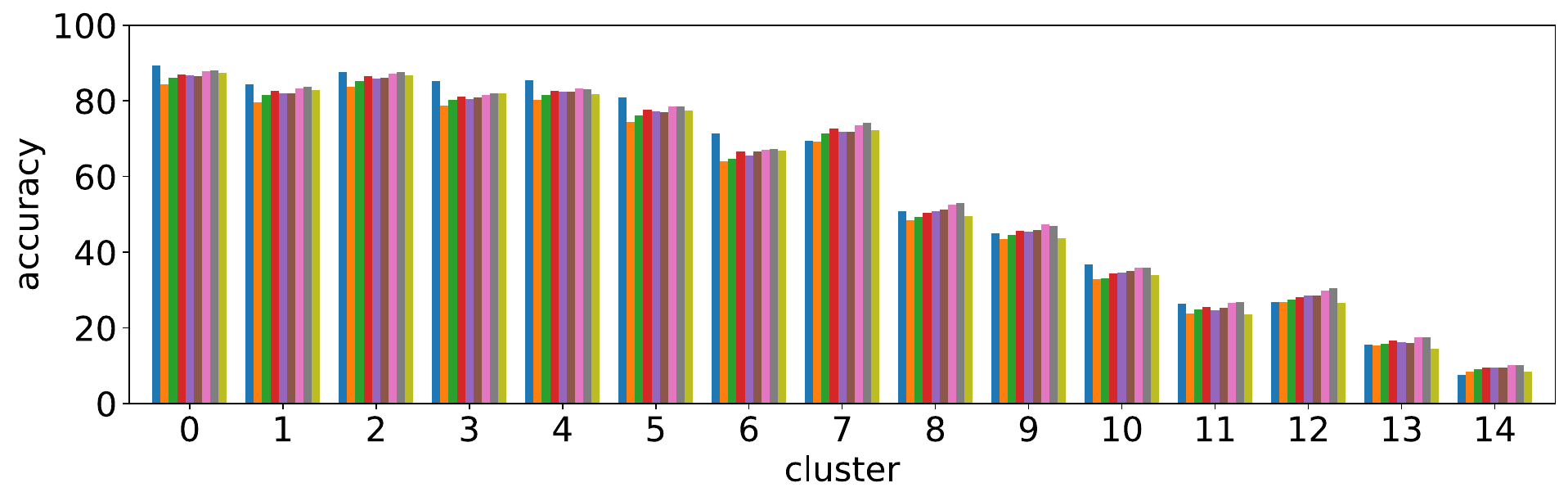}
    \includegraphics[height=\myfigheight]{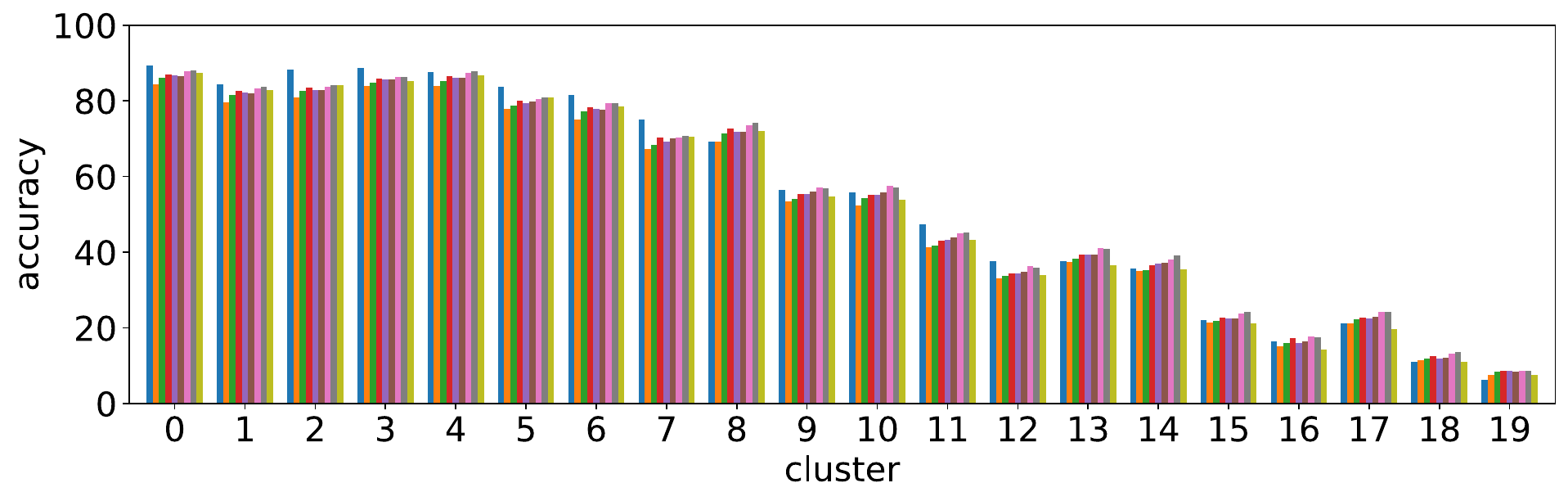}
    \includegraphics[height=\myfigheight]{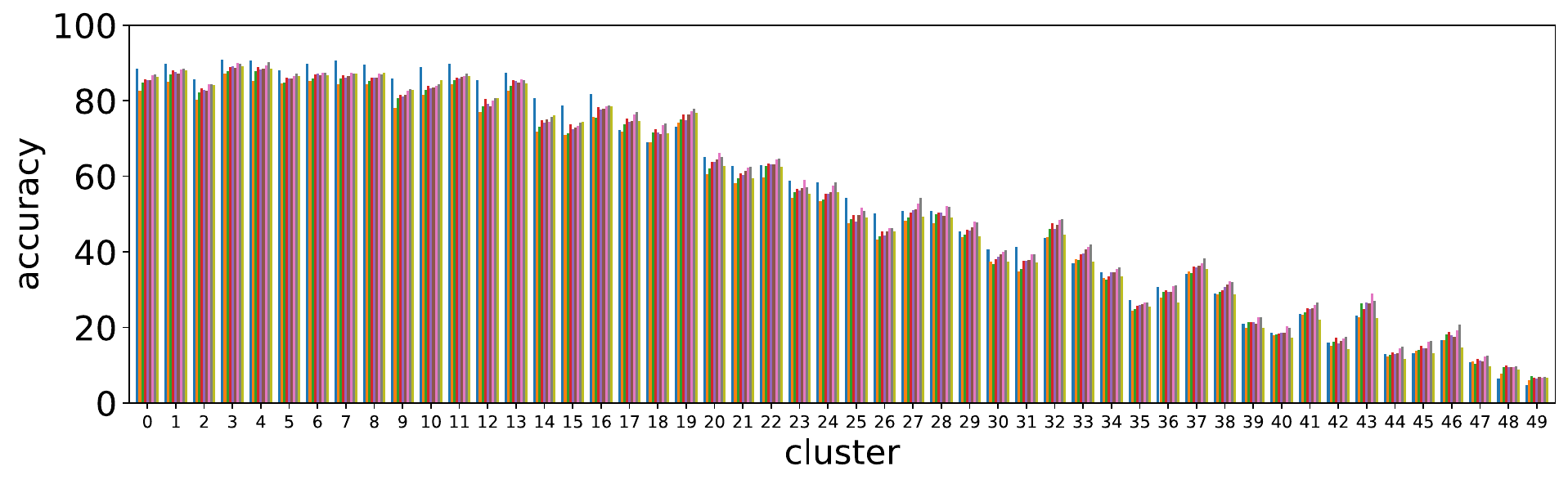}
    \includegraphics[height=\myfigheight]{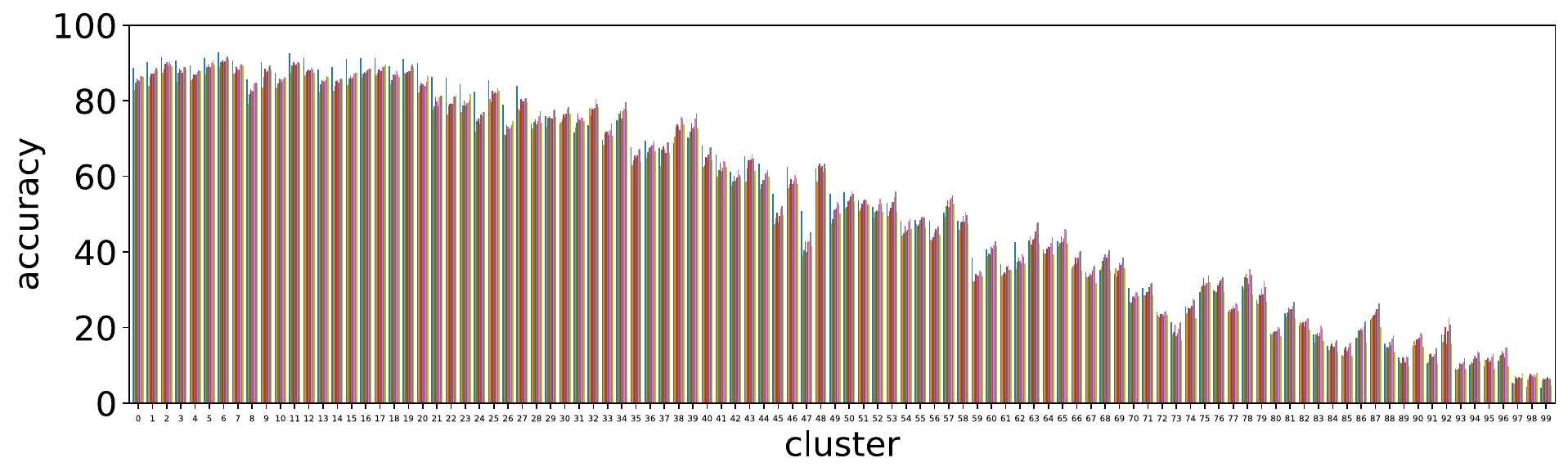}
    \end{minipage}
\hfill
\begin{minipage}[t]{.48\linewidth}
    \centering
    \includegraphics[height=\myfigheight]{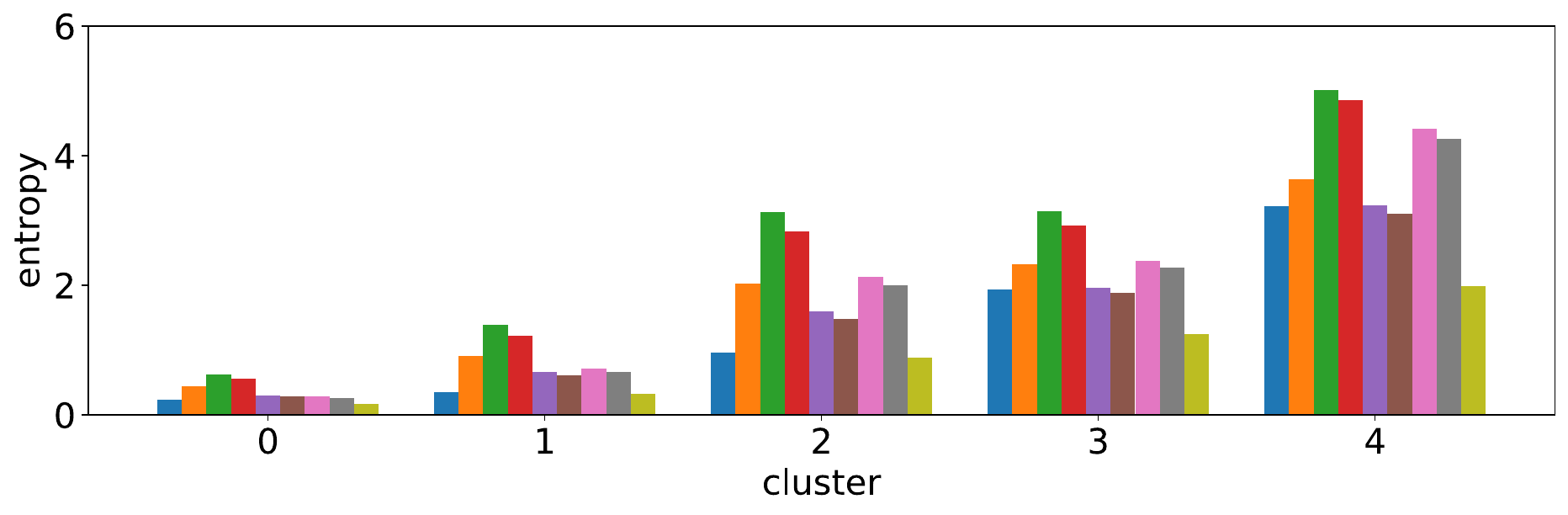}
    \includegraphics[height=\myfigheight]{supp_nolegend/i_q_iq_k10/entropy_cluster_i_q_iq_0.pdf}
    \includegraphics[height=\myfigheight]{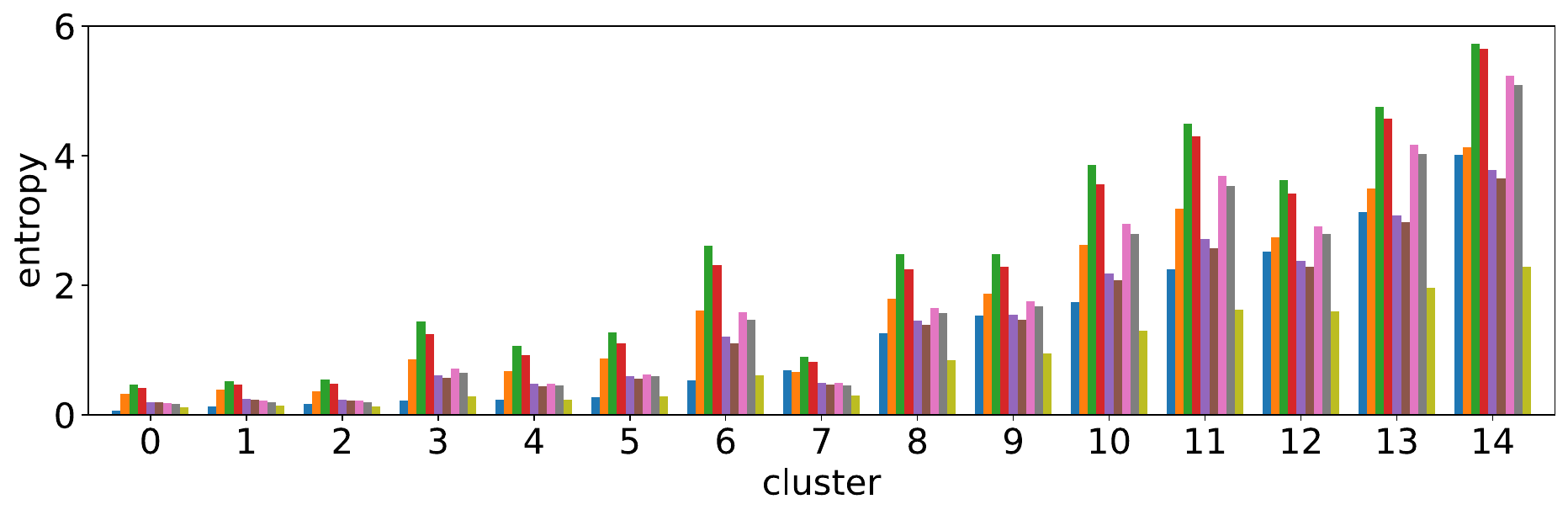}
    \includegraphics[height=\myfigheight]{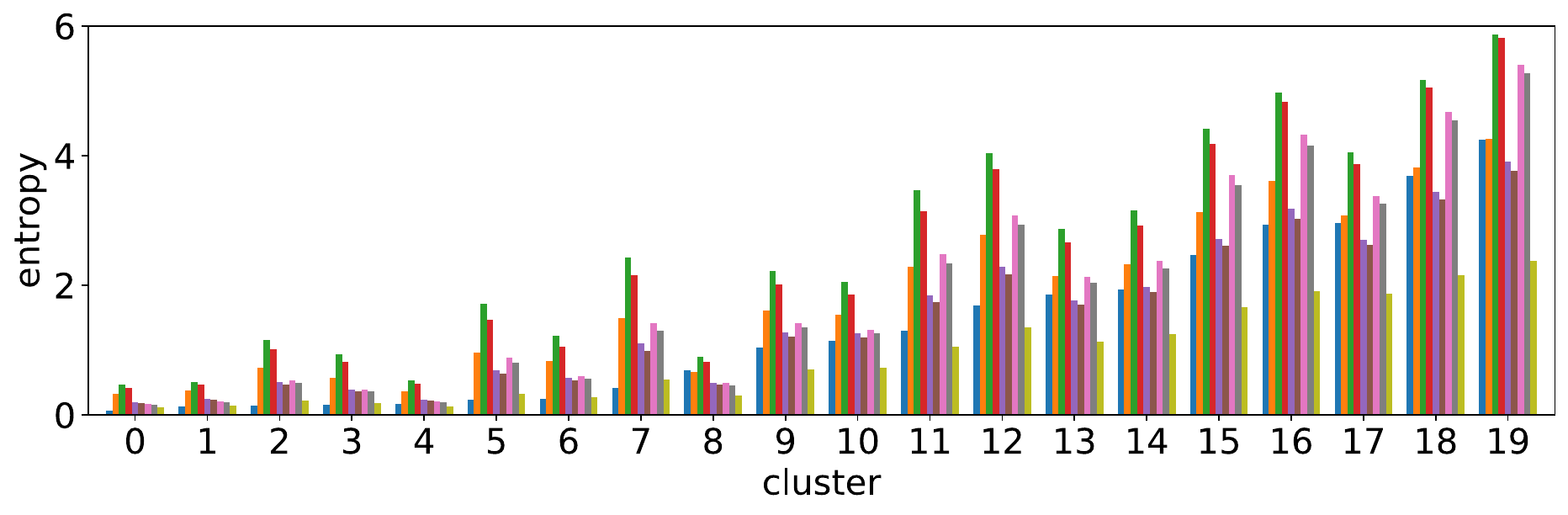}
    \includegraphics[height=\myfigheight]{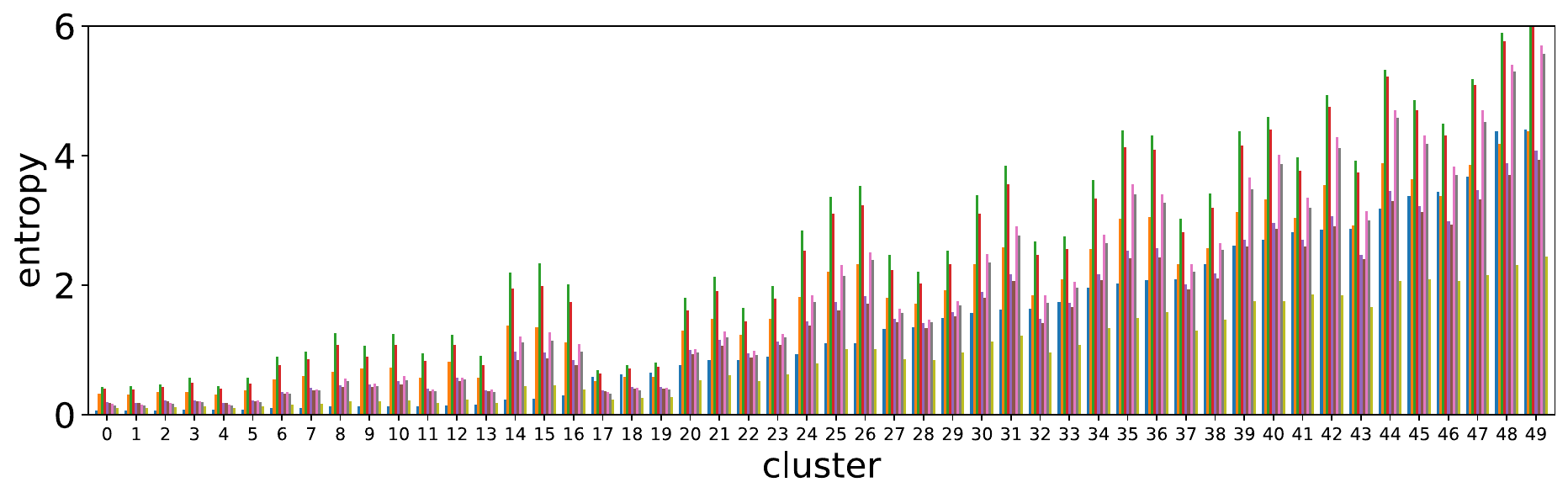}
    \includegraphics[height=\myfigheight]{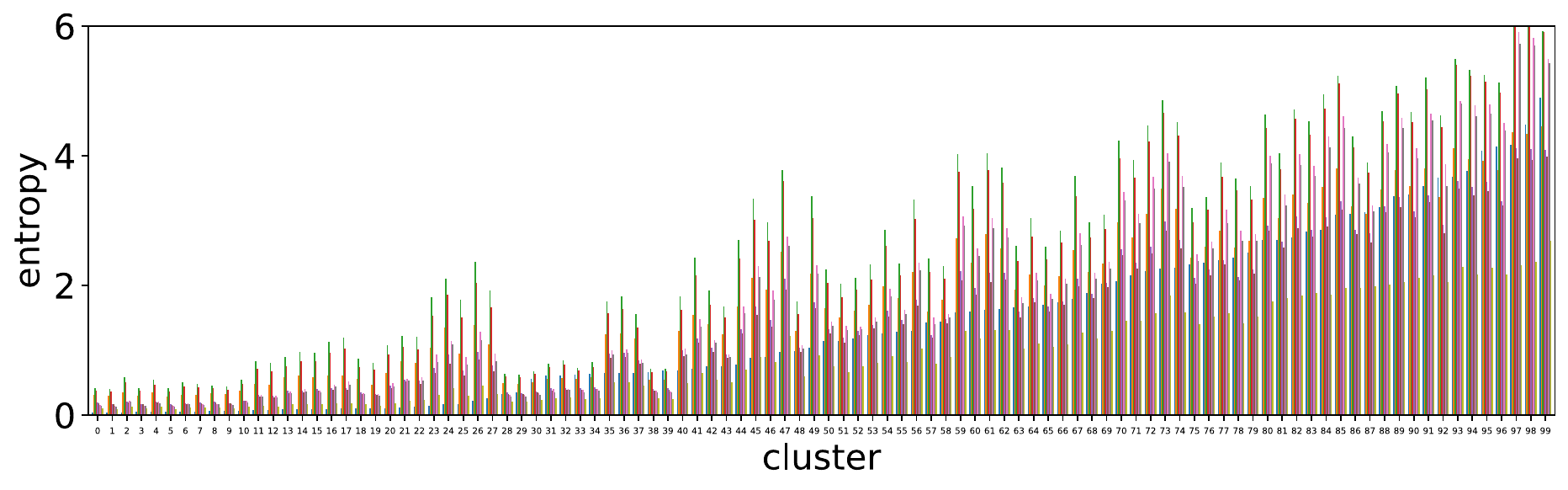}
\end{minipage}

    \caption{Values of (left) accuracy and (right) entropy
    for each cluster when using different values of $k$ in the $k$-means clustering algorithm.
    From top to bottom, values of $k$ are 5, 10, 15, 20, 50, and 100.
    The row for $k=10$ is the same with Figure \ref{fig:entropy_accuracy_relation}.
    }
    \label{fig:accuracy_entropy_different_k_values}
\end{figure}

Figure \ref{fig:accuracy_entropy_different_k_values}
shows results corresponding to Figure \ref{fig:entropy_accuracy_relation},
but with different number of clusters for the $k$-means clustering.
As stated before, the number of clusters affects
the clustering result.
However we can see in these figures that
similar results are obtained with both fewer clusters $k=5$
and more clusters $k=15,20,50,100$.
This also demonstrate the robustness of our approach to the number of clusters.

\subsubsection{Effect of clustering with different features}

\begin{figure}[t]
\def\myfigheight{1.3cm}
\begin{minipage}[t]{.48\linewidth}
    \centering
    \includegraphics[height=\myfigheight]{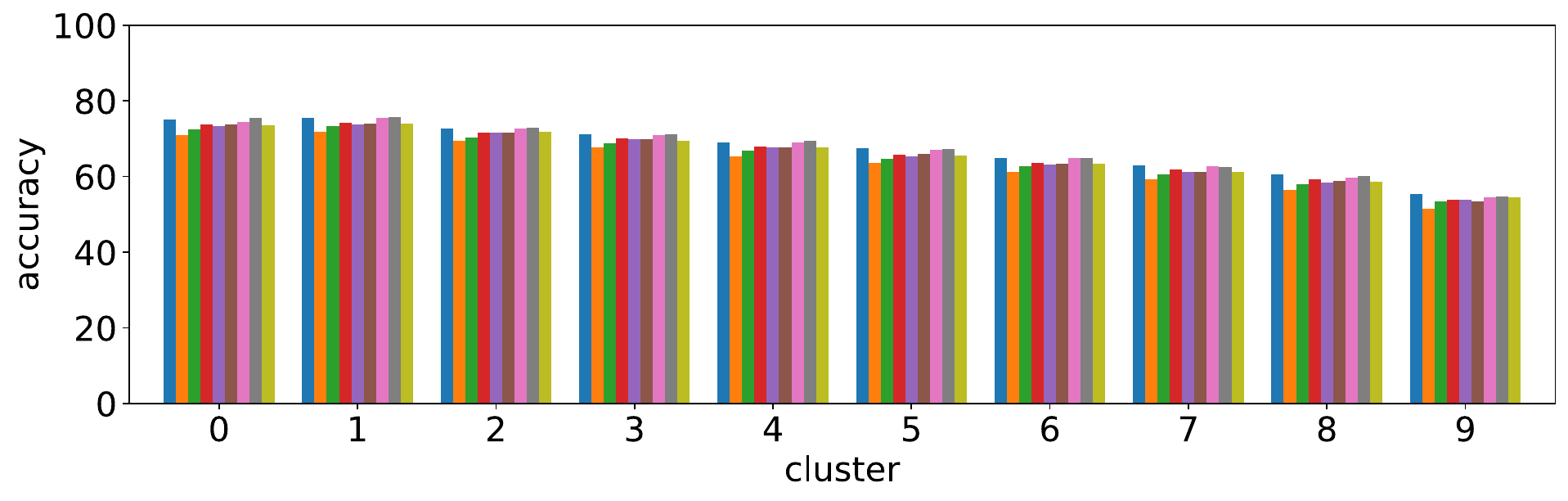}
    \includegraphics[height=\myfigheight]{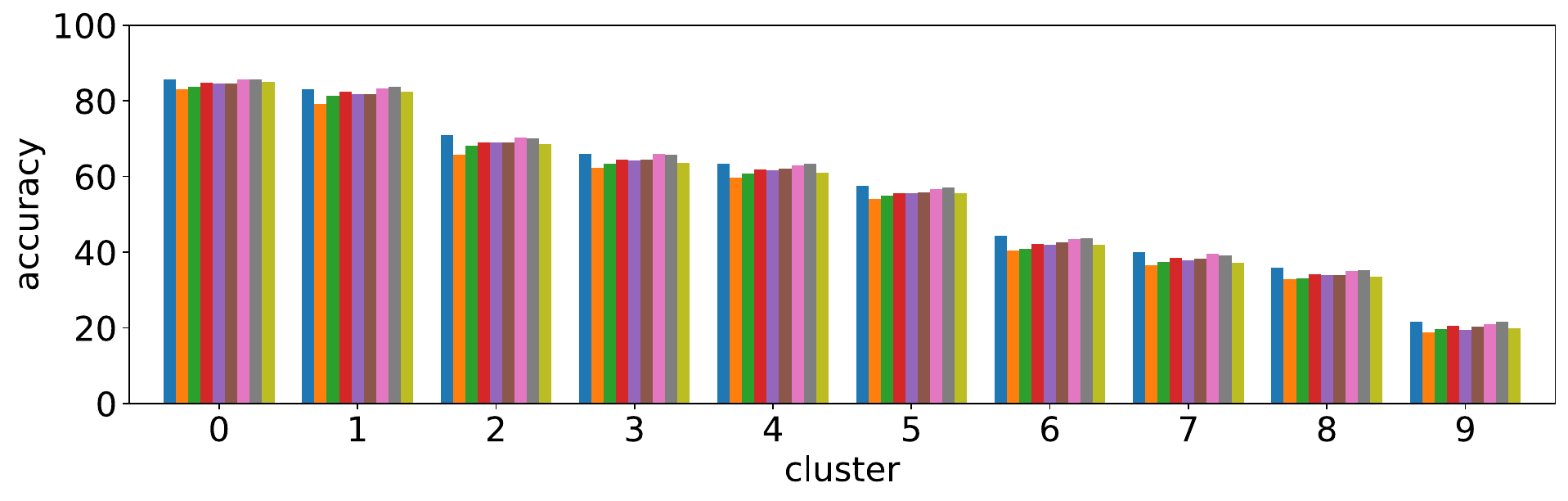}
    \includegraphics[height=\myfigheight]{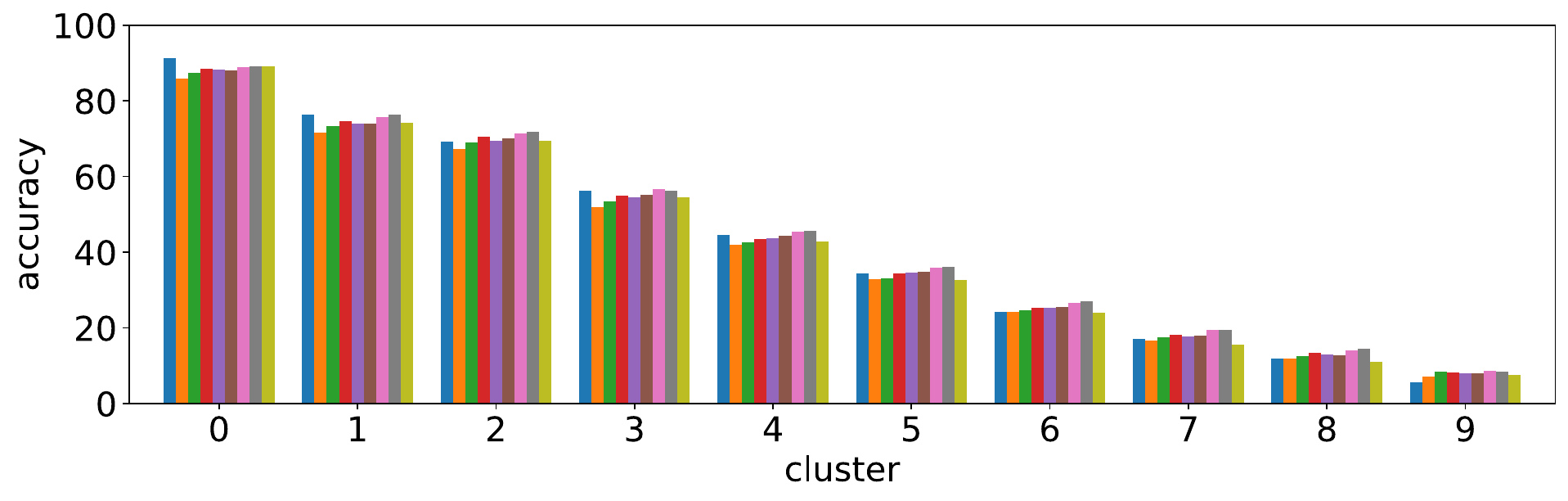}
    \includegraphics[height=\myfigheight]{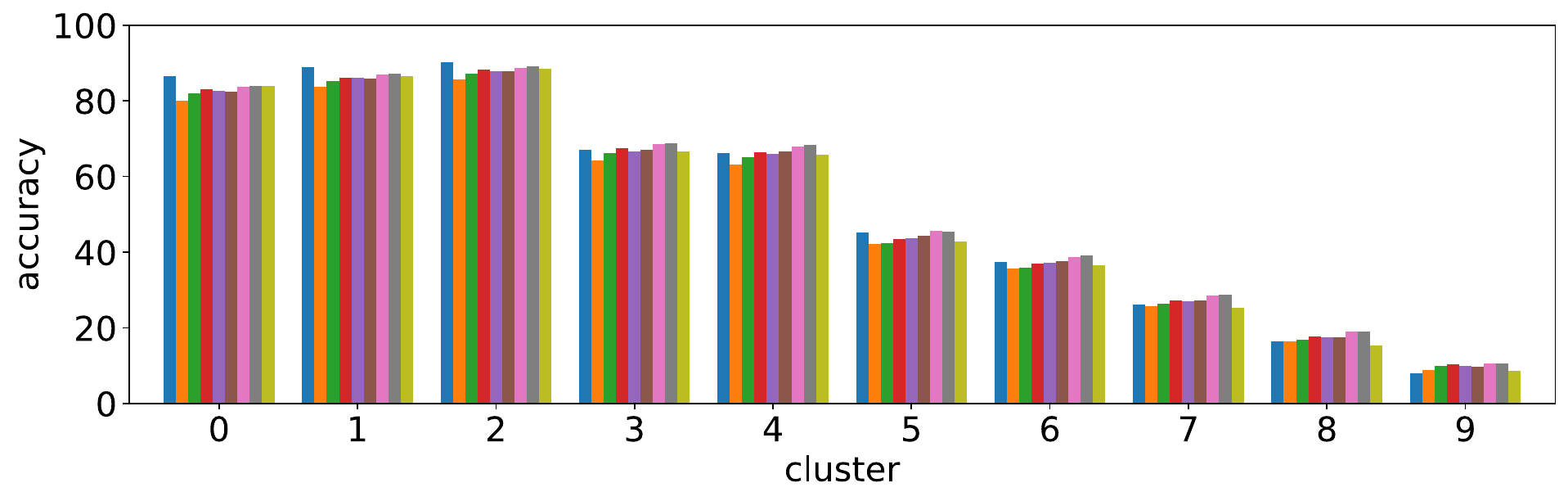}
    \includegraphics[height=\myfigheight]{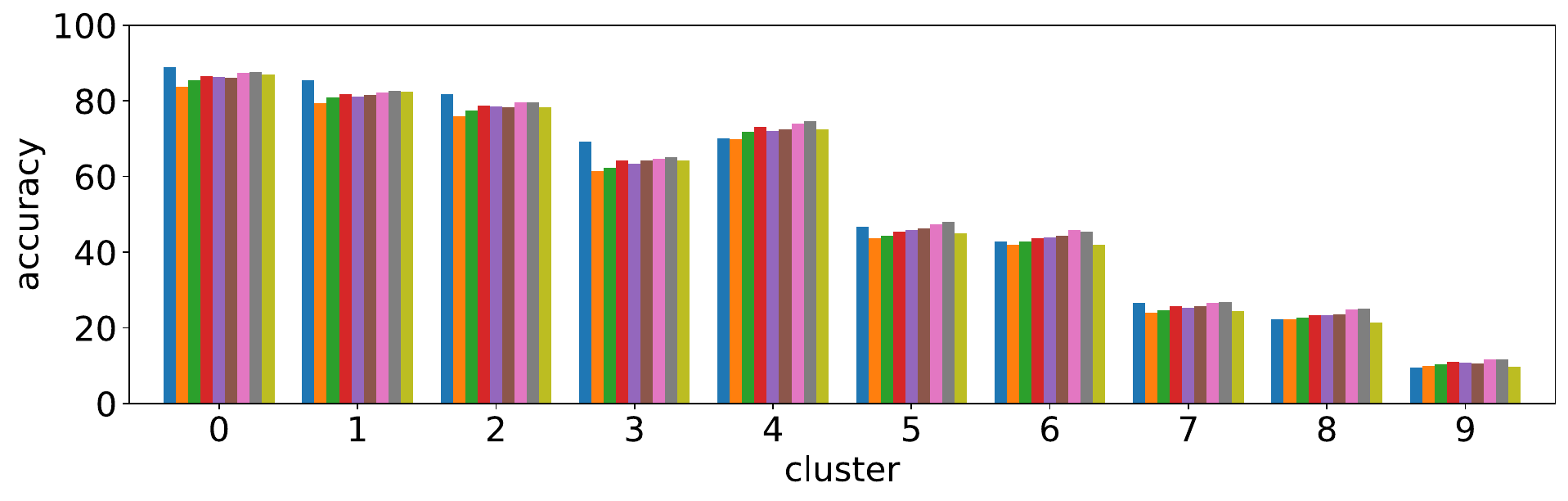}
    \includegraphics[height=\myfigheight]{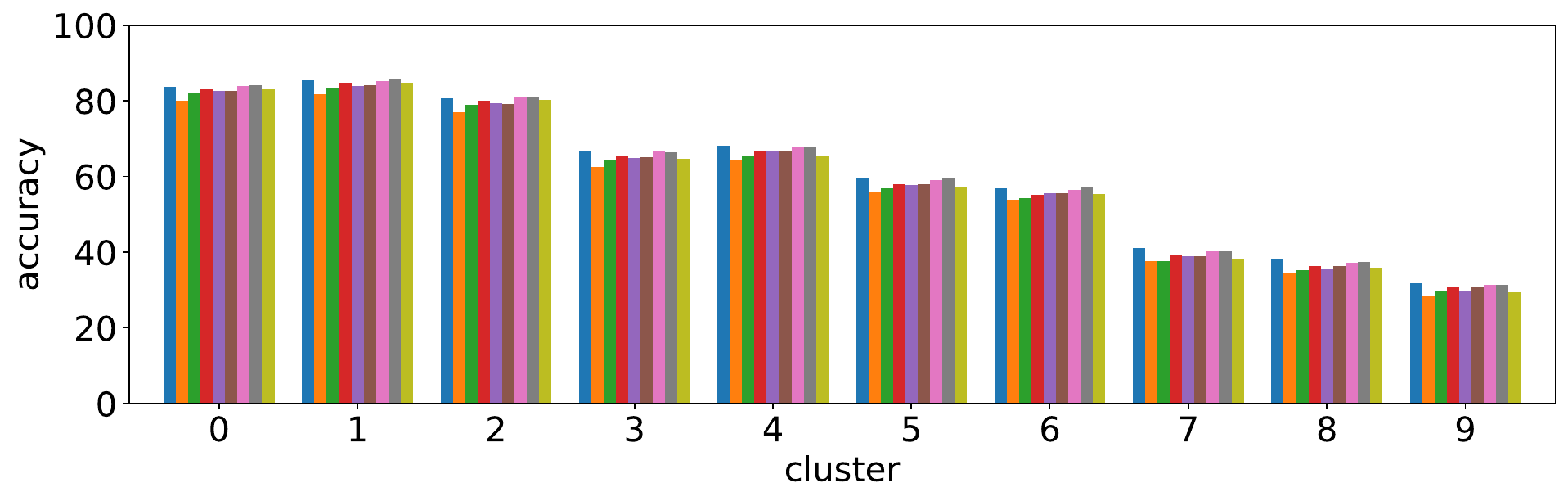}
    \includegraphics[height=\myfigheight]{supp_nolegend/i_q_iq_k10/accuracy_cluster_i_q_iq_0.pdf}
\end{minipage}
\hfill
\begin{minipage}[t]{.48\linewidth}
    \centering
    \includegraphics[height=\myfigheight]{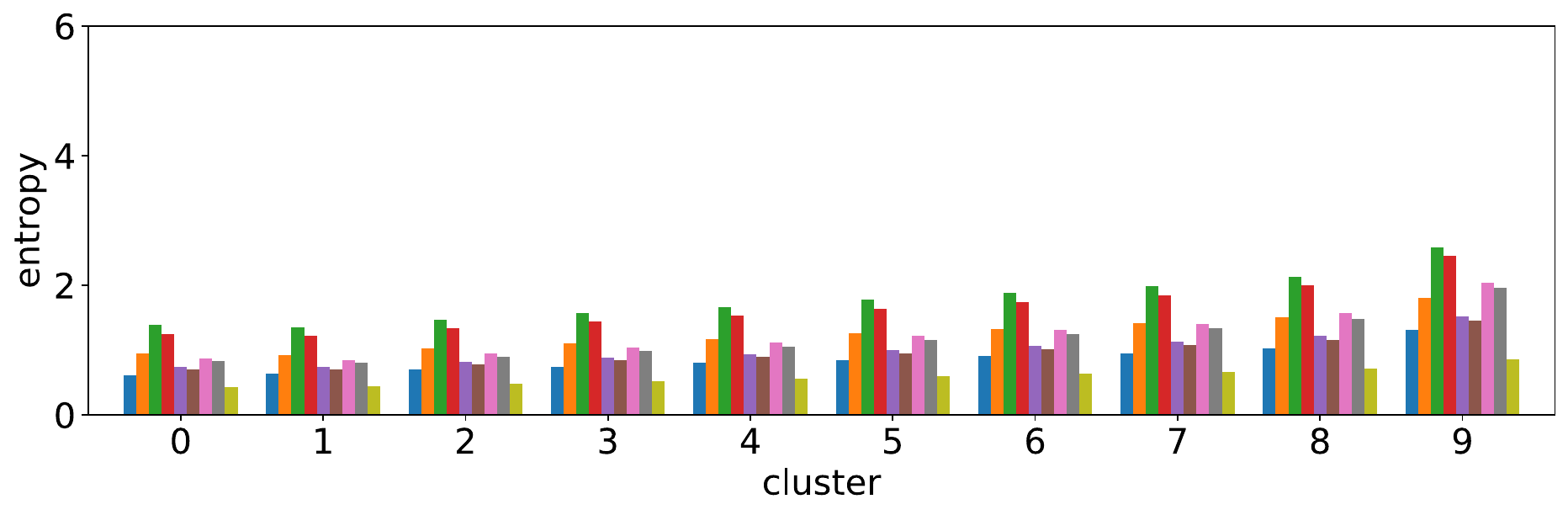}
    \includegraphics[height=\myfigheight]{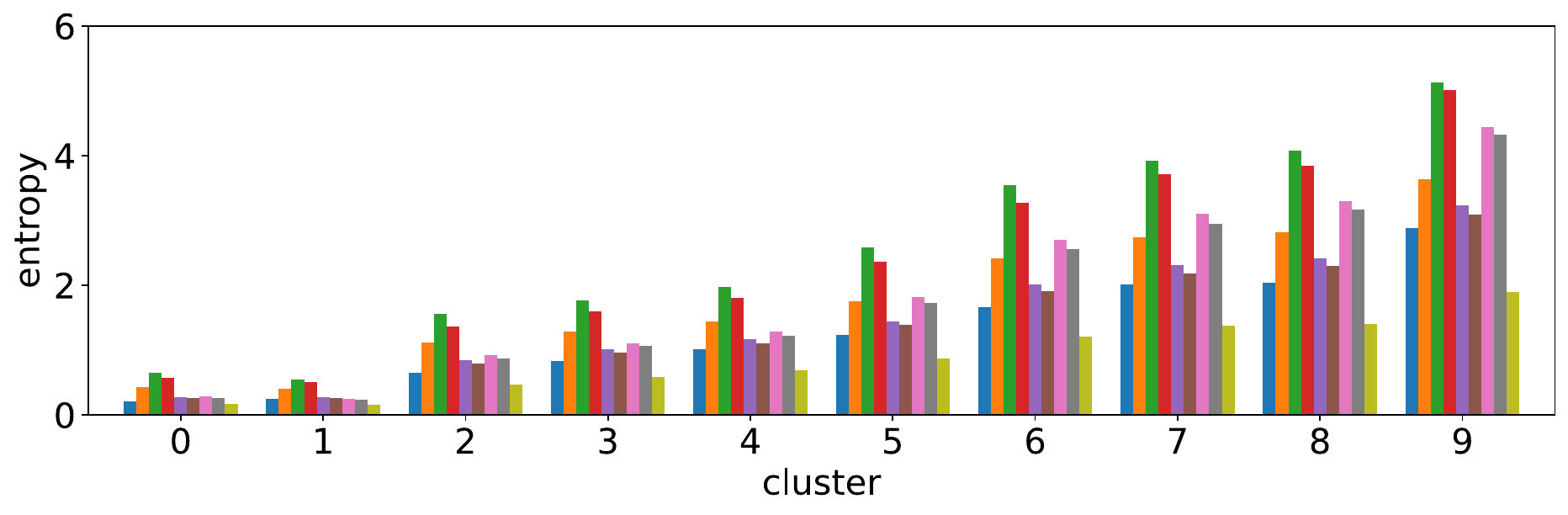}
    \includegraphics[height=\myfigheight]{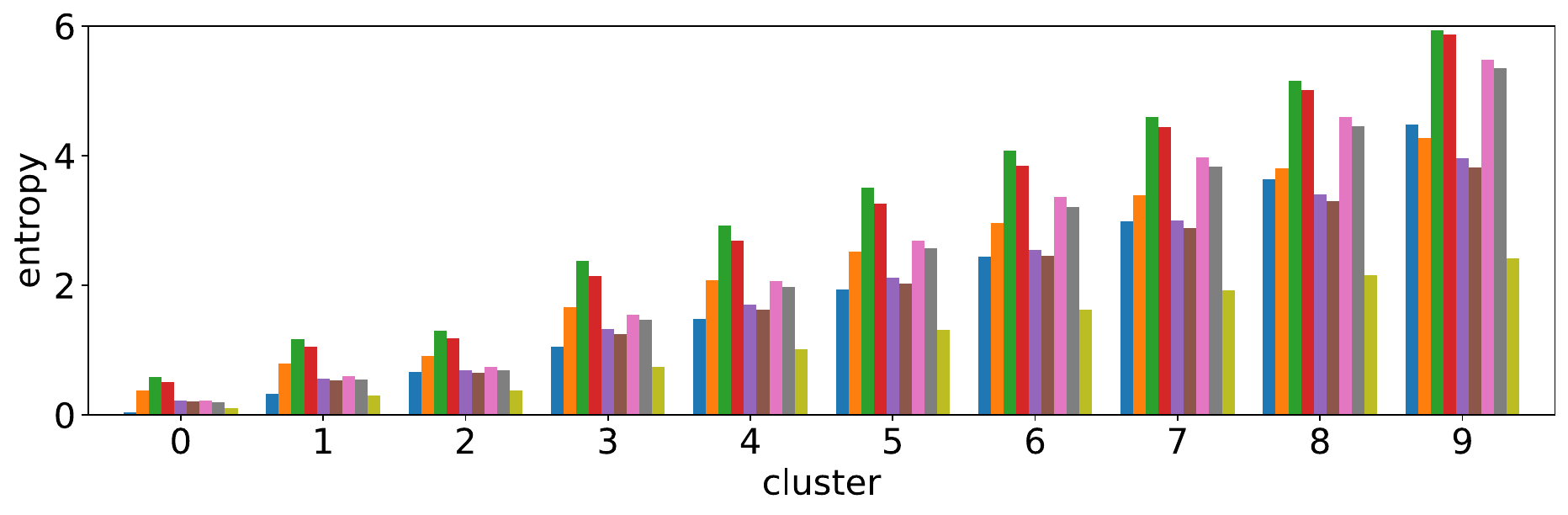}
    \includegraphics[height=\myfigheight]{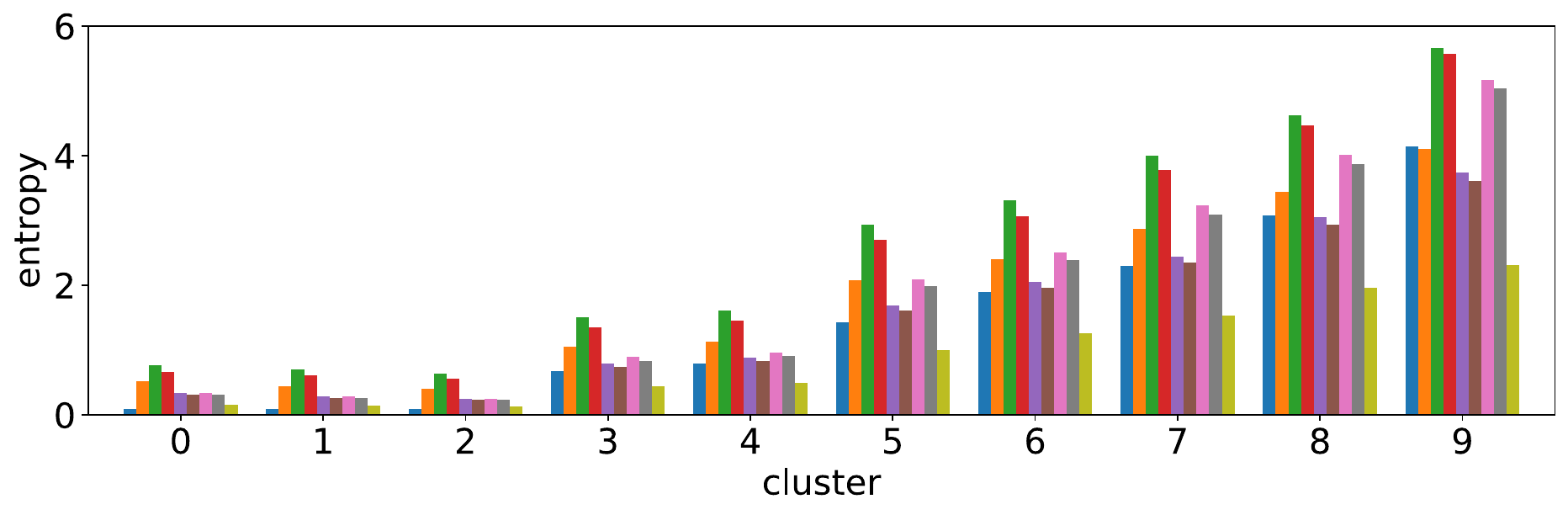}
    \includegraphics[height=\myfigheight]{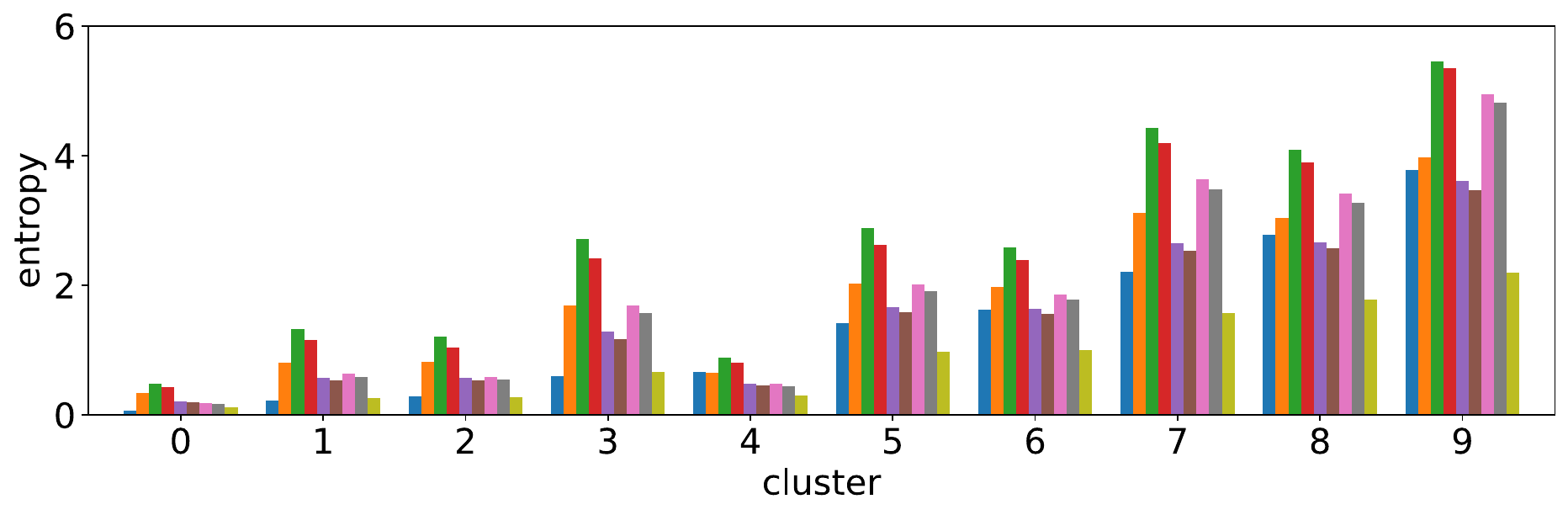}
    \includegraphics[height=\myfigheight]{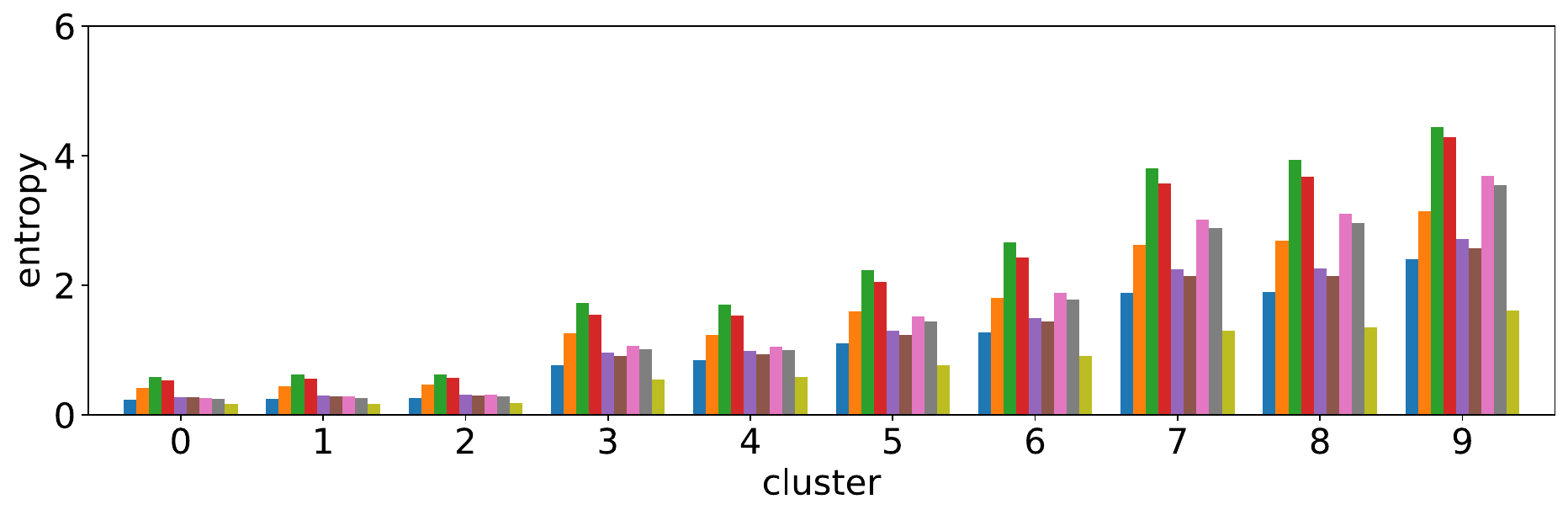}    
    \includegraphics[height=\myfigheight]{supp_nolegend/i_q_iq_k10/entropy_cluster_i_q_iq_0.pdf}
\end{minipage}

    \caption{Values of (left) accuracy and (right) entropy for each cluster when using different features in the $k$-means clustering algorithm.
    From top to bottom,
    (I), (Q), (Q+I),
    (I and Q+I),
    (Q an Q+I),
    (I and Q),
    and (I, Q, and Q+I)
    are used.
    The bottom row is the same with Figure \ref{fig:entropy_accuracy_relation}.
    }
    \label{fig:accuracy_entropy_i_q_iq}
\end{figure}

Figure \ref{fig:accuracy_entropy_i_q_iq}
shows results corresponding to Figure \ref{fig:entropy_accuracy_relation},
but when using different features in the  $k$-means clustering algorithm.
We use the entropy values of answer predictions 
obtained from three different models: I, Q, and Q+I.
Therefore we can use different subsets of the three models for clustering.

The first three rows of
Figure \ref{fig:accuracy_entropy_i_q_iq}
show the clustering results obtained when using only one of the three models.
The row (I) is obtained from the clustering result with the I model only,
and so on.
As expected, the results for the I model only show that there is
little correlation between accuracy and the order of clusters,
while the Q+I and Q models shows better correlations.
The next three rows of
Figure \ref{fig:accuracy_entropy_i_q_iq}
show clustering results obtained when using two of the three models together.
The row (I and Q+I) is obtained from the clustering result with the I and Q+I models,
and so on.
Again, the combinations with the Q model seems to affect the correlation.

\begin{figure}[t]
    \centering
    \includegraphics[width=.8\linewidth]{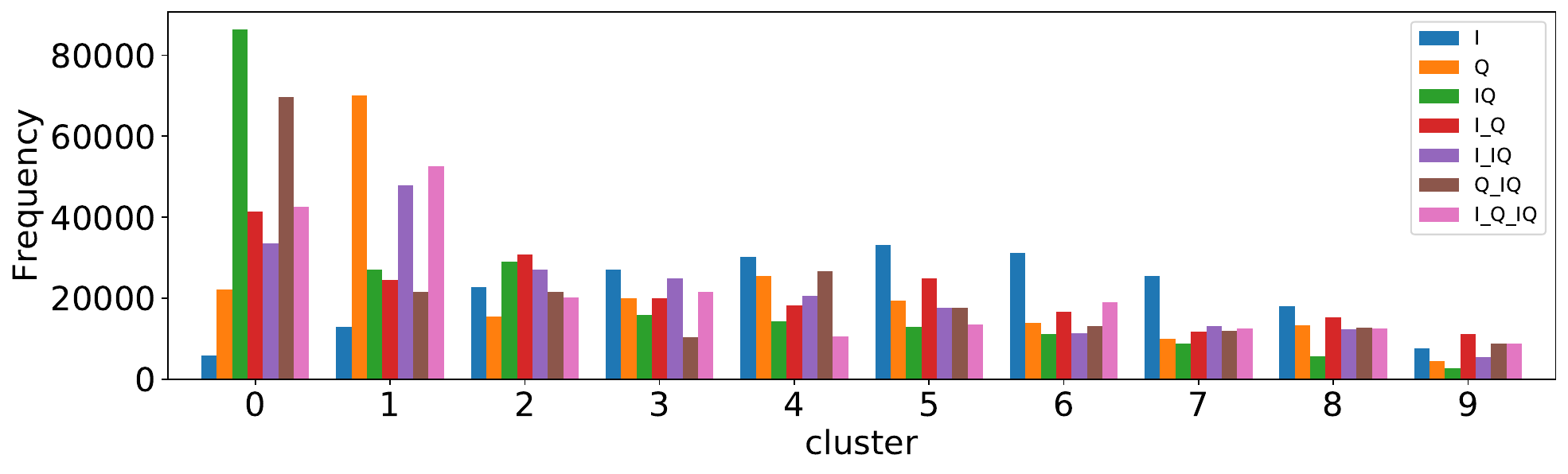}
        
    \caption{Number of samples for each cluster
    with different features for $k$-means clustering.
    }

    \label{fig:count_each_cluster}
\end{figure}

The row (Q+I) exhibits the largest correlation,
however this is due to the sample unbalance between clusters,
as shown in Figure \ref{fig:count_each_cluster}.
In contrast, the combination (I, Q, and Q+I)
has a better sample balance,
\ie samples are well distributed across clusters.
Future work includes a more qualitative investigation of this issue.

\subsubsection{Normalization of accuracy}

The clustering result (Figure \ref{fig:entropy_accuracy_relation}) shows
that large entropy leads to low accuracy.
However this might be due to the disagreement of answers by annotators.
As shown in Table \ref{tab:clustering_result},
clusters with large entropy values tend to disagree, \ie,
the entropy values and average number of unique answers of ground truth 
are larger.

The accuracy is defined by
Eq. \eqref{eq:accuracy_definition},
and it is bounded above 
by the number of ground truth answers.
If annotators disagree completely on a certain visual question,
each unique answer is provided by one annotator,
then the accuracy is at most 33\%.
This might cause an apparent reduction of accuracy for clusters with large entropy.

To remove this 
effect, we define a normalized version of accuracy as follows;
\begin{equation}
\text{normalized accuracy}(a)
= 
\frac{\text{accuracy}(a)}{\max_{A \in \mathcal{A}} \text{accuracy}(A)},
\end{equation}
where $\mathcal{A}$ is the the set of all possible answers.
This becomes 100\% for 
the case of complete disagreement
if the predicted answer is one among 10 different ground truth answers.

\begin{figure}[t]
    \centering
    \includegraphics[width=.8\linewidth]{supp_nolegend/i_q_iq_k10/accuracy_cluster_i_q_iq_0.pdf}
    \includegraphics[width=.8\linewidth]{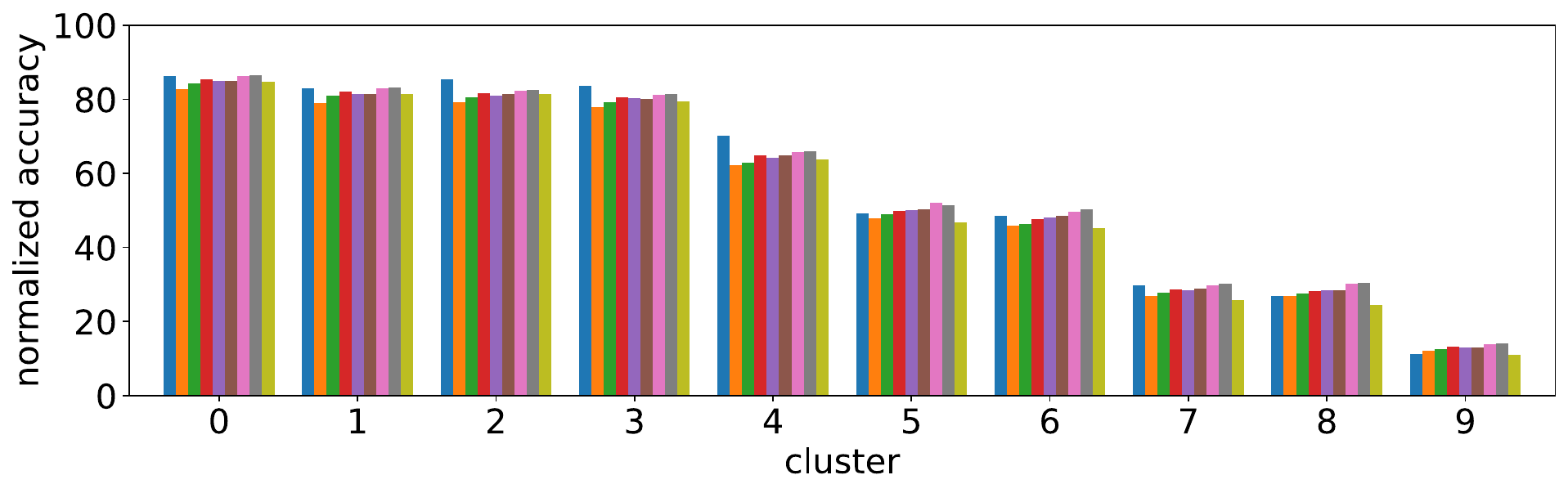}
        
    \caption{Values of (top) accuracy and (bottom) normalized accuracy for each cluster.
    }

    \label{fig:accuracy_norm}
\end{figure}

Figure \ref{fig:accuracy_norm}
shows results corresponding to Figure \ref{fig:entropy_accuracy_relation},
and results with the normalized accuracy.
As can be seen, the result for the normalized accuracy
is very similar to that obtained with the original accuracy.
Therefore we can conclude that
the result is due to the nature of our approach,
and not due to the effect of apparent reduction.


\section*{Acknowledgement}

We would like to thank 
Yoshitaka Ushiku for his support as a mentor of the PRMU mentorship program.

\bibliographystyle{IEEEtran}
\bibliography{egbib}

\begin{thebibliography}{10}
\providecommand{\url}[1]{#1}
\csname url@samestyle\endcsname
\providecommand{\newblock}{\relax}
\providecommand{\bibinfo}[2]{#2}
\providecommand{\BIBentrySTDinterwordspacing}{\spaceskip=0pt\relax}
\providecommand{\BIBentryALTinterwordstretchfactor}{4}
\providecommand{\BIBentryALTinterwordspacing}{\spaceskip=\fontdimen2\font plus
\BIBentryALTinterwordstretchfactor\fontdimen3\font minus
  \fontdimen4\font\relax}
\providecommand{\BIBforeignlanguage}[2]{{%
\expandafter\ifx\csname l@#1\endcsname\relax
\typeout{** WARNING: IEEEtran.bst: No hyphenation pattern has been}%
\typeout{** loaded for the language `#1'. Using the pattern for}%
\typeout{** the default language instead.}%
\else
\language=\csname l@#1\endcsname
\fi
#2}}
\providecommand{\BIBdecl}{\relax}
\BIBdecl

\bibitem{WU2017CVIU}
\BIBentryALTinterwordspacing
Q.~Wu, D.~Teney, P.~Wang, C.~Shen, A.~Dick, and A.~van~den Hengel, ``Visual
  question answering: A survey of methods and datasets,'' \emph{Computer Vision
  and Image Understanding}, vol. 163, pp. 21 -- 40, 2017, language in Vision.
  [Online]. Available:
  \url{http://www.sciencedirect.com/science/article/pii/S1077314217300772}
\BIBentrySTDinterwordspacing

\bibitem{VQAv1}
S.~Antol, A.~Agrawal, J.~Lu, M.~Mitchell, D.~Batra, C.~L. Zitnick, and
  D.~Parikh, ``{VQA}: {V}isual {Q}uestion {A}nswering,'' in \emph{International
  Conference on Computer Vision (ICCV)}, 2015.

\bibitem{Gurari_2018_CVPR}
D.~Gurari, Q.~Li, A.~J. Stangl, A.~Guo, C.~Lin, K.~Grauman, J.~Luo, and J.~P.
  Bigham, ``Vizwiz grand challenge: Answering visual questions from blind
  people,'' in \emph{The IEEE Conference on Computer Vision and Pattern
  Recognition (CVPR)}, June 2018.

\bibitem{embodiedqa}
A.~Das, S.~Datta, G.~Gkioxari, S.~Lee, D.~Parikh, and D.~Batra, ``{E}mbodied
  {Q}uestion {A}nswering,'' in \emph{Proceedings of the IEEE Conference on
  Computer Vision and Pattern Recognition (CVPR)}, 2018.

\bibitem{visual_dialog}
A.~Das, S.~Kottur, K.~Gupta, A.~Singh, D.~Yadav, J.~M. Moura, D.~Parikh, and
  D.~Batra, ``{V}isual {D}ialog,'' in \emph{Proceedings of the IEEE Conference
  on Computer Vision and Pattern Recognition (CVPR)}, 2017.

\bibitem{balanced_vqa_v2}
Y.~Goyal, T.~Khot, D.~Summers{-}Stay, D.~Batra, and D.~Parikh, ``Making the {V}
  in {VQA} matter: Elevating the role of image understanding in {V}isual
  {Q}uestion {A}nswering,'' in \emph{Conference on Computer Vision and Pattern
  Recognition (CVPR)}, 2017.

\bibitem{vizwiz}
\BIBentryALTinterwordspacing
J.~P. Bigham, C.~Jayant, H.~Ji, G.~Little, A.~Miller, R.~C. Miller, R.~Miller,
  A.~Tatarowicz, B.~White, S.~White, and T.~Yeh, ``Vizwiz: Nearly real-time
  answers to visual questions,'' in \emph{Proceedings of the 23Nd Annual ACM
  Symposium on User Interface Software and Technology}, ser. UIST '10.\hskip
  1em plus 0.5em minus 0.4em\relax New York, NY, USA: ACM, 2010, pp. 333--342.
  [Online]. Available: \url{http://doi.acm.org/10.1145/1866029.1866080}
\BIBentrySTDinterwordspacing

\bibitem{Gurari2017CHI}
\BIBentryALTinterwordspacing
D.~Gurari and K.~Grauman, ``Crowdverge: Predicting if people will agree on the
  answer to a visual question,'' in \emph{Proceedings of the 2017 CHI
  Conference on Human Factors in Computing Systems}, ser. CHI '17.\hskip 1em
  plus 0.5em minus 0.4em\relax New York, NY, USA: ACM, 2017, pp. 3511--3522.
  [Online]. Available: \url{http://doi.acm.org/10.1145/3025453.3025781}
\BIBentrySTDinterwordspacing

\bibitem{Bhattacharya_2019_ICCV}
N.~Bhattacharya, Q.~Li, and D.~Gurari, ``Why does a visual question have
  different answers?'' in \emph{The IEEE International Conference on Computer
  Vision (ICCV)}, October 2019.

\bibitem{Daniel:2018:QCC}
\BIBentryALTinterwordspacing
F.~Daniel, P.~Kucherbaev, C.~Cappiello, B.~Benatallah, and M.~Allahbakhsh,
  ``Quality control in crowdsourcing: A survey of quality attributes,
  assessment techniques, and assurance actions,'' \emph{ACM Comput. Surv.},
  vol.~51, no.~1, pp. 7:1--7:40, Jan. 2018. [Online]. Available:
  \url{http://doi.acm.org/10.1145/3148148}
\BIBentrySTDinterwordspacing

\bibitem{Soberon:2013:MCT}
\BIBentryALTinterwordspacing
G.~Sober\'{o}n, L.~Aroyo, C.~Welty, O.~Inel, H.~Lin, and M.~Overmeen,
  ``Measuring crowd truth: Disagreement metrics combined with worker behavior
  filters,'' in \emph{Proceedings of the 1st International Conference on
  Crowdsourcing the Semantic Web - Volume 1030}, ser. CrowdSem'13.\hskip 1em
  plus 0.5em minus 0.4em\relax Aachen, Germany, Germany: CEUR-WS.org, 2013, pp.
  45--58. [Online]. Available:
  \url{http://dl.acm.org/citation.cfm?id=2874376.2874381}
\BIBentrySTDinterwordspacing

\bibitem{corr-1301-2774}
\BIBentryALTinterwordspacing
J.~Muhammadi and H.~R. Rabiee, ``Crowd computing: a survey,'' \emph{CoRR}, vol.
  abs/1301.2774, 2013. [Online]. Available:
  \url{http://arxiv.org/abs/1301.2774}
\BIBentrySTDinterwordspacing

\bibitem{Malinowski_2015_ICCV}
M.~Malinowski, M.~Rohrbach, and M.~Fritz, ``Ask your neurons: A neural-based
  approach to answering questions about images,'' in \emph{The IEEE
  International Conference on Computer Vision (ICCV)}, December 2015.

\bibitem{DBLP:conf/hcomp/YangGG18}
\BIBentryALTinterwordspacing
C.~Yang, K.~Grauman, and D.~Gurari, ``Visual question answer diversity,'' in
  \emph{Proceedings of the Sixth {AAAI} Conference on Human Computation and
  Crowdsourcing, {HCOMP} 2018, Z{\"{u}}rich, Switzerland, July 5-8, 2018},
  2018, pp. 184--192. [Online]. Available:
  \url{https://aaai.org/ocs/index.php/HCOMP/HCOMP18/paper/view/17936}
\BIBentrySTDinterwordspacing

\bibitem{Agrawal_2018_CVPR}
A.~Agrawal, D.~Batra, D.~Parikh, and A.~Kembhavi, ``Don't just assume; look and
  answer: Overcoming priors for visual question answering,'' in \emph{The IEEE
  Conference on Computer Vision and Pattern Recognition (CVPR)}, June 2018.

\bibitem{Goyal2016}
\BIBentryALTinterwordspacing
Y.~Goyal, A.~Mohapatra, D.~Parikh, and D.~Batra, ``Towards transparent ai
  systems: Interpreting visual question answering models,'' in
  \emph{International Conference on Machine Learning (ICML) Workshop on
  Visualization for Deep Learning}, 2016. [Online]. Available:
  \url{https://icmlviz.github.io/icmlviz2016/assets/papers/22.pdf}
\BIBentrySTDinterwordspacing

\bibitem{Das2016}
\BIBentryALTinterwordspacing
A.~Das, H.~Agrawal, L.~Zitnick, D.~Parikh, and D.~Batra, ``Human attention in
  visual question answering: Do humans and deep networks look at the same
  regions?'' in \emph{International Conference on Machine Learning (ICML)
  Workshop on Visualization for Deep Learning}, 2016. [Online]. Available:
  \url{https://icmlviz.github.io/icmlviz2016/assets/papers/17.pdf}
\BIBentrySTDinterwordspacing

\bibitem{Jing_2020_AAAI}
C.~Jing, Y.~Wu, X.~Zhang, Y.~Jia, and Q.~Wu, ``Overcoming language priors in
  vqa via decomposed linguistic representations,'' in \emph{Thirty-Fourth AAAI
  Conference on Artificial Intelligence}, Feb 2020.

\bibitem{NIPS2018_7427}
\BIBentryALTinterwordspacing
S.~Ramakrishnan, A.~Agrawal, and S.~Lee, ``Overcoming language priors in visual
  question answering with adversarial regularization,'' in \emph{Advances in
  Neural Information Processing Systems 31}, S.~Bengio, H.~Wallach,
  H.~Larochelle, K.~Grauman, N.~Cesa-Bianchi, and R.~Garnett, Eds.\hskip 1em
  plus 0.5em minus 0.4em\relax Curran Associates, Inc., 2018, pp. 1541--1551.
  [Online]. Available:
  \url{http://papers.nips.cc/paper/7427-overcoming-language-priors-in-visual-question-answering-with-adversarial-regularization.pdf}
\BIBentrySTDinterwordspacing

\bibitem{singh2019TowardsVM}
A.~Singh, V.~Natarajan, M.~Shah, Y.~Jiang, X.~Chen, D.~Batra, D.~Parikh, and
  M.~Rohrbach, ``Towards vqa models that can read,'' in \emph{Proceedings of
  the IEEE Conference on Computer Vision and Pattern Recognition}, 2019.

\bibitem{fvqa}
\BIBentryALTinterwordspacing
P.~Wang, Q.~Wu, C.~Shen, A.~R. Dick, and A.~van~den Hengel, ``{FVQA:}
  fact-based visual question answering,'' \emph{{IEEE} Trans. Pattern Anal.
  Mach. Intell.}, vol.~40, no.~10, pp. 2413--2427, 2018. [Online]. Available:
  \url{https://doi.org/10.1109/TPAMI.2017.2754246}
\BIBentrySTDinterwordspacing

\bibitem{Marino_2019_CVPR}
K.~Marino, M.~Rastegari, A.~Farhadi, and R.~Mottaghi, ``Ok-vqa: A visual
  question answering benchmark requiring external knowledge,'' in \emph{The
  IEEE Conference on Computer Vision and Pattern Recognition (CVPR)}, June
  2019.

\bibitem{Singh_2019_ICCV}
A.~K. Singh, A.~Mishra, S.~Shekhar, and A.~Chakraborty, ``From strings to
  things: Knowledge-enabled vqa model that can read and reason,'' in \emph{The
  IEEE International Conference on Computer Vision (ICCV)}, October 2019.

\bibitem{Garcia_2020_AAAI}
N.~Garcia, M.~Otani, C.~Chu, and Y.~Nakashima, ``Knowit vqa: Answering
  knowledge-based questions about videos,'' in \emph{Thirty-Fourth AAAI
  Conference on Artificial Intelligence}, Feb 2020.

\bibitem{mostafazadeh2016ACL_VQG}
\BIBentryALTinterwordspacing
N.~Mostafazadeh, I.~Misra, J.~Devlin, M.~Mitchell, X.~He, and L.~Vanderwende,
  ``Generating natural questions about an image,'' in \emph{Proceedings of the
  54th Annual Meeting of the Association for Computational Linguistics (Volume
  1: Long Papers)}.\hskip 1em plus 0.5em minus 0.4em\relax Berlin, Germany:
  Association for Computational Linguistics, Aug. 2016, pp. 1802--1813.
  [Online]. Available: \url{https://www.aclweb.org/anthology/P16-1170}
\BIBentrySTDinterwordspacing

\bibitem{Li_2018_CVPR}
Y.~Li, N.~Duan, B.~Zhou, X.~Chu, W.~Ouyang, X.~Wang, and M.~Zhou, ``Visual
  question generation as dual task of visual question answering,'' in \emph{The
  IEEE Conference on Computer Vision and Pattern Recognition (CVPR)}, June
  2018.

\bibitem{Jain_2018_CVPR}
U.~Jain, S.~Lazebnik, and A.~G. Schwing, ``Two can play this game: Visual
  dialog with discriminative question generation and answering,'' in \emph{The
  IEEE Conference on Computer Vision and Pattern Recognition (CVPR)}, June
  2018.

\bibitem{Shen_2019_ICCV}
T.~Shen, A.~Kar, and S.~Fidler, ``Learning to caption images through a lifetime
  by asking questions,'' in \emph{The IEEE International Conference on Computer
  Vision (ICCV)}, October 2019.

\bibitem{zhu2016cvpr}
Y.~Zhu, O.~Groth, M.~Bernstein, and L.~Fei-Fei, ``{Visual7W: Grounded Question
  Answering in Images},'' in \emph{{IEEE Conference on Computer Vision and
  Pattern Recognition}}, 2016.

\bibitem{Yu_2015_ICCV}
L.~Yu, E.~Park, A.~C. Berg, and T.~L. Berg, ``Visual madlibs: Fill in the blank
  description generation and question answering,'' in \emph{The IEEE
  International Conference on Computer Vision (ICCV)}, December 2015.

\bibitem{Johnson_2017_CVPR}
J.~Johnson, B.~Hariharan, L.~van~der Maaten, L.~Fei-Fei, C.~Lawrence~Zitnick,
  and R.~Girshick, ``Clevr: A diagnostic dataset for compositional language and
  elementary visual reasoning,'' in \emph{The IEEE Conference on Computer
  Vision and Pattern Recognition (CVPR)}, July 2017.

\bibitem{COCO-QA}
\BIBentryALTinterwordspacing
M.~Ren, R.~Kiros, and R.~Zemel, ``Exploring models and data for image question
  answering,'' in \emph{Advances in Neural Information Processing Systems 28},
  C.~Cortes, N.~D. Lawrence, D.~D. Lee, M.~Sugiyama, and R.~Garnett, Eds.\hskip
  1em plus 0.5em minus 0.4em\relax Curran Associates, Inc., 2015, pp.
  2953--2961. [Online]. Available:
  \url{http://papers.nips.cc/paper/5640-exploring-models-and-data-for-image-question-answering.pdf}
\BIBentrySTDinterwordspacing

\bibitem{kafle2017analysis}
K.~Kafle and C.~Kanan, ``An analysis of visual question answering algorithms,''
  in \emph{ICCV}, 2017.

\bibitem{krishnavisualgenome}
\BIBentryALTinterwordspacing
R.~Krishna, Y.~Zhu, O.~Groth, J.~Johnson, K.~Hata, J.~Kravitz, S.~Chen,
  Y.~Kalantidis, L.-J. Li, D.~A. Shamma, M.~Bernstein, and L.~Fei-Fei, ``Visual
  genome: Connecting language and vision using crowdsourced dense image
  annotations,'' in \emph{CoRR}, 2016. [Online]. Available:
  \url{https://arxiv.org/abs/1602.07332}
\BIBentrySTDinterwordspacing

\bibitem{FM-IQA}
\BIBentryALTinterwordspacing
H.~Gao, J.~Mao, J.~Zhou, Z.~Huang, L.~Wang, and W.~Xu, ``Are you talking to a
  machine? dataset and methods for multilingual image question,'' in
  \emph{Advances in Neural Information Processing Systems 28}, C.~Cortes, N.~D.
  Lawrence, D.~D. Lee, M.~Sugiyama, and R.~Garnett, Eds.\hskip 1em plus 0.5em
  minus 0.4em\relax Curran Associates, Inc., 2015, pp. 2296--2304. [Online].
  Available:
  \url{http://papers.nips.cc/paper/5641-are-you-talking-to-a-machine-dataset-and-methods-for-multilingual-image-question.pdf}
\BIBentrySTDinterwordspacing

\bibitem{yu2018beyond}
Z.~Yu, J.~Yu, C.~Xiang, J.~Fan, and D.~Tao, ``Beyond bilinear: Generalized
  multimodal factorized high-order pooling for visual question answering,''
  \emph{IEEE Transactions on Neural Networks and Learning Systems}, vol.~29,
  no.~12, pp. 5947--5959, 2018.

\bibitem{Nam_2017_CVPR}
H.~Nam, J.-W. Ha, and J.~Kim, ``Dual attention networks for multimodal
  reasoning and matching,'' in \emph{The IEEE Conference on Computer Vision and
  Pattern Recognition (CVPR)}, July 2017.

\bibitem{Xu_2016_ECCV}
H.~Xu and K.~Saenko, ``Ask, attend and answer: Exploring question-guided
  spatial attention for visual question answering,'' in \emph{Computer Vision
  -- ECCV 2016}, B.~Leibe, J.~Matas, N.~Sebe, and M.~Welling, Eds.\hskip 1em
  plus 0.5em minus 0.4em\relax Cham: Springer International Publishing, 2016,
  pp. 451--466.

\bibitem{Nguyen_2018_CVPR}
D.-K. Nguyen and T.~Okatani, ``Improved fusion of visual and language
  representations by dense symmetric co-attention for visual question
  answering,'' in \emph{The IEEE Conference on Computer Vision and Pattern
  Recognition (CVPR)}, June 2018.

\bibitem{NIPS2018_7429}
\BIBentryALTinterwordspacing
J.-H. Kim, J.~Jun, and B.-T. Zhang, ``Bilinear attention networks,'' in
  \emph{Advances in Neural Information Processing Systems 31}, S.~Bengio,
  H.~Wallach, H.~Larochelle, K.~Grauman, N.~Cesa-Bianchi, and R.~Garnett,
  Eds.\hskip 1em plus 0.5em minus 0.4em\relax Curran Associates, Inc., 2018,
  pp. 1564--1574. [Online]. Available:
  \url{http://papers.nips.cc/paper/7429-bilinear-attention-networks.pdf}
\BIBentrySTDinterwordspacing

\bibitem{Yu_2019_CVPR}
Z.~Yu, J.~Yu, Y.~Cui, D.~Tao, and Q.~Tian, ``Deep modular co-attention networks
  for visual question answering,'' in \emph{The IEEE Conference on Computer
  Vision and Pattern Recognition (CVPR)}, June 2019.

\bibitem{Chen_2020_CVPR}
L.~Chen, X.~Yan, J.~Xiao, H.~Zhang, S.~Pu, and Y.~Zhuang, ``Counterfactual
  samples synthesizing for robust visual question answering,'' in
  \emph{IEEE/CVF Conference on Computer Vision and Pattern Recognition (CVPR)},
  June 2020.

\bibitem{Wang_2020_CVPR}
X.~Wang, Y.~Liu, C.~Shen, C.~C. Ng, C.~Luo, L.~Jin, C.~S. Chan, A.~v.~d.
  Hengel, and L.~Wang, ``On the general value of evidence, and bilingual
  scene-text visual question answering,'' in \emph{IEEE/CVF Conference on
  Computer Vision and Pattern Recognition (CVPR)}, June 2020.

\bibitem{Agarwal_2020_CVPR}
V.~Agarwal, R.~Shetty, and M.~Fritz, ``Towards causal vqa: Revealing and
  reducing spurious correlations by invariant and covariant semantic editing,''
  in \emph{IEEE/CVF Conference on Computer Vision and Pattern Recognition
  (CVPR)}, June 2020.

\bibitem{Patro_2020_AAAI}
B.~N. Patro, Anupriy, and V.~P. Namboodiri, ``Explanation vs attention: A
  two-player game to obtain attention for vqa,'' in \emph{Thirty-Fourth AAAI
  Conference on Artificial Intelligence}, Feb 2020.

\bibitem{Teney_2018_CVPR}
D.~Teney, P.~Anderson, X.~He, and A.~van~den Hengel, ``Tips and tricks for
  visual question answering: Learnings from the 2017 challenge,'' in \emph{The
  IEEE Conference on Computer Vision and Pattern Recognition (CVPR)}, June
  2018.

\bibitem{Wang_2015_ICCV}
X.~Wang and A.~Gupta, ``Unsupervised learning of visual representations using
  videos,'' in \emph{The IEEE International Conference on Computer Vision
  (ICCV)}, December 2015.

\bibitem{Shrivastava_2016_CVPR}
A.~Shrivastava, A.~Gupta, and R.~Girshick, ``Training region-based object
  detectors with online hard example mining,'' in \emph{The IEEE Conference on
  Computer Vision and Pattern Recognition (CVPR)}, June 2016.

\bibitem{Wang_2018_ECCV}
P.~Wang and N.~Vasconcelos, ``Towards realistic predictors,'' in \emph{The
  European Conference on Computer Vision (ECCV)}, September 2018.

\bibitem{pythia2018arxiv}
\BIBentryALTinterwordspacing
Y.~Jiang, V.~Natarajan, X.~Chen, M.~Rohrbach, D.~Batra, and D.~Parikh, ``Pythia
  v0.1: the winning entry to the {VQA} challenge 2018,'' \emph{CoRR}, vol.
  abs/1807.09956, 2018. [Online]. Available:
  \url{http://arxiv.org/abs/1807.09956}
\BIBentrySTDinterwordspacing

\bibitem{singh2018pythia}
A.~Singh, V.~Natarajan, Y.~Jiang, X.~Chen, M.~Shah, M.~Rohrbach, D.~Batra, and
  D.~Parikh, ``Pythia-a platform for vision \& language research,'' in
  \emph{SysML Workshop, NeurIPS}, vol. 2018, 2018.

\bibitem{Anderson_2018_CVPR}
P.~Anderson, X.~He, C.~Buehler, D.~Teney, M.~Johnson, S.~Gould, and L.~Zhang,
  ``Bottom-up and top-down attention for image captioning and visual question
  answering,'' in \emph{The IEEE Conference on Computer Vision and Pattern
  Recognition (CVPR)}, June 2018.

\bibitem{lu2019vilbert}
J.~Lu, D.~Batra, D.~Parikh, and S.~Lee, ``Vilbert: Pretraining task-agnostic
  visiolinguistic representations for vision-and-language tasks,'' \emph{arXiv
  preprint arXiv:1908.02265}, 2019.

\bibitem{fukui-etal-2016-multimodal}
\BIBentryALTinterwordspacing
A.~Fukui, D.~H. Park, D.~Yang, A.~Rohrbach, T.~Darrell, and M.~Rohrbach,
  ``Multimodal compact bilinear pooling for visual question answering and
  visual grounding,'' in \emph{Proceedings of the 2016 Conference on Empirical
  Methods in Natural Language Processing}.\hskip 1em plus 0.5em minus
  0.4em\relax Austin, Texas: Association for Computational Linguistics, Nov.
  2016, pp. 457--468. [Online]. Available:
  \url{https://www.aclweb.org/anthology/D16-1044}
\BIBentrySTDinterwordspacing

\bibitem{NIPS2016_6202}
\BIBentryALTinterwordspacing
J.~Lu, J.~Yang, D.~Batra, and D.~Parikh, ``Hierarchical question-image
  co-attention for visual question answering,'' in \emph{Advances in Neural
  Information Processing Systems 29}, D.~D. Lee, M.~Sugiyama, U.~V. Luxburg,
  I.~Guyon, and R.~Garnett, Eds.\hskip 1em plus 0.5em minus 0.4em\relax Curran
  Associates, Inc., 2016, pp. 289--297. [Online]. Available:
  \url{http://papers.nips.cc/paper/6202-hierarchical-question-image-co-attention-for-visual-question-answering.pdf}
\BIBentrySTDinterwordspacing

\bibitem{Yu_2017_ICCV}
Z.~Yu, J.~Yu, J.~Fan, and D.~Tao, ``Multi-modal factorized bilinear pooling
  with co-attention learning for visual question answering,'' in \emph{The IEEE
  International Conference on Computer Vision (ICCV)}, Oct 2017.

\bibitem{vqd}
T.~Tamaki, ``{Visual Question Difficulty (VQD)},''
  \url{https://github.com/tttamaki/vqd/}, March 2020,
  doi:10.5281/zenodo.3725534.

\bibitem{Arthur2007kmeans++}
D.~Arthur and S.~Vassilvitskii, ``K-means++: The advantages of careful
  seeding,'' in \emph{Proceedings of the Eighteenth Annual ACM-SIAM Symposium
  on Discrete Algorithms}, ser. SODA ’07.\hskip 1em plus 0.5em minus
  0.4em\relax USA: Society for Industrial and Applied Mathematics, 2007, p.
  1027–1035.

\end{thebibliography}


\vspace{-1.5cm}
\begin{IEEEbiography}[{\includegraphics[width=1in,height=1.25in,clip,keepaspectratio]{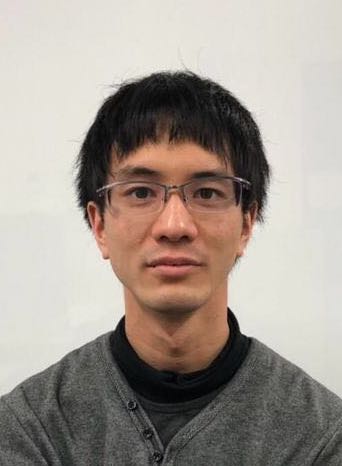}}]{Kento Terao} received his B.E. and M.S. degrees in information engineering from Hiroshima University, Japan, in 2018 and 2020, respectively. His research interests include visual question answering and generation.
\end{IEEEbiography}
\vspace{-1.5cm}
\begin{IEEEbiography}[{\includegraphics[width=1in,height=1.25in,clip,keepaspectratio]{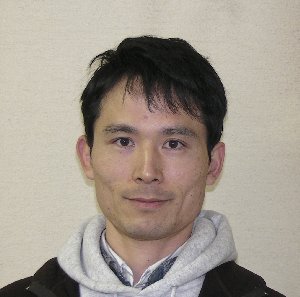}}]{Toru Tamaki} received his B.E., M.S., and Ph.D. degrees in information engineering from Nagoya University, Japan, in 1996, 1998 and 2001, respectively. After being an assistant professor at Niigata University, Japan, from 2001 to 2005, he is currently an associate professor at the Department of Information Engineering, Graduate School of Engineering, Hiroshima University, Japan. He was an associate researcher at ESIEE Paris, France, in 2015. His research interests include computer vision, image recognition, and machine learning.
\end{IEEEbiography}
\vspace{-1.5cm}
\begin{IEEEbiography}[{\includegraphics[width=1in,height=1.25in,clip,keepaspectratio]{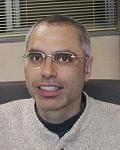}}]{Bisser Raytchev} received his Ph.D. in Informatics from Tsukuba University, Japan in 2000. After being a research associate at NTT Communication Science Labs and AIST, he is presently an associate professor in the Department of Information Engineering, Hiroshima University, Japan. His current research interests include machine learning, computer vision, natural language processing and brain-inspired computing.
\end{IEEEbiography}
\vspace{-1.5cm}
\begin{IEEEbiography}[{\includegraphics[width=1in,height=1.25in,clip,keepaspectratio]{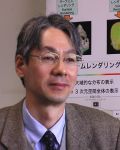}}]{Kazufumi Kaneda} is a professor in the Department of Information Engineering, Hiroshima University, Japan. He joined Hiroshima University in 1986. He was a visiting researcher in the Engineering Computer Graphics laboratory at Brigham Young University in 1991. Kaneda received the B.E., M.E., and D.E. in 1982, 1984 and 1991, respectively, from Hiroshima University. His research interests include computer graphics, scientific visualization, and image processing.
\end{IEEEbiography}
\vspace{-1.5cm}
\begin{IEEEbiography}[{\includegraphics[width=1in,height=1.25in,clip,keepaspectratio]{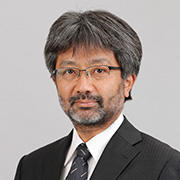}}]{Shin'ichi Satoh} received the B.E. degree in electronics engineering and the M.E. and Ph.D. degrees in information engineering from the University of Tokyo, Tokyo, Japan, in 1987, 1989, and 1992, respectively. He was a Visiting Scientist with the Robotics Institute, Carnegie Mellon University, Pittsburgh, PA, USA, from 1995 to 1997. He has been a Full Professor with the National Institute of Informatics, Tokyo, Japan, since 2004. His current research interests include image processing, video content analysis, and multimedia databases.
\end{IEEEbiography}
\EOD

\end{document}